\documentclass[10pt,twocolumn,letterpaper,table]{article}
\usepackage[pagenumbers]{iccv} % To force page numbers, e.g. for an arXiv version
\usepackage{titletoc}
\usepackage{listings}

\newcommand{\mysection}[1]{\noindent\textbf{#1.}}

\definecolor{iccvblue}{rgb}{0.21,0.49,0.74}
\usepackage[pagebackref,breaklinks,colorlinks,allcolors=iccvblue]{hyperref}

\def\paperID{8498} % *** Enter the Paper ID here
\def\confName{ICCV}
\def\confYear{2025}

\title{4D-Bench: Benchmarking Multi-modal Large Language Models for 4D Object Understanding}

\author{
	\textbf{Wenxuan Zhu$^{1*}$, Bing Li$^{1*}$, Cheng Zheng$^{1*}$ 
        Jinjie Mai$^{1}$, Jun Chen$^{1}$, Letian Jiang$^{1}$, Abdullah Hamdi$^{2}$,} \\ \textbf{Sara Rojas Martinez$^{1}$, Chia-Wen Lin$^{3}$, Mohamed Elhoseiny$^{1}$, Bernard Ghanem$^{1}$} \\
	$^1$King Abdullah University of Science and Technology, $^2$University of Oxford, $^3$National Tsing Hua University
}

\begin{document}
\maketitle
\begin{abstract}
Multimodal Large Language Models (MLLMs) have demonstrated impressive 2D image/video understanding capabilities.
However, there are no publicly standardized benchmarks to assess the abilities of MLLMs in understanding the 4D objects (3D objects with temporal evolution over time).
In this paper, we introduce 4D-Bench, the first benchmark to evaluate the capabilities of MLLMs in 4D object understanding, featuring tasks in 4D object Question Answering (4D object QA) and 4D object captioning.
4D-Bench provides 4D objects with diverse categories, high-quality annotations, and tasks necessitating multi-view spatial-temporal understanding, different from existing 2D image/video-based benchmarks.
With 4D-Bench, we evaluate a wide range of open-source and closed-source MLLMs.
The results from the 4D object captioning experiment indicate that MLLMs generally exhibit weaker temporal understanding compared to their appearance understanding, notably, while open-source models approach closed-source performance in appearance understanding, they show larger performance gaps in temporal understanding.
4D object QA yields surprising findings: even with simple single-object videos, MLLMs perform poorly, with state-of-the-art GPT-4o achieving only 63\% accuracy compared to the human baseline of 91\%.
These findings highlight a substantial gap in 4D object understanding and the need for further advancements in MLLMs. Project page: \href{https://wenxuanzhu1103.github.io/4dbench.github.io/}{https://4dbench.github.io/}
\end{abstract}

\section{Introduction}
\label{sec:intro}
\begin{figure}[!htb]
\centering
\includegraphics[width=1.0\linewidth]{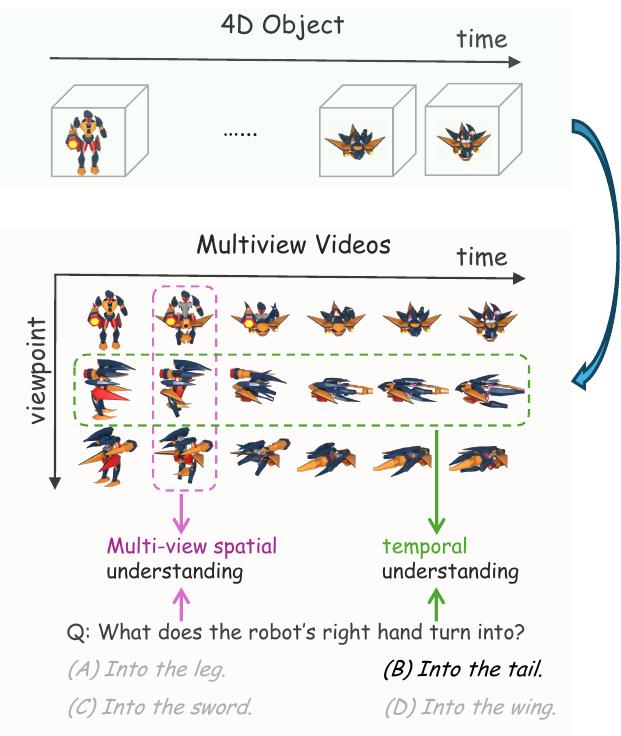}
\caption{\textbf{An example demonstrating the challenges of 4D object understanding involves multi-view spatial-temporal reasoning.} Given the 4D object, the robot's right hand seems ambiguous in some views at first and eventually disappears over time.
Hence, answering the question needs to (1) address multi-view ambiguity and choose proper views and time that the right hand is visible, (2) localize the right hand, (3) and track its evolutions along the time dimension.}
\label{fig:example_4Dobject}
\vspace{-10pt}
\end{figure}

Digital 4D (\ie dynamic 3D) assets have received increasing attention from both academia \cite{4dfy,4diffusion,diffusion4d,vividzoo,DreamGaussian4D} and industry \cite{4dasset1,4dasset2}, as they are important to many real-world applications such as digital twins, augmented reality, and gaming.
With the increasing demand for dynamic and interactive virtual experiences,  it is desirable to understand and interact with 4D assets using language, necessitating 4D-object-language understanding for 4D assets.

While many efforts \cite{minigptv2,minigpt,blip2,alayrac2022flamingo,LLaVA,internvl,qwen2_vl} have been devoted to 2D image/video language understanding, 4D object language understanding has been much less underexplored, yet it poses new challenges.
First, unlike 2D images, where parts of an object are occluded or ambiguous, a 4D object can be observed from different views, exhibiting different appearances among views and dynamic motions over time.
As a result, 4D object understanding requires both multi-view spatial and temporal understanding (see Fig. \ref{fig:example_4Dobject}).
Additionally, diverse 4D representations (\eg point cloud squences \cite{deform4dthing,motion2vecsets}, 4DGS \cite{4DGS}),  add more difficulties in 4D object understanding.
Second, unlike the massive availability of 2D image-text data on the Internet, large-scale 4D-object-text data are scarce, hindering the development of 4D-object-centric foundation models.

In this paper, instead of costly building a large-scale 4D-object-text dataset and establishing a 4D object understanding model on advanced 4D representation (\eg point clouds, 4DGS), we explore a new question. \textit{Can we directly expand advanced Multi-modal Large Language Models (MLLMs) to 4D object understanding?}  
Current MLLMs, such as GPT-4o \cite{gpt4} and Qwen2-VL \cite{qwen2_vl}, have learned rich world knowledge from massive text, image and video data. 
By representing 4D objects as multi-view videos, we can leverage MLLMs for 4D object language understanding.
However, a significant challenge arises:  there are no such public benchmarks designed for evaluating 4D object language understanding abilities, to the best of our knowledge.
Without a dedicated benchmark, it is unclear what the strengths and limitations of these models are in 4D object understanding, thereby making it difficult to improve MLLMs and unlock their potential.

To fill the gap, we step towards 4D object language understanding by introducing a new benchmark, dubbed 4D-Bench. The 4D-bench presents 4D object captioning and 4D object Question Answering (QA) tasks,  enabling an in-depth evaluation of MLLMs.
Due to the lack of publicly available high-quality text descriptions for 4D objects,  it is non-trivial to construct annotations through leveraging text information in existing 4D object datasets, unlike 2D images/videos~\cite{mvbench}.
Instead, we devote great human efforts to manually ensure that most questions necessitate multi-view spatial-temporal understanding for 4D object QA, so that our 4D-Bench provides high-quality annotations yet challenging evaluations.

Our 4D-Bench introduces new dimensions in evaluating MLLMs, compared to 2D image/video benchmarks. First,  our benchmark necessitates both multi-view spatial and temporal understanding, which has been ignored by existing 3D- and 2D-language understanding benchmarks. For example, 3D-language understanding benchmarks (\eg \cite{scanqa,sceneverse}) focus on static 3D scene understanding, ignoring motion information, while 2D video benchmarks (\eg \cite{mmbenchvideo,videomme} ) ignore multi-view understanding.
%Second, the multi-view temporal understanding is similar to  how human beings perceive objects  and reasoning in the real world
Second, our 4D-Bench comprises digital 4D assets, which are synthetic and include counterfactual objects and motions, typically absent in real-world datasets. This enables our 4D-Bench to be an Out-Of-Distribution (OOD) evaluation for MLLMs trained on real-world, scene-level 2D images/videos.

With 4D-Bench, we evaluate various MLLMs ranging from closed-source models such as Gemini 1.5 Pro \cite{gemini2024} and GPT-4o \cite{openai2024gpt4} to open-source ones (\eg Qwen2-VL \cite{qwen2_vl}).
Our extensive experiments reveal several key insights about current MLLMs' 4D object understanding capabilities: (1) Even state-of-the-art models still perform notably worse than humans across both question answering and captioning tasks; (2) On the 4D object QA task, MLLMs demonstrate a clear performance hierarchy across different understanding dimensions: they perform relatively better on appearance and spatial relationship subtasks but struggle considerably with object counting (37.29\% average accuracy), action recognition, and temporal relationship understanding; (3) 4D object captioning experimental results shows a similar pattern where MLLMs generally achieved higher GPT-Appearance scores than GPT-Action scores. Notably, closed-source models generally outperform open-source alternatives, particularly in action understanding, some open-source models show competitive performance in appearance comprehension.

Our contributions can be summarized as follows:
\begin{itemize}
    \item We introduce 4D-Bench, the first comprehensive benchmark for evaluating MLLMs' capabilities in understanding 4D objects, featuring both captioning and question-answering tasks.
    
   \item Our benchmark provides new challenges, necessitating multi-view spatial-temporal understanding, while it can serve as a generalization evaluation benchmark for image/video MLLMs.
    \item Evaluation results effectively reveal the strengths and shortcomings of the evaluated MLLMs in 4D object understanding.
    
\end{itemize}
\begin{figure*}[t]
    \centering
    \includegraphics[width=1.0\textwidth]{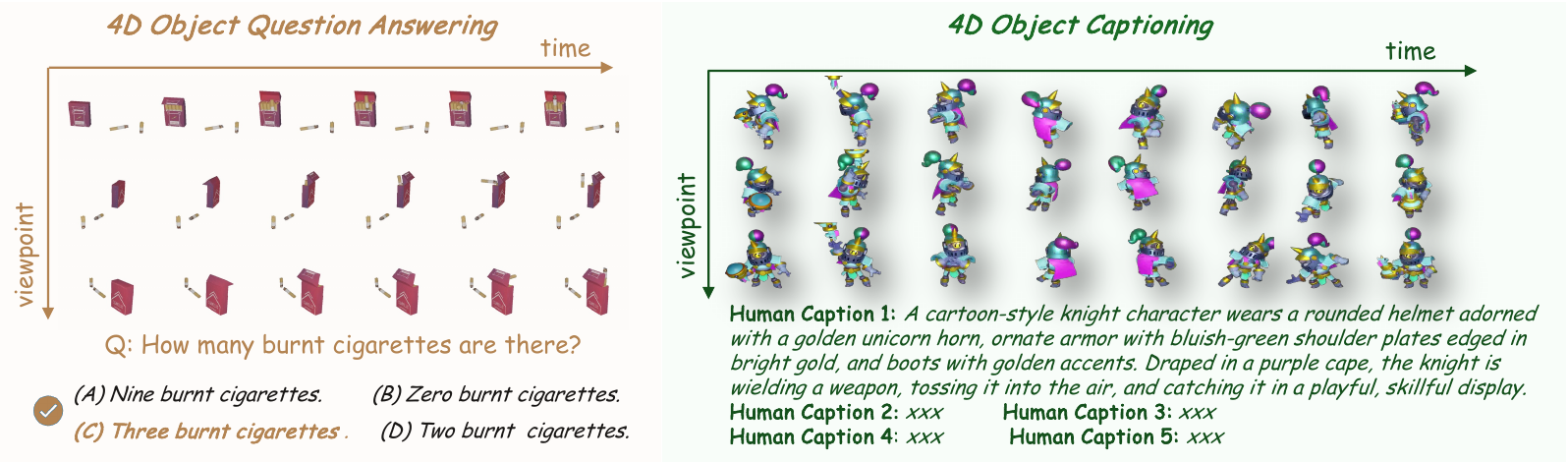}
\caption{\textbf{Illustration of the 4D-Bench.} 4D-Bench consists of two critical tasks (a) 4D object QA and (b) 4D object captioning. 4D object QA provides one question and four choices per QA to evaluate MLLMs. 4D object captioning provides five human captions per 4D object.}
    \label{fig:benchmarks}
    \vspace{-0pt}
\end{figure*}
\section{Related Work}   
% {\color{red} internVL2,LLaVA-onevision ,Qwen2}
\mysection{Multimodal Large Language Models (MLLMs)} 
Large Language Models (LLMs) such as GPT-4o\cite{gpt4}, LLaMA \cite{llama,llama2}, and Gemini \cite{gemini2023} have demonstrated substantial capabilities in language comprehension, generation, and knowledge retrieval. Concurrently, vision-language models like CLIP \cite{radford2021learning} have successfully aligned visual and textual modalities. To understand information across multiple modalities, MLLMs \cite{minigptv2,minigpt,blip2,alayrac2022flamingo,LLaVA,internvl,qwen2_vl} extend the capabilities of LLMs to modalities such as 2D images, videos, and audio by introducing alignment modules and visual instruction tuning.
Models like MiniGPT-4 \cite{minigptv2,minigpt} and LLaVA \cite{LLaVA-15,LLaVA,llavaonevision,llava_video} use multilayer perceptrons (MLPs) to align features extracted by pre-trained vision backbones to the latent space of LLMs, while 2D-Video LLMs such as VideoChat \cite{videochat} and Video-LLaMA \cite{videollama} employ Q-former modules for 2D video understanding. In the realm of 3D vision-language tasks, models like 3D-LLM \cite{3dllm}, 3DVista \cite{3dvista}, and GPT4Point \cite{GPT4Point} have been proposed.

Recent works like InstructBLIP \cite{instructblip}, ShareGPT4V \cite{sharegpt4v}, and ShareGPT4Video \cite{sharegpt4video} leverage GPT-4 Vision to generate large-scale, highly descriptive image-text and video-text datasets, improving captioning capabilities. 
VImageBind-LLM \cite{imagebind-llm} extends multimodal understanding by aligning embeddings from various modalities, including audio and 3D point clouds, to LLMs using a learnable binding network. 
Our findings highlight significant room for improvement in fine-grained temporal understanding within 4D object comprehension, underscoring the need for systematic evaluation and further research to address these challenges.

\mysection{Evaluations of MLLMs} To evaluate image and video tasks in MLLMs, a range of benchmarks has emerged~\cite{MMVP,cambrian1,mmbench,liu2023hidden,mmmu,Mm-vet,POPE2023,videomme,autobench}.
Initial efforts~\cite{LLaVA-Bench,OwlEval} provided foundational assessments but lacked scale, leading to benchmarks that assess perception and cognition across diverse subtasks~\cite{MME}. Liu et al.
~\cite{mmbench} leveraged GPT-4~\cite{gpt4} for scalable, labor-free evaluations.
More recent developments like SEED-Bench and SEED-Bench-2~\cite{seedbench,li2024seed} introduced six-fold larger annotations with extensive multi-modal questions, categorizing MLLM capabilities into hierarchical levels.
Image understanding benchmarks evolved from object counting~\cite{GVT-bench} to high-resolution detail assessments~\cite{MagnifierBench,HR-Bench}.
Fine-grained image-text alignment and relational understanding are evaluated through complex semantic matching~\cite{Winoground,VALSE} and paired image relationships~\cite{compbench}.
For further details on these benchmarks, please refer to~\cite{li2024survey}.

Video understanding benchmarks~\cite{mvbench,tempcompass,autoevalvideo,moviechat,mmbenchvideo,egoschema,videomme,MLVU} focus on temporal coherence and action recognition, progressing from early tasks~\cite{TimeIT} to more granular temporal and causal assessments~\cite{mvbench,VilMA,tempcompass,OsCaR}.
Real-world activities with overlapping actions are assessed in~\cite{LLAVIDAL}, while comprehensive video evaluations encompass diverse tasks and long-form content~\cite{MLVU,Event-Bench,Video-Bench,autoevalvideo}.
In addition to MLLMs, T3bench \cite{T3bench} introduces a benchmark to evaluate text-to-3D generation methods.
Different from these benchmarks, our benchmark focuses on evaluating the capability of MLLMs on 4D-object-centric understanding.

\section{A New Benchmark: 4D-Bench}
\label{sec:method}

We establish a new benchmark named 4D-Bench to evaluate MLLMs on 4D object understanding.
We define the 4D object question answering task in Sec. \ref{sec:qa_task_definition} and the 4D object captioning task in Sec. \ref{sec:caption_task_definition}.
We then describe the data collection and the annotations of these two tasks in Sec. \ref{sec:data_collection}.

\subsection{Task 1: 4D Object Question Answering}
\label{sec:qa_task_definition}
 
We propose the following five subtasks of 4D object QA to evaluate MLLMs' 4D object understanding capability. While some subtask definitions may be similar to those in 2D video benchmarks, the complexity of 4D objects introduces new challenges for MLLMs.

\textit{\textbf{Appearance.}} This subtask evaluates MLLMs to analyze and describe the visual attributes of objects. This subtask presents two key challenges: (1) many objects in our dataset are synthetic or fictional, presenting attributes and configurations that may deviate significantly from real-world examples that MLLMs were trained on, and (2) the multi-view nature requires MLLMs to integrate appearance information across different viewpoints (e.g., ``From the front view, what color is the main part of the character's outfit? From the side view, does the character appear to have any accessories attached to their back?").

\textit{\textbf{Action.}}   Different from 2D video-based benchmarks that focus on scene-level videos, our benchmark enables the deep study of the activities of an object and the motions of its local parts from multiple viewpoints. 
The action subtask evaluates MLLMs in three additional aspects:  (1) typical action recognition; 
 (2) fine-grained motion detection that recognizes subtle movements of specific parts; (3) directional movement analysis that determines specific movement directions.

\textit{\textbf{Object Counting.}} This evaluation subtask evaluates MLLMs by performing precise object enumeration under dynamic and spatially complex scenarios. The key challenges lie in two aspects: (1) temporal dynamics where objects may appear or disappear during the sequence, requiring continuous tracking and count adjustment, and (2) occlusion handling where objects may be partially or fully obscured from certain viewpoints, necessitating cross-view information integration to arrive at accurate counts.
% These scenarios test MLLMs' ability to maintain consistent object tracking across temporal transitions and synthesize spatial information from multiple perspectives.

\textit{\textbf{Spatial Relationship.}} This subtask tests MLLMs' ability to understand spatial configurations across multiple viewpoints, requiring them to analyze object relationships and transformations while integrating information from different angles to handle occlusions.

\textit{\textbf{Temporal Relationship.}} This subtask examines MLLMs' ability to comprehend the temporal evolution of objects or sequential actions.

\subsection{Task 2: 4D Object Captioning}
\label{sec:caption_task_definition}
The 4D object captioning task is to generate text descriptions for the 4D objects. Here, our task requires MLLM to interpret and describe the objects' appearance and actions.
Unlike 2D image/video captioning~\cite{dong2024benchmarking, chen2015microsoft, xu2016msr, chen:acl11, krishna2017dense, agrawal2019nocaps}, 4D object captioning necessitates multi-view spatial-temporal understanding in two aspects:
(1) appearance description requires aggregating visual details observed from different angles to form a complete understanding of the object's characteristics, and (2) action description demands observing the motion sequence from various perspectives to accurately capture complex movements that may be ambiguous or partially visible from a single viewpoint.

\subsection{Data Collection and Annotation} 
\label{sec:data_collection}
\begin{figure}[tb]
    \centering
\includegraphics[width=1.0\linewidth]{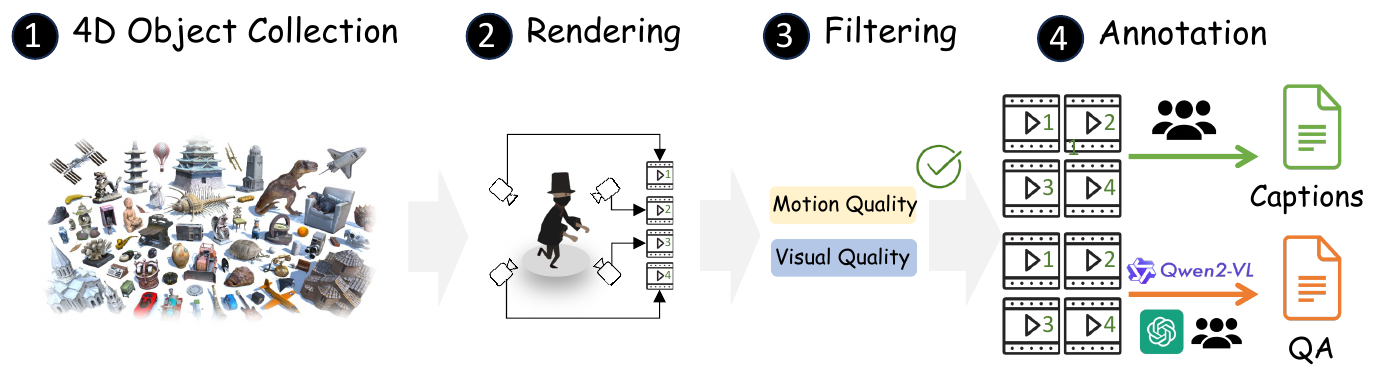} 
\caption{\textbf{Pipeline for constructing the 4D-Bench dataset.} The pipeline includes rendering multi-view videos for 4D objects from Objaverse-XL, motion filtering, visual quality filtering, and multi-stage annotations for QA pairs and captions. Captions are purely human-annotated, while QA pairs are generated through a hybrid approach using MLLMs and human validation.}
\label{fig:dataset_pipeline}
\end{figure}

In this section, we describe the construction of our 4D-Bench dataset shown in \cref{fig:dataset_pipeline}.

\subsubsection{4D Data Collection and Curation.} 
% \textbf{4D Data collection, curation and presentation.} 
We choose multi-view videos as the representation for 4D objects to make the benchmarking of MLLMs possible.
To build our dataset, we render tens of thousands of dynamic 3D objects collected from Objaverse-XL \cite{objaversexl}.
Due to the noisy nature of the data, we designed a data-cleaning pipeline to filter out low-quality samples.
The data-cleaning process consists of two main stages.

\mysection{Object motion analysis} We perform pixel change detection of the rendered videos to identify the temporal boundaries of object motion, allowing us to extract relevant video segments. This ensures the dataset contains exclusively dynamic objects.

\mysection{Object visual quality assessment} Many 4D objects exhibit undesirable visual characteristics, such as oversimplified geometry, lack of texture, and poor aesthetic quality.
Here, we propose a CLIP-based\cite{radford2021learning} filtration framework. We manually annotated thousands of images as high or low quality, then we fine-tuned the CLIP image encoder to serve as a quality classifier to distinguish between high and low-quality objects. The resulting classifier effectively filters out low-quality objects, ensuring that only visually appealing and geometrically complex objects are included.

\subsubsection{4D Object Question Answering Annotation.} 
Designing challenging 4D object question-answer pairs necessitating both multi-view spatial and temporal understanding is challenging, given that our multi-view videos feature only a single object and cover a short time span.
% Designing challenging 4D object question-answer pairs is non-trivial.

We began by engaging professional (have done similar tasks before) annotators who were instructed to carefully observe the rendered multi-view videos and design challenging questions with four choices. Each annotation was subsequently manually verified by us. However, this process proved to be not only costly but also suffered from quality degradation over time. Specifically, the retention rate of annotations from the annotation team initially stood at 92.0\% but dramatically declined to 62.5\% in later stages. During this preliminary exploration phase, we retained 164 high-quality QA pairs that met our rigorous standards.

Inspired by recent work \cite{mvbench,egoschema,hourvideo}, we leveraged MLLMs, specifically GPT-4o and Qwen2-VL, to generate QA pairs from tens of thousands multi-view videos of 4D objects. By prompting the model to analyze multi-view videos through chain-of-thought reasoning, we facilitated the generation of challenging questions and options.
The generated QA pairs underwent an initial validation process using the Qwen2-VL 7B model to ensure strict adherence to the predefined task-specific guidelines and quality criteria.
Then we run blind filtering by inputting only the QA text content (without visual input) to Qwen2.5\cite{qwen25} and Llama 3.1\cite{llama31} and drop those where both models answer correctly. Finally, we performed a manual review to refine the remaining pairs and removed any inappropriate 4D object question-answering pairs.

 \subsubsection{4D Object Captioning Annotation.} 
We manually examined approximately 8,000 candidate 4D objects and carefully selected 580 representative samples, prioritizing diversity in object types and motion characteristics (see \cref{fig:dataset_sta} for 4D object category distribution). For each object, five professional annotators independently provided one caption based on the multi-view video, resulting in five unique descriptions per 4D object. A dedicated reviewer ensured that captions captured significant details and exhibited diversity, unsatisfactory captions were revised accordingly.

 \subsection{Statistics of 4D-Bench.}
The statistics of 4D-Bench are shown in \cref{fig:dataset_sta}, we provide more details in the Appendix.

Our 4D Object QA task contains 751 question-answer pairs for 736 4D objects, where the \textit{Action} subtask comprises the largest portion of the question-answer pairs.
The remaining four subtasks (Appearance, Object Counting, Spatial Relationship, and Temporal Relationship) are distributed in relatively balanced proportions.
4D object captioning task of 4D-Bench covers 580 4D objects with diverse categories.

\begin{figure}[tb]
    \centering
    \begin{minipage}{0.48\linewidth}
        \centering
        \includegraphics[width=\linewidth]{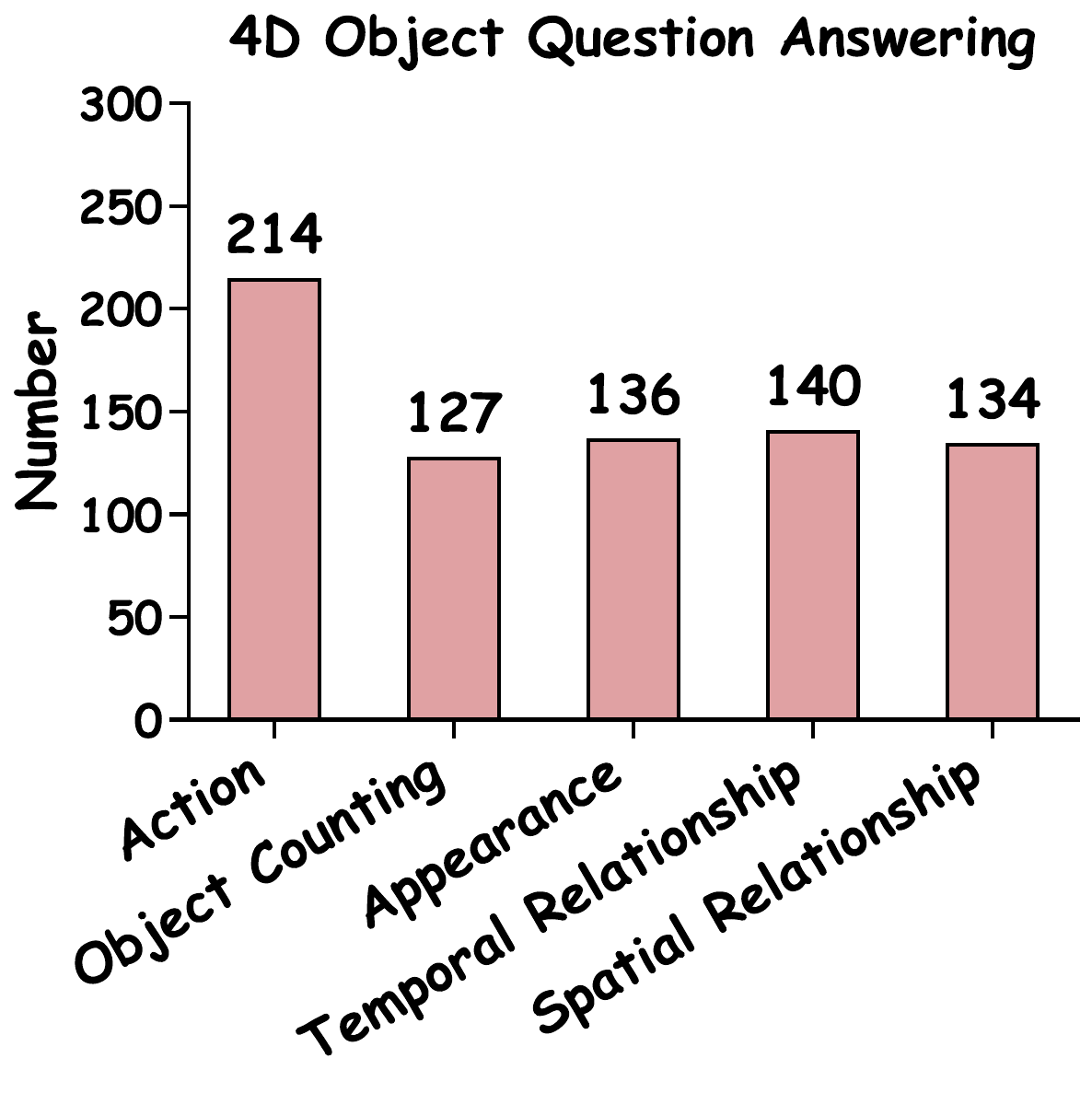}
    \end{minipage}
    \hspace{0.1cm} % 控制两张图像之间的间距
    \begin{minipage}{0.48\linewidth}
        \centering
        \includegraphics[width=\linewidth]{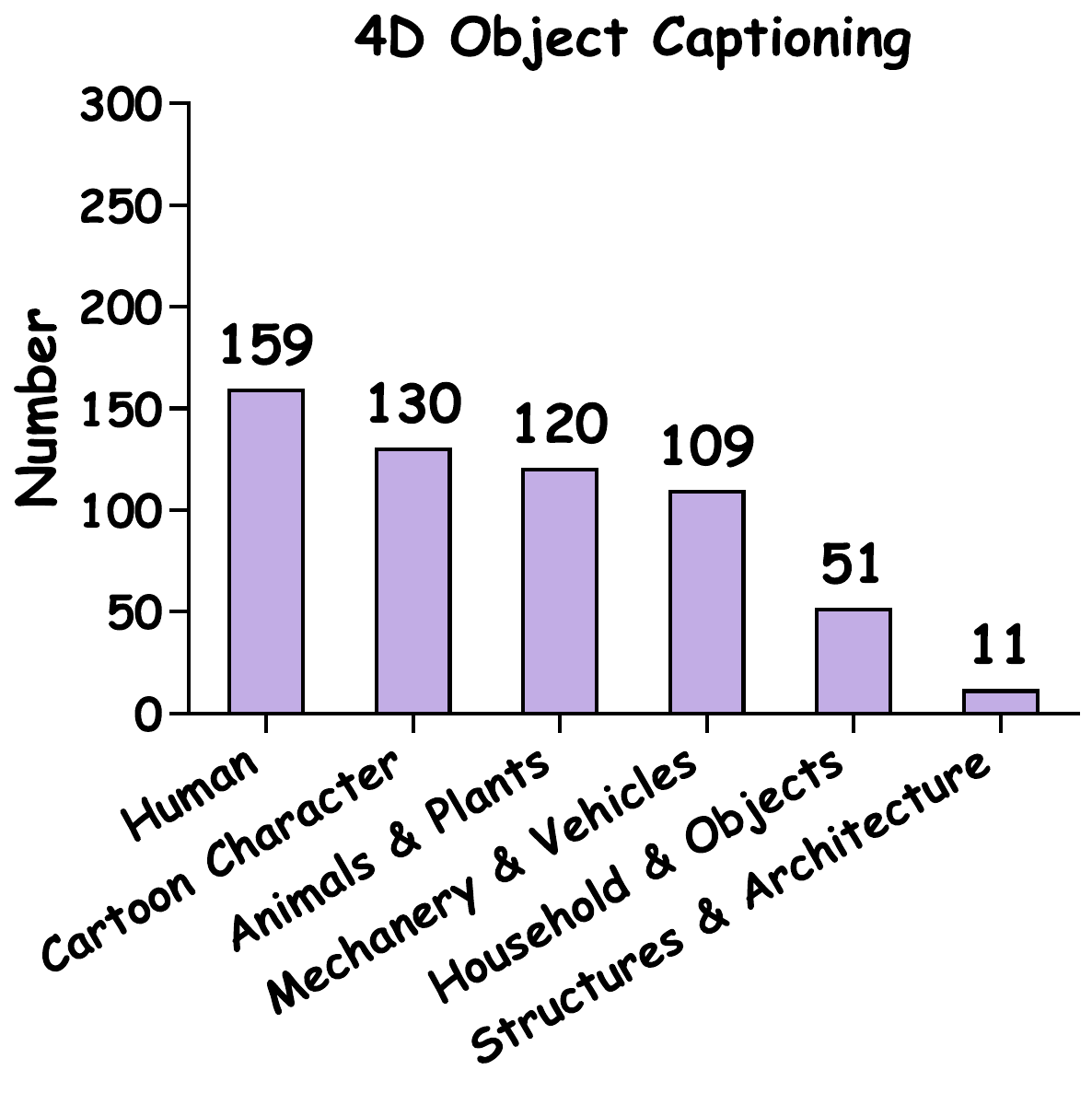}
    \end{minipage}
    \caption{\textbf{Subtask and category distributions in 4D object QA and captioning.} Left: Distribution of five subtasks in the 4D object QA task, 751 question-answering pairs in total. Right: Distribution of 4D object categories in 4D object captioning task, 580 4D objects in total.}
    \label{fig:dataset_sta}
\end{figure}

\section{Experiments}
\subsection{Evaluation Metrics}
\mysection{4D object question answering metrics} 
The 4D object QA consists of questions with four choices where only one choice is correct.
We report both task-specific accuracies and the aggregate performance across the entire benchmark dataset.

\mysection{4D object captioning metrics}
To evaluate the generated captions against the five human annotations provided for each 4D object, we employ a comprehensive evaluation framework.
This includes traditional n-gram-based metrics such as BLEU~\cite{bleu}, ROUGE~\cite{rouge}, METEOR~\cite{meteor}, and CIDEr~\cite{cider}, which remain standard in the caption evaluation literature despite some noted limitations.
We also incorporate embedding-based metrics like BERTScore~\cite{bertscore} and Sentence-BERT~\cite{sentencebert}.

Furthermore, inspired by recent findings~\cite{dong2024benchmarking, VideoChatGPT, Neptune, moviechat} that have widely validated and adopted LLM-based evaluation for its stronger correlation with human judgment~\cite{dong2024benchmarking, Neptune}, we introduce GPT-4o as our LLM evaluator.
The GPT-Appearance and GPT-Action scores evaluate the similarity between the predicted and human-annotated captions in terms of object appearance and actions, respectively.
Both scores range from 0 to 5, and the GPT-Eval score is the average of these two scores. For more information about GPT evaluation, please refer to the Appendix.

\subsection{Evaluation Settings}
We evaluate a range of advanced MLLMs, including two leading closed-source models, GPT-4o \cite{gpt4} and Gemini 1.5 Pro~\cite{gemini2024}, as well as widely used open-source models:  MiniGPT4-Video \cite{minigpt4-video}, VideoChat2~\cite{mvbench}, InternVL2 \cite{internvl}, Qwen2-VL \cite{qwen2_vl}, LLaVA-OneVision \cite{llavaonevision} and LLaVA-Video \cite{llava_video}.

We uniformly select $K$ views around the 4D object from the rendered multi-view videos, then sample $N$ frames from each selected view's video sequence, resulting in a  $K \times N$ frames input.
In our experiments, we empirically set $K=3$ and $N=6$. Such sampling strategies ensure that the evaluations fulfill GPU memory constraints while covering the multi-view and temporal information of 4D objects well.

\subsection{Evaluation Results on 4D Object QA}

4D object question answering experimental results are showed in \cref{tab:vqa_main_exp}. Here, we provide our key findings.

\textit{\textbf{MLLMs underperform humans.}} Our experimental results demonstrate a clear performance hierarchy, with GPT-4o achieving the highest \textit{Overall} accuracy (62.98\%).
However, it should be noted that even the best-performing model achieves relatively modest accuracy.
This is particularly striking given that our test cases primarily involve simple 4D objects - when presented with carefully designed questions requiring multi-view spatial and temporal understanding, current MLLMs struggle to provide accurate responses.

\begin{figure}[tb]
    \centering
    \includegraphics[width=\columnwidth]{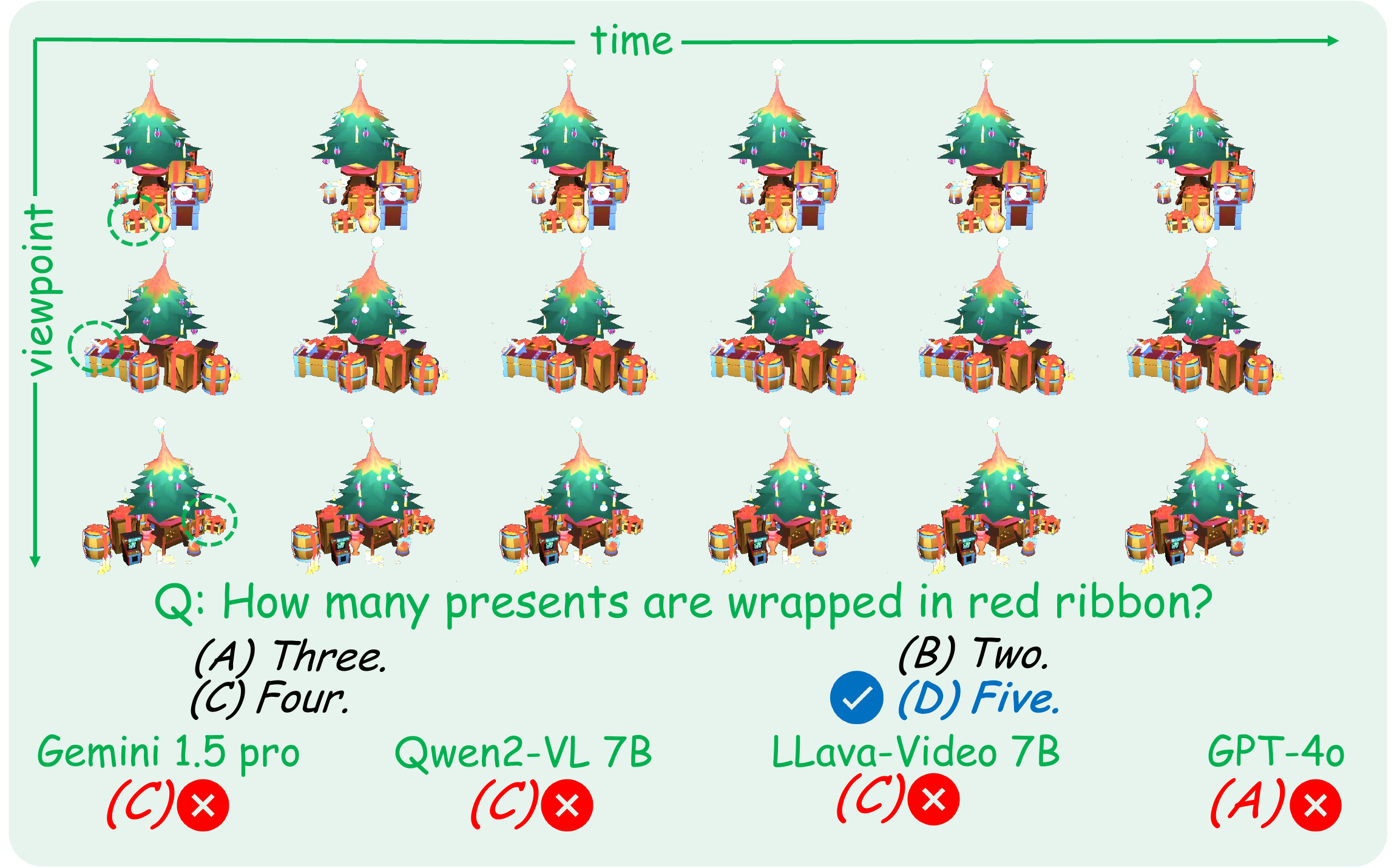}
\caption{\textbf{An example from \textit{Object Counting} subtask.} Answering this question requires integrating multi-view information and capturing cross-view correspondences to count the presents, necessitating multi-view reasoning abilities.  If relying solely on a single view (\eg the middle row), it would lead to wrong answers (\eg four), since some boxes are occluded and invisible in this view.}
    \label{fig:vqa_example_tree}
\end{figure}

\textbf{\textit{MLLMs struggle most with Object Counting task.}} A large performance gap between object counting and other subtasks.  All models struggle in \textit{Object Counting} (37.29\% average accuracy), in contrast, even for the challenging subtask \textit{Temporal Relationship} understanding, models achieve higher performance (49.29\% average accuracy). 
\cref{fig:vqa_example_tree} shows the performance of MLLMs on a counting problem. Although the absence of motion information lowers the complexity of answering the question, Gemini 1.5 pro, Qwen2-VL 7B, LLava-Video 7B and GPT-4o still wrongly answer the question. Such results uncover the limitations of these advanced MLLMs in fusing information from different views to reason accurate counts.

\textbf{\textit{MLLMs are better at appearance and spatial understanding than action and temporal understanding.}}
This pattern is also validated in the following 4D object captioning experimental results. 
As shown in \cref{tab:vqa_main_exp}, many MLLMs achieve over 70\% accuracy in the \textit{Appearance} subtask.
In the subtask of \textit{Spatial Relation}, half of the MLLMs achieve over 60\% accuracy.
% whereas the performances of open-source models \eg LLaVA-Video 72B and Qwen2-VL 72B, are comparable to closed-source models.
However, all MLLMs perform worse in subtasks of \textit{Temporal Relationship} and \textit{Action}, with average accuracies of only 49.29\% and 49.37\%, respectively.

The above evaluation results highlight the new challenges posed by 4D object understanding and showcase the shortcomings of MLLMs in detailed aspects. 
On the other hand, the revealed shortcomings provide valuable guidance for future improvements. 
% For example, poor object counting accuracy indicates that MLLMs' 4D object understanding performance can be improved by explicitly capturing multi-view and temporal correlations.
For example, weak performance in action understanding suggests more advanced temporal-aware visual encoders can enhance MLLMs' performance.
\begin{table*}[ht]
\centering
\small
\resizebox{\textwidth}{!}{%
\begin{tabular}{lcccccc}
\toprule
\textbf{Model} & \textbf{Object Counting (\%)}  & \textbf{Temporal Relationship (\%)} & \textbf{Action (\%)} & \textbf{Spatial Relationship (\%)} & \textbf{Appearance (\%)} & \textbf{Overall (\%)} \\
\midrule
MiniGPT4-Video \cite{minigpt4-video} & 22.05 & 26.43 & 22.90 & 22.39 & 22.06 & 23.17 \\
VideoChat2 \cite{videochat} & 22.83 & 31.43 & 33.18 & 38.81 & 34.56 & 32.36 \\
InternVL2 8B \cite{internvl} & 18.11 & 31.43 & 35.98 & 32.09 & 39.71 & 32.09 \\
LLaVA-OneVision 7B \cite{llavaonevision} & 42.52 & 52.86 & 42.99 & 57.46 & 74.26 & 53.00 \\
LLaVA-Video 7B \cite{llava_video} & 42.52 & 55.00 & 52.80 & 56.72 & \textbf{78.68} & 56.86 \\
Qwen2-VL 7B \cite{qwen2_vl} & 38.58 & 56.43 & 57.94 & 58.96 & 71.32 & 56.99 \\
InternVL2 76B \cite{internvl} & 28.35 & 45.00 & 42.52 & 38.81 & 64.71 & 43.94 \\
LLaVA-OneVision 72B \cite{llavaonevision} & 49.61 & 58.57 & 60.75 & 61.19 & 76.47 & 61.38 \\
LLaVA-Video 72B \cite{llava_video} & \textbf{54.33 }& 58.57 & 57.48 & 66.42 & 77.21 & 62.32 \\
Qwen2-VL 72B \cite{qwen2_vl} & 45.67 & 55.71 & 58.41 & 61.19 & 72.06 & 58.72 \\
Gemini 1.5 Flash \cite{gemini2024} & 26.77 & 50.00 & 53.27 & 60.45 & 66.18 & 51.80 \\
GPT-4o mini \cite{gpt4} & 40.16 & 50.71 & 50.00 & 61.94 & 72.06 & 54.59 \\
Gemini 1.5 Pro \cite{gemini2024} & 46.46 & 58.57 & 59.35 & 64.18 & 68.38 & 59.52 \\
GPT-4o \cite{gpt4} & 44.09 & \textbf{59.29} & \textbf{63.55} & \textbf{69.40} & 77.21 & \textbf{62.98} \\
\midrule
Average & 37.29 & 49.29 & 49.37 & 53.57 & 63.92 & 50.69 \\
\rowcolor{gray!20} Human & 88.98 & 89.29 & 94.39 & 91.04 & 89.71 & 91.08 \\
\bottomrule
\end{tabular}%
}
\caption{\textbf{4D object question answering results.} The Overall column refers to average accuracy across all sub-tasks. The Average row represents the mean performance of all tested models in each category. We provide human performance as a reference.}
\label{tab:vqa_main_exp}
\end{table*}

\subsection{Evaluation Results on 4D Object Captioning}
\begin{table*}[htb]
\centering
\small
\resizebox{\textwidth}{!}{%
\begin{tabular}{l*{11}{c}}
\toprule
\textbf{Model} & \textbf{CIDEr} & \textbf{BLEU@4} & \textbf{METEOR} & \textbf{ROUGE} & \textbf{BERT} & \textbf{SBERT} & \textcolor{green}{\textbf{GPT-Appearance}} & \textcolor{green}{\textbf{GPT-Action}} & \textcolor{green}{\textbf{GPT-Eval}} \\
\midrule
MiniGPT4-Video \cite{minigpt4-video} & 18.4 & 0.6 & 23.1 & 13.2 & 50.7 & 51.2 & 1.737/5 & 1.351/5 & 1.544/5 \\
InternVL2 8B \cite{internvl} & 48.4 & 2.5 & 27.9 & 22.6 & 58.2 & 60.3 & 2.531/5 & 1.877/5 & 2.204/5 \\
VideoChat2-Mistral \cite{mvbench} & 79.0 & 6.9 & 33.5 & 33.5 & 65.4 & 59.7 & 2.578/5 & 1.912/5 & 2.245/5 \\
LLaVA-OneVison 7B \cite{llavaonevision} & 86.4 & 10.0 & 39.2 & 32.7 & 63.2 & 65.6 & 3.166/5 & 2.479/5 & 2.823/5 \\
LLaVA-Video 7B \cite{llava_video} & 102.6 & 14.6 & \textbf{41.7} & 38.8 & 66.7 & 68.1 & 3.235/5 & 2.552/5 & 2.894/5 \\
Qwen2-VL 7B \cite{qwen2_vl} & 84.5 & 10.1 & 36.9 & 36.4 & 65.7 & 66.9 & 3.170/5 & 2.666/5 & 2.918/5 \\
InternVL2 76B \cite{internvl} & 72.0 & 5.5 & 34.2 & 27.1 & 60.9 & 65.3 & 3.099/5 & 2.637/5 & 2.868/5 \\
LLaVA-OneVision 72B \cite{llavaonevision} & \textbf{107.4} & \textbf{16.1} & 41.1 & \textbf{41.5} & \textbf{68.5} & 68.0 & 3.180/5 & 2.268/5 & 2.724/5 \\
LLaVA-Video 72B \cite{llava_video} & 106.2 & 15.1 & 39.8 & 40.9 & \textbf{68.5} & 68.1 & 3.138/5 & 2.471/5 & 2.804/5 \\
Qwen2-VL 72B \cite{qwen2_vl} & 95.1 & 12.4 & 40.3 & 38.0 & 66.8 & 67.5 & 3.324/5 & 2.791/5 & 3.057/5 \\
Gemini 1.5 Flash \cite{gemini2024} & 84.3 & 7.3 & 36.5 & 32.9 & 65.3 & \textbf{68.9} & 3.246/5 & 2.931/5 & 3.088/5 \\
GPT-4o mini \cite{gpt4} & 51.1 & 2.7 & 30.8 & 24.0 & 59.3 & 63.5 & \textcolor{gray}{3.311/5} & \textcolor{gray}{3.131/5} & \textcolor{gray}{3.221/5} \\
Gemini 1.5 Pro \cite{gemini2024} & 94.8 & 11.2 & 38.7 & 39.0 & \textbf{68.5} & 68.8 & 3.311/5 & 2.983/5 & 3.147/5 \\
GPT-4o \cite{gpt4} & 69.0 & 6.4 & 35.9 & 32.1 & 64.1 & 66.4 & \textbf{\textcolor{gray}{3.507/5}} & \textbf{\textcolor{gray}{3.258/5}} & \textbf{\textcolor{gray}{3.382/5}} \\
\midrule
Average & - & - & - & - & - & - & 3.038/5 & 2.522/5 & 2.780/5 \\
\rowcolor{gray!20} Human & 126.6 & 14.12 & 45.01 & 43.48 & 71.69 & 76.30 & 3.772/5 & 3.879/5 & 3.826/5 \\
\bottomrule
\end{tabular}%
}
\caption{\textbf{4D object captioning results.} The Average row represents the mean performance of all tested MLLM models under each metric. The Human row represents the performance of human annotator under each metric. For each metric, we \textbf{bold} the best performing MLLM model. We \textcolor{green}{highlight} GPT metrics as they demonstrate better alignment with human preferences in evaluating caption quality, and our analysis also primarily focuses on models' performance across these metrics. GPT-4o's GPT metrics are marked in \textcolor{gray}{gray} due to the potential self-evaluation bias when using GPT-based metrics to evaluate a GPT model\cite{panickssery2404llm}. We provide human performance as a reference.}
\vspace{-4pt}
\label{tab:caption_main_exp}
\end{table*}

\cref{tab:caption_main_exp} illustrates the evaluation results of various MLLMs on the 4D object captioning task of 4D-Bench.
The following analysis primarily relies on GPT-Appearance, GPT-Action, and GPT-Eval scores \cite{dong2024benchmarking, Neptune}. 

\textit{\textbf{MLLMs underperform humans.}} Current state-of-the-art multi-modal large models (MLLMs) still underperform compared to humans. As shown in \cref{tab:caption_main_exp}, humans achieve better scores with a GPT-Eval score of 3.826 out of 5, compared to even the best-performing MLLM, GPT-4o, with a score of 3.382 out of 5.

\textit{\textbf{MLLMs are better at appearance understanding than action understanding.}} A deeper analysis across different evaluation metrics reveals interesting patterns in model capabilities. We observe that both open-source and closed-source models generally achieve higher scores in GPT-Appearance compared to GPT-Action. For instance, Qwen2-VL 72B achieves a GPT-Appearance score of 3.324/5 but drops to 2.791/5 for GPT-Action. 

\textit{\textbf{Open-source models lag behind closed-source models in action understanding.}} All the closed-source models (such as Gemini 1.5 Pro and GPT-4o mini) achieve a higher overall performance in 4D object captioning compared to open-source models, where their GPT-Eval scores are higher than 3 (out of a maximum score of 5). In contrast, among open-source models, only Qwen2-VL 72B achieves the  GPT-Eval score above 3. Notably, in terms of appearance understanding, open-source models demonstrate competitive performance with their closed-source counterparts, with models like LLaVA-Video 7B and Qwen2-VL 72B achieving GPT-Appearance scores (3.235/5 and 3.324/5, respectively) comparable to Gemini 1.5 Pro (3.311/5). However, when it comes to action understanding, there exists a noticeable gap between open-source and closed-source models. Closed-source models like GPT-4o and Gemini 1.5 Pro maintain stronger performance in GPT-Action (3.258/5 and 2.983/5, respectively), while open-source alternatives show relatively weaker capabilities in this aspect, typically scoring below 2.8.

\subsection{Discussions}
\mysection{Impact of view number and sampling frequency}
Here, we study MLLMs' performance by varying the number of views and sampling frequency of video frames that fed into the model independently. 

\begin{figure}[hbt]
    \centering
    \includegraphics[width=\columnwidth]{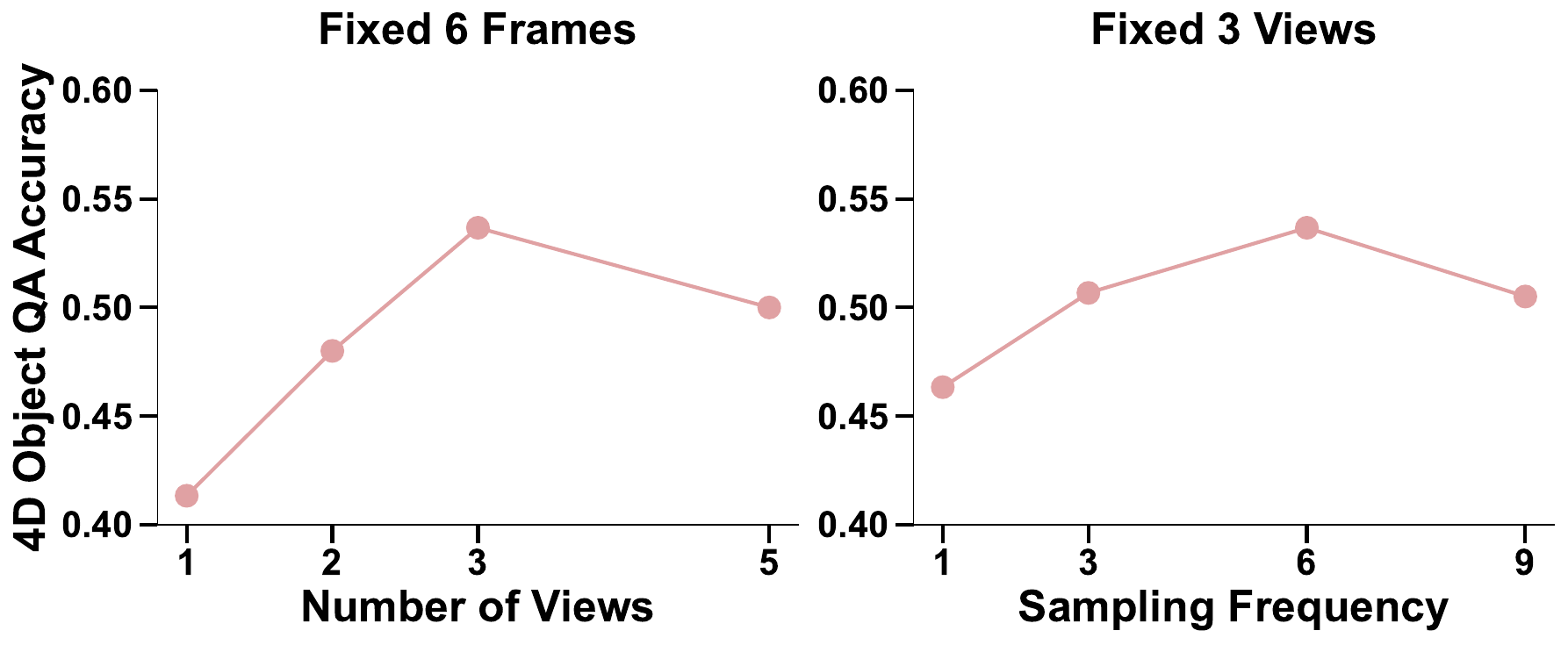}
\caption{\textbf{Effect of view number and temporal sampling on the 4D object QA performance.} Tested on Gemini 1.5 Flash. Left: Accuracies across different numbers of views with fixed 6 frames. Right: Accuracies across different temporal frequencies with fixed 3 views.}
    \label{fig:vqa_ablation}
\end{figure}

For 4D object question answering, \cref{fig:vqa_ablation} shows consistent accuracy improvements with both increased views (41.3\% to 53.7\% with fixed frames) and increased sampling frequencies (46.3\% to 53.7\% with fixed views), confirming that our questions effectively require both multi-view and temporal understanding rather than being solvable from limited viewpoints or timestamps. However, we observed that performance degrades when exceeding 3 views or 6 frames, likely due to information redundancy that may overwhelm the model's processing capacity.

\begin{figure}[htb]
    \centering
    \includegraphics[width=\columnwidth]{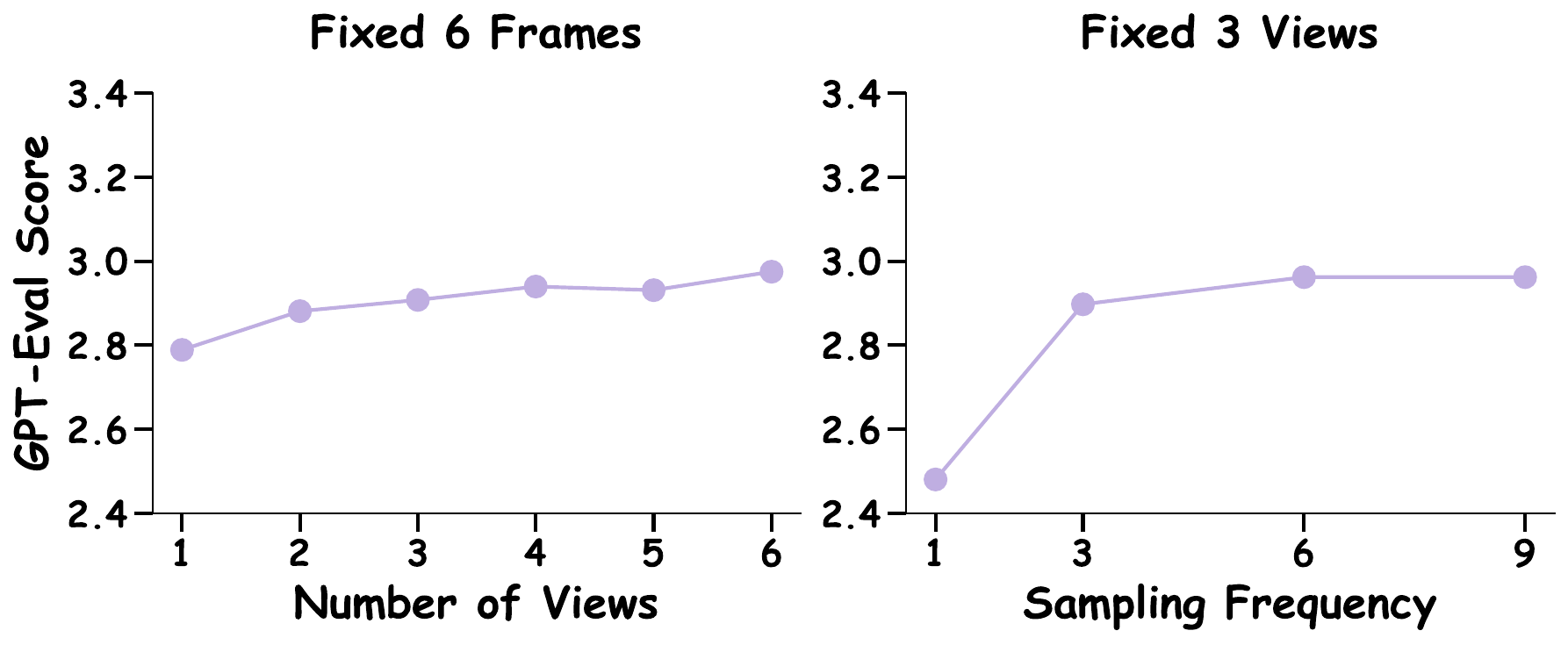}
\caption{\textbf{Effect of view number and temporal sampling on the 4D object captioning performance.} Tested on Qwen2-VL 7B. Left: GPT-Eval scores across different numbers of views with fixed 6 frames. Right: GPT-Eval scores across different temporal frequencies with fixed 3 views.}
    \label{fig:caption_ablation}
\end{figure}

For 4D object captioning, \cref{fig:caption_ablation} shows that increasing the number of views from 1 to 6 improves the GPT-Eval scores from 2.79 to 2.98.  
For temporal sampling, increasing frames from 1 to 3 boosts the GPT-Eval score from 2.48 to 2.89, and a sampling frequency of 6 further improves the GPT-Eval score to 2.96. However, when the sampling frequency is increased from 6 to 9, the performance improvement becomes negligible.

\mysection{Robustness evaluation}
We propose the following two concerns:
(1) In the original experiment design, when inputting images to the large language model, we prioritized viewpoint order (all frames from viewpoint 1, then all frames from viewpoint 2). How would the results differ if we prioritized temporal order instead.
(2) In the original experimental design, we didn't include timestamp information for each image in the prompt (since they were all short videos). What would the results be if we included timestamp information?

To answer those questions, we run corresponding experiments on 4D object question answering and the results are shown in \cref{tab:evalution_robustness}.
The minimal variations in model performance across different input configurations (temporal vs. viewpoint-first ordering and with/without timestamps) demonstrate the robustness of our original experimental design.
\begin{table}[htb]
\centering
\resizebox{0.48\textwidth}{!}{%
\begin{tabular}{lccc}
\toprule
\textbf{Model} & \textbf{Original Setting(\%)} & \textbf{Frame Order(\%)}  & \textbf{w/ Time Stamp(\%)}  \\
\midrule
MiniGPT4-Video \cite{minigpt4-video} & 23.17 & 17.58 \textcolor{red}{(↓5.59)} & 17.18 \textcolor{red}{(↓5.99)} \\
VideoChat2 \cite{mvbench} & 32.36 & 33.95 \textcolor{green}{(↑1.59)} & 23.04 \textcolor{red}{(↓9.32)} \\
InternVL2 8B \cite{internvl} & 32.09 & 38.88 \textcolor{green}{(↑6.79)} & 33.69 \textcolor{green}{(↑1.60)} \\
LLaVA-OneVision 7B \cite{llavaonevision} & 53.00 & 51.40 \textcolor{red}{(↓1.60)} & 53.53 \textcolor{green}{(↑0.53)} \\
LLaVA-Video 7B \cite{llava_video} & 56.86 & 59.25 \textcolor{green}{(↑2.39)} & 57.52 \textcolor{green}{(↑0.66)} \\
Qwen2-VL 7B \cite{qwen2_vl} & 56.99 & 49.80 \textcolor{red}{(↓7.19)} & 57.52 \textcolor{green}{(↑0.53)} \\
InternVL2 76B \cite{internvl} & 43.94 & 47.54 \textcolor{green}{(↑3.60)} & 46.07 \textcolor{green}{(↑2.13)} \\
LLaVA-OneVision 72B \cite{llavaonevision} & 61.38 & 61.25 \textcolor{red}{(↓0.13)} & 60.59 \textcolor{red}{(↓0.79)} \\
LLaVA-Video 72B \cite{llava_video} & 62.32 & 62.72 \textcolor{green}{(↑0.40)} & 61.92 \textcolor{red}{(↓0.40)} \\
Qwen2-VL 72B \cite{qwen2_vl} & 58.72 & 54.46 \textcolor{red}{(↓4.26)} & 59.25 \textcolor{green}{(↑0.53)} \\
Gemini 1.5 Flash \cite{gemini2024} & 51.80 & 51.80 \textcolor{green}{(↑0.00)} & 52.86 \textcolor{green}{(↑1.06)} \\
GPT-4o mini \cite{gpt4} & 54.59 & 53.66 \textcolor{red}{(↓0.93)} & 53.79 \textcolor{red}{(↓0.80)} \\
Gemini 1.5 Pro \cite{gemini2024} & 59.52 & 58.72 \textcolor{red}{(↓0.80)} & 59.25 \textcolor{red}{(↓0.27)} \\
GPT-4o \cite{gpt4} & 62.98 & 60.85 \textcolor{red}{(↓2.13)} & 63.12 \textcolor{green}{(↑0.14)} \\
\midrule
Average & 50.69 & 50.13 \textcolor{red}{(↓0.56)} & 49.95 \textcolor{red}{(↓0.74)} \\
\bottomrule
\end{tabular}%
}
\caption{\textbf{Robustness study of 4D object QA experiment.} Green arrows (\textcolor{green}{↑}) indicate improvement over Original Setting's Overall accuracy, while red arrows (\textcolor{red}{↓}) show decline. }
\label{tab:evalution_robustness}
\end{table}

\mysection{When MLLMs encounter counterfactual 4D data}  
Unlike existing benchmarks based on real-world videos, our dataset is built on artificially created 4D objects and hence provides some counterfactual 4D data that deviates from physical laws and behaves differently from its real-world counterpart.
These data serve as a valuable testbed to examine whether MLLMs truly understand the input or simply rely on learned world knowledge.

\begin{figure}[t]
    \centering
    \includegraphics[width=\columnwidth]{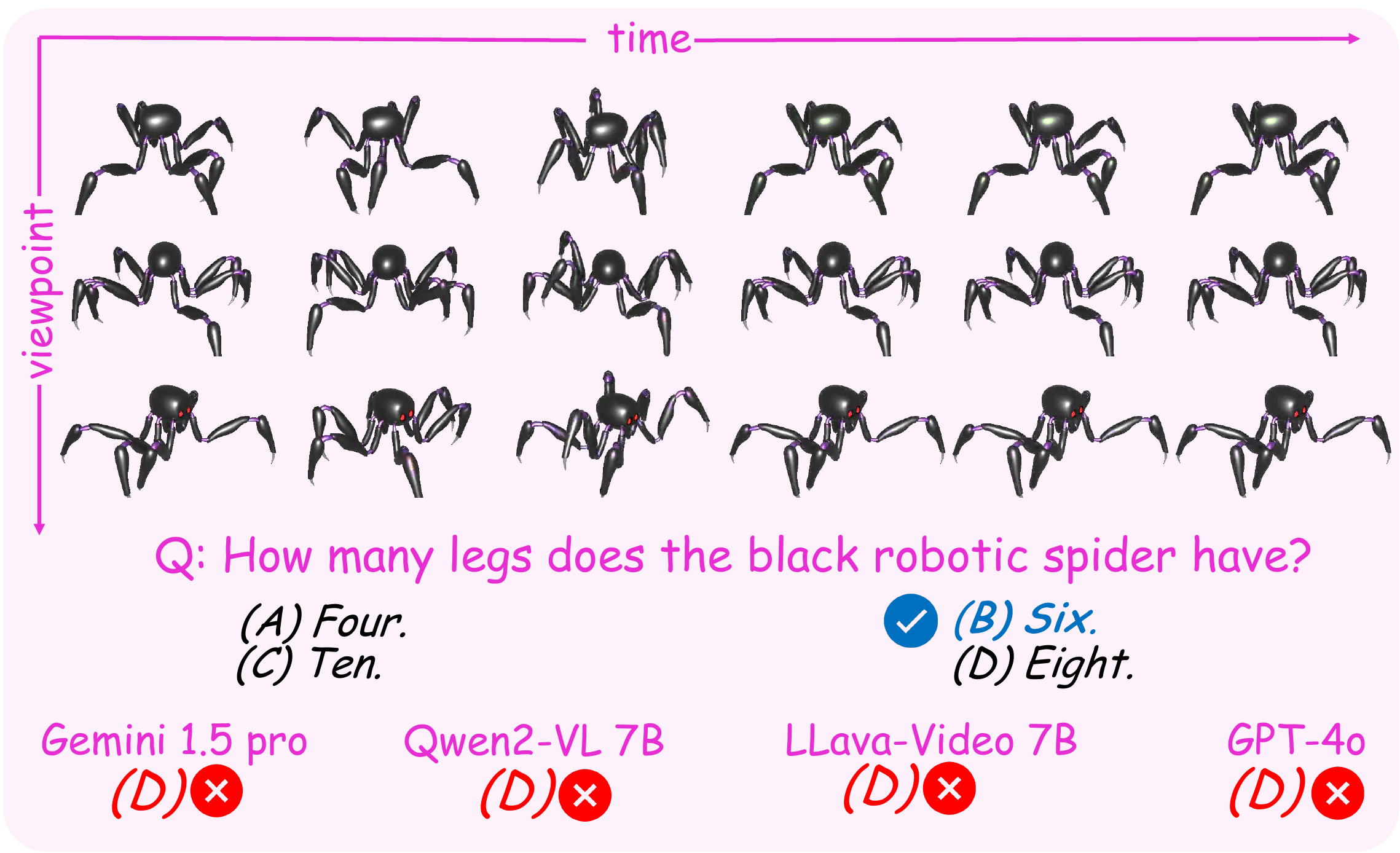}
\caption{\textbf{A counterfactual example from 4D object QA task.} A synthetic spider with six legs, illustrating a counterfactual scenario for testing model understanding, as real spiders typically have eight legs.}
    \label{fig:vqa_example_spider}
% \vspace{-2pt}
\end{figure}

\begin{figure}[t]
    \centering
    \includegraphics[width=\columnwidth]{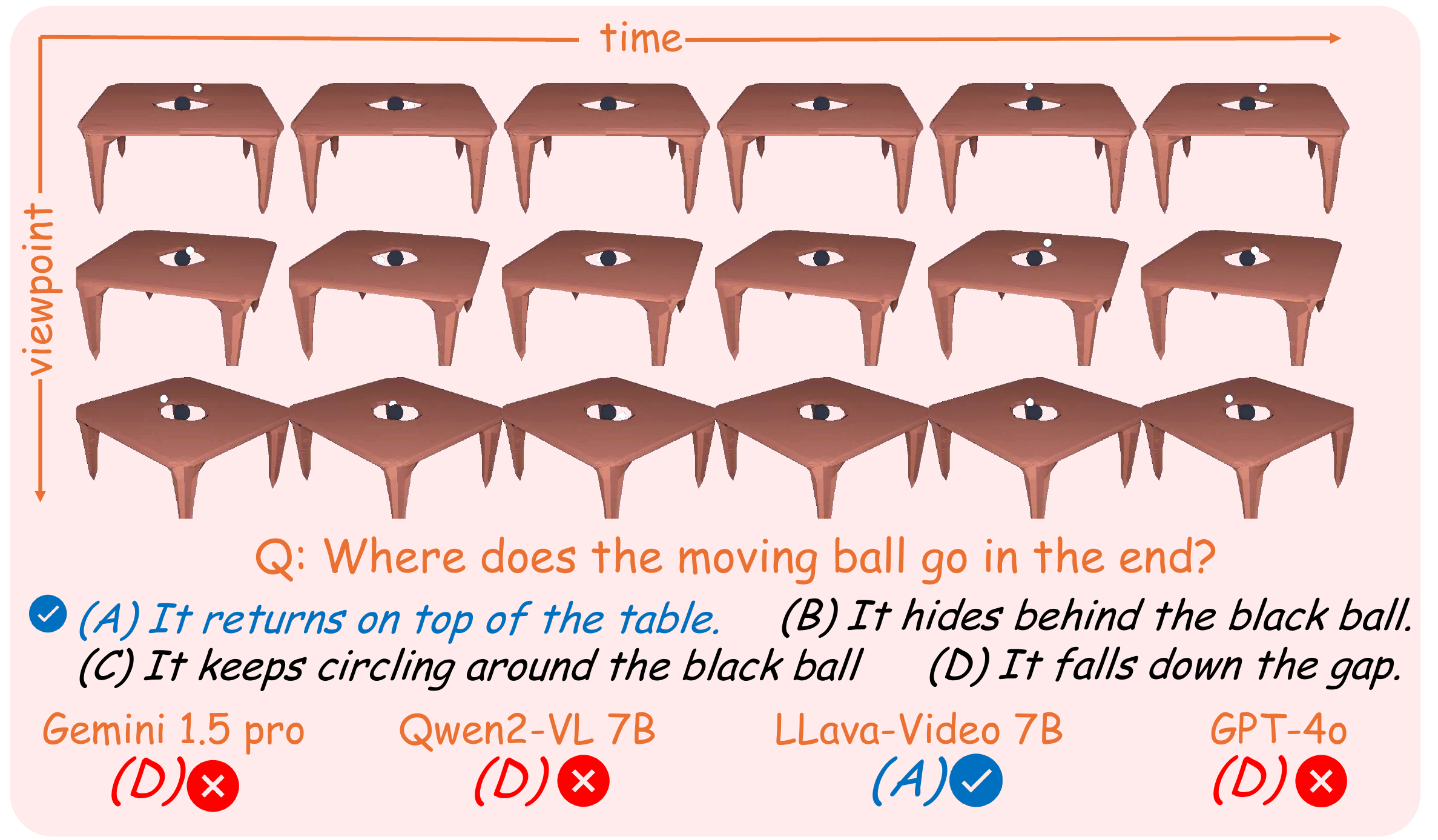}
\caption{\textbf{A counterfactual example from 4D object QA task.} A ball rolling into a downward-facing hole and then rolling back out, depicting a counterfactual scenario that violates physical laws, as a ball would normally stay trapped in the hole.}
    \label{fig:vqa_example_desk}
    % \vspace{-2pt}
\end{figure}

For example, as illustrated in \cref{fig:vqa_example_spider}, our benchmark includes counterfactual testing data where a synthetic spider has 6 legs, contrary to the fact that real spiders typically have 8 legs.
Similarly,  \cref{fig:vqa_example_desk} presents counterfactual testing data where a ball rolls into a downward-facing hole and then rolls back out, defying the laws of physics, as a ball would normally remain trapped in a hole in the real world.
Given these testing data, all advanced MLLMs, including Gemini 1.5 Pro, Qwen2-VL 7B, LLaVA-Video 7B, and GPT-4o, choose the wrong answer. These results highlight that these advanced MLLMs are not robust enough to understand counterfactual data.

\section{Conclusion}

We present 4D-Bench, a novel benchmark for assessing the 4D object understanding capabilities of MLLMs.
Compared with existing benchmarks for 2D image and video understanding, 4D-Bench is 4D-object-centric, providing 4D objects with diverse categories for benchmarking MLLMs.
4D-Bench presents two critical tasks regarding 4D object question answering and 4D object captioning, necessitating multi-view spatial-temporal understanding.
Benchmarking results reveal that the capabilities of existing MLLMs are limited in 4D object understanding. 
We hope that 4D-Bench facilitates the development of MLLMs in 4D object understanding and other related research areas.
For example, our benchmark on 4D object captioning fills in the gap of quantitatively evaluating 4D object captioning performance, which drives research on leveraging MLLMs to generate high-quality text descriptions from 4D objects for improving text-to-4D generative models.
Our benchmark on 4D object question answering enables the community to conduct an in-depth evaluation of the capabilities of MLLMs in specific aspects.

\textbf{Acknowledgement.} This work was supported by funding from King Abdullah University of Science and Technology (KAUST) - Center of Excellence for Generative AI, under award number 5940.

\textbf{License.} 4D-Bench is strictly for academic research purposes, and any form of commercial use is prohibited. The copyright of all 4D objects is retained by their respective owners, and proper acknowledgement will be given in the dataset. The dataset as a whole is licensed under the ODC-By v1.0 license, consistent with the licensing of Objaverse-XL\cite{objaversexl}. 
\clearpage
{
    \small
    \bibliographystyle{ieeenat_fullname}
    \bibliography{main}

\begin{thebibliography}{130}
\providecommand{\natexlab}[1]{#1}
\providecommand{\url}[1]{\texttt{#1}}
\expandafter\ifx\csname urlstyle\endcsname\relax
  \providecommand{\doi}[1]{doi: #1}\else
  \providecommand{\doi}{doi: \begingroup \urlstyle{rm}\Url}\fi

\bibitem[4da(2025{\natexlab{a}})]{4dasset1}
4d technology market size, share, growth report, 2025{\natexlab{a}}.

\bibitem[4da(2025{\natexlab{b}})]{4dasset2}
3d digital asset market share, forecast, 2025{\natexlab{b}}.

\bibitem[Achiam et~al.(2023)Achiam, Adler, Agarwal, Ahmad, Akkaya, Aleman, Almeida, Altenschmidt, Altman, Anadkat, et~al.]{gpt4}
Josh Achiam, Steven Adler, Sandhini Agarwal, Lama Ahmad, Ilge Akkaya, Florencia~Leoni Aleman, Diogo Almeida, Janko Altenschmidt, Sam Altman, Shyamal Anadkat, et~al.
\newblock Gpt-4 technical report.
\newblock \emph{arXiv preprint arXiv:2303.08774}, 2023.

\bibitem[Agrawal et~al.(2019)Agrawal, Desai, Wang, Chen, Jain, Johnson, Batra, Parikh, Lee, and Anderson]{agrawal2019nocaps}
Harsh Agrawal, Karan Desai, Yufei Wang, Xinlei Chen, Rishabh Jain, Mark Johnson, Dhruv Batra, Devi Parikh, Stefan Lee, and Peter Anderson.
\newblock Nocaps: Novel object captioning at scale.
\newblock In \emph{Proceedings of the IEEE/CVF international conference on computer vision}, pages 8948--8957, 2019.

\bibitem[Alayrac et~al.(2022)Alayrac, Donahue, Luc, Miech, Barr, Hasson, Lenc, Mensch, Millican, Reynolds, et~al.]{alayrac2022flamingo}
Jean-Baptiste Alayrac, Jeff Donahue, Pauline Luc, Antoine Miech, Iain Barr, Yana Hasson, Karel Lenc, Arthur Mensch, Katherine Millican, Malcolm Reynolds, et~al.
\newblock Flamingo: a visual language model for few-shot learning.
\newblock \emph{Advances in neural information processing systems}, 35:\penalty0 23716--23736, 2022.

\bibitem[Ataallah et~al.(2024)Ataallah, Shen, Abdelrahman, Sleiman, Zhu, Ding, and Elhoseiny]{minigpt4-video}
Kirolos Ataallah, Xiaoqian Shen, Eslam Abdelrahman, Essam Sleiman, Deyao Zhu, Jian Ding, and Mohamed Elhoseiny.
\newblock Minigpt4-video: Advancing multimodal llms for video understanding with interleaved visual-textual tokens.
\newblock \emph{arXiv preprint arXiv:2404.03413}, 2024.

\bibitem[Awadalla et~al.(2023)Awadalla, Gao, Gardner, Hessel, Hanafy, Zhu, Marathe, Bitton, Gadre, Sagawa, et~al.]{awadalla2023openflamingo}
Anas Awadalla, Irena Gao, Josh Gardner, Jack Hessel, Yusuf Hanafy, Wanrong Zhu, Kalyani Marathe, Yonatan Bitton, Samir Gadre, Shiori Sagawa, et~al.
\newblock Openflamingo: An open-source framework for training large autoregressive vision-language models.
\newblock \emph{arXiv preprint arXiv:2308.01390}, 2023.

\bibitem[Azuma et~al.(2022)Azuma, Miyanishi, Kurita, and Kawanabe]{scanqa}
Daichi Azuma, Taiki Miyanishi, Shuhei Kurita, and Motoaki Kawanabe.
\newblock Scanqa: 3d question answering for spatial scene understanding.
\newblock In \emph{Proceedings of the IEEE/CVF Conference on Computer Vision and Pattern Recognition (CVPR)}, 2022.

\bibitem[Bahmani et~al.(2023)Bahmani, Skorokhodov, Rong, Wetzstein, Guibas, Wonka, Tulyakov, Park, Tagliasacchi, and Lindell]{4dfy}
Sherwin Bahmani, Ivan Skorokhodov, Victor Rong, Gordon Wetzstein, Leonidas Guibas, Peter Wonka, Sergey Tulyakov, Jeong~Joon Park, Andrea Tagliasacchi, and David~B. Lindell.
\newblock 4d-fy: Text-to-4d generation using hybrid score distillation sampling.
\newblock \emph{arXiv preprint arXiv:2311.17984}, 2023.

\bibitem[Banerjee and Lavie(2005)]{meteor}
Satanjeev Banerjee and Alon Lavie.
\newblock Meteor: An automatic metric for mt evaluation with improved correlation with human judgments.
\newblock In \emph{Proceedings of the acl workshop on intrinsic and extrinsic evaluation measures for machine translation and/or summarization}, pages 65--72, 2005.

\bibitem[Bao et~al.(2024)Bao, Huang, Wang, Ye, Wang, Chen, Elhoseiny, and Zhang]{autobench}
Han Bao, Yue Huang, Yanbo Wang, Jiayi Ye, Xiangqi Wang, Xiuying Chen, Mohamed Elhoseiny, and Xiangliang Zhang.
\newblock Autobench-v: Can large vision-language models benchmark themselves?, 2024.

\bibitem[Cao et~al.(2024)Cao, Luo, Zhang, Nießner, and Tang]{motion2vecsets}
Wei Cao, Chang Luo, Biao Zhang, Matthias Nießner, and Jiapeng Tang.
\newblock Motion2vecsets: 4d latent vector set diffusion for non-rigid shape reconstruction and tracking, 2024.

\bibitem[Chakraborty et~al.(2024)Chakraborty, Sinha, Reilly, Govind, Wang, Bremond, and Das]{LLAVIDAL}
Rajatsubhra Chakraborty, Arkaprava Sinha, Dominick Reilly, Manish~Kumar Govind, Pu Wang, Francois Bremond, and Srijan Das.
\newblock Llavidal: Benchmarking large language vision models for daily activities of living.
\newblock \emph{arXiv preprint arXiv:2406.09390}, 2024.

\bibitem[Chandrasegaran et~al.(2024)Chandrasegaran, Gupta, Hadzic, Kota, He, Eyzaguirre, Durante, Li, Wu, and Li]{hourvideo}
Keshigeyan Chandrasegaran, Agrim Gupta, Lea~M. Hadzic, Taran Kota, Jimming He, Cristobal Eyzaguirre, Zane Durante, Manling Li, Jiajun Wu, and Fei-Fei Li.
\newblock Hourvideo: 1-hour video-language understanding.
\newblock In \emph{Advances in Neural Information Processing Systems}, 2024.

\bibitem[Chen and Dolan(2011{\natexlab{a}})]{chen2011collecting}
David Chen and William~B Dolan.
\newblock Collecting highly parallel data for paraphrase evaluation.
\newblock In \emph{Proceedings of the 49th annual meeting of the association for computational linguistics: human language technologies}, pages 190--200, 2011{\natexlab{a}}.

\bibitem[Chen and Dolan(2011{\natexlab{b}})]{chen:acl11}
David~L. Chen and William~B. Dolan.
\newblock Collecting highly parallel data for paraphrase evaluation.
\newblock In \emph{Proceedings of the 49th Annual Meeting of the Association for Computational Linguistics (ACL-2011)}, Portland, OR, 2011{\natexlab{b}}.

\bibitem[Chen et~al.(2023{\natexlab{a}})Chen, Zhu, Shen, Li, Liu, Zhang, Krishnamoorthi, Chandra, Xiong, and Elhoseiny]{minigptv2}
Jun Chen, Deyao Zhu, Xiaoqian Shen, Xiang Li, Zechu Liu, Pengchuan Zhang, Raghuraman Krishnamoorthi, Vikas Chandra, Yunyang Xiong, and Mohamed Elhoseiny.
\newblock Minigpt-v2: large language model as a unified interface for vision-language multi-task learning.
\newblock \emph{arXiv preprint arXiv:2310.09478}, 2023{\natexlab{a}}.

\bibitem[Chen et~al.(2023{\natexlab{b}})Chen, Li, Dong, Zhang, He, Wang, Zhao, and Lin]{sharegpt4v}
Lin Chen, Jisong Li, Xiaoyi Dong, Pan Zhang, Conghui He, Jiaqi Wang, Feng Zhao, and Dahua Lin.
\newblock Sharegpt4v: Improving large multi-modal models with better captions.
\newblock \emph{arXiv preprint arXiv:2311.12793}, 2023{\natexlab{b}}.

\bibitem[Chen et~al.(2024)Chen, Wei, Li, Dong, Zhang, Zang, Chen, Duan, Lin, Tang, et~al.]{sharegpt4video}
Lin Chen, Xilin Wei, Jinsong Li, Xiaoyi Dong, Pan Zhang, Yuhang Zang, Zehui Chen, Haodong Duan, Bin Lin, Zhenyu Tang, et~al.
\newblock Sharegpt4video: Improving video understanding and generation with better captions.
\newblock \emph{arXiv preprint arXiv:2406.04325}, 2024.

\bibitem[Chen et~al.(2015)Chen, Fang, Lin, Vedantam, Gupta, Doll{\'a}r, and Zitnick]{chen2015microsoft}
Xinlei Chen, Hao Fang, Tsung-Yi Lin, Ramakrishna Vedantam, Saurabh Gupta, Piotr Doll{\'a}r, and C~Lawrence Zitnick.
\newblock Microsoft coco captions: Data collection and evaluation server.
\newblock \emph{arXiv preprint arXiv:1504.00325}, 2015.

\bibitem[Chen et~al.(2023{\natexlab{c}})Chen, Lin, Zhang, and Huang]{autoevalvideo}
Xiuyuan Chen, Yuan Lin, Yuchen Zhang, and Weiran Huang.
\newblock Autoeval-video: An automatic benchmark for assessing large vision language models in open-ended video question answering.
\newblock \emph{arXiv preprint arXiv:2311.14906}, 2023{\natexlab{c}}.

\bibitem[Chen et~al.(2023{\natexlab{d}})Chen, Wu, Wang, Su, Chen, Xing, Zhong, Zhang, Zhu, Lu, Li, Luo, Lu, Qiao, and Dai]{internvl}
Zhe Chen, Jiannan Wu, Wenhai Wang, Weijie Su, Guo Chen, Sen Xing, Muyan Zhong, Qinglong Zhang, Xizhou Zhu, Lewei Lu, Bin Li, Ping Luo, Tong Lu, Yu Qiao, and Jifeng Dai.
\newblock Internvl: Scaling up vision foundation models and aligning for generic visual-linguistic tasks.
\newblock \emph{arXiv preprint arXiv:2312.14238}, 2023{\natexlab{d}}.

\bibitem[Cheng et~al.(2024)Cheng, Leng, Zhang, Xin, Li, Chen, Zhu, Zhang, Luo, Zhao, et~al.]{cheng2024videollama}
Zesen Cheng, Sicong Leng, Hang Zhang, Yifei Xin, Xin Li, Guanzheng Chen, Yongxin Zhu, Wenqi Zhang, Ziyang Luo, Deli Zhao, et~al.
\newblock Videollama 2: Advancing spatial-temporal modeling and audio understanding in video-llms.
\newblock \emph{arXiv preprint arXiv:2406.07476}, 2024.

\bibitem[Dai et~al.(2023)Dai, Li, Li, Tiong, Zhao, Wang, Li, Fung, and Hoi]{instructblip}
Wenliang Dai, Junnan Li, Dongxu Li, Anthony Tiong, Junqi Zhao, Weisheng Wang, Boyang Li, Pascale Fung, and Steven Hoi.
\newblock Instruct{BLIP}: Towards general-purpose vision-language models with instruction tuning.
\newblock In \emph{Thirty-seventh Conference on Neural Information Processing Systems}, 2023.

\bibitem[Dai et~al.(2024)Dai, Lee, Wang, Yang, Liu, Barker, Rintamaki, Shoeybi, Catanzaro, and Ping]{dai2024nvlm}
Wenliang Dai, Nayeon Lee, Boxin Wang, Zhuolin Yang, Zihan Liu, Jon Barker, Tuomas Rintamaki, Mohammad Shoeybi, Bryan Catanzaro, and Wei Ping.
\newblock Nvlm: Open frontier-class multimodal llms.
\newblock \emph{arXiv preprint arXiv:2409.11402}, 2024.

\bibitem[Deitke et~al.(2024)Deitke, Liu, Wallingford, Ngo, Michel, Kusupati, Fan, Laforte, Voleti, Gadre, et~al.]{objaversexl}
Matt Deitke, Ruoshi Liu, Matthew Wallingford, Huong Ngo, Oscar Michel, Aditya Kusupati, Alan Fan, Christian Laforte, Vikram Voleti, Samir~Yitzhak Gadre, et~al.
\newblock Objaverse-xl: A universe of 10m+ 3d objects.
\newblock \emph{Advances in Neural Information Processing Systems}, 36, 2024.

\bibitem[Doddington(2002)]{doddington2002automatic}
George Doddington.
\newblock Automatic evaluation of machine translation quality using n-gram co-occurrence statistics.
\newblock In \emph{Proceedings of the second international conference on Human Language Technology Research}, pages 138--145, 2002.

\bibitem[Dong et~al.(2024)Dong, Li, Wu, Wang, Zhang, and Guo]{dong2024benchmarking}
Hongyuan Dong, Jiawen Li, Bohong Wu, Jiacong Wang, Yuan Zhang, and Haoyuan Guo.
\newblock Benchmarking and improving detail image caption.
\newblock \emph{arXiv preprint arXiv:2405.19092}, 2024.

\bibitem[Du et~al.(2024)Du, Zhou, Huo, Li, Zhao, Lu, Zhao, Wang, Chen, and Wen]{Event-Bench}
Yifan Du, Kun Zhou, Yuqi Huo, Yifan Li, Wayne~Xin Zhao, Haoyu Lu, Zijia Zhao, Bingning Wang, Weipeng Chen, and Ji-Rong Wen.
\newblock Towards event-oriented long video understanding.
\newblock \emph{arXiv preprint arXiv:2406.14129}, 2024.

\bibitem[Dubey et~al.(2024)Dubey, Jauhri, Pandey, Kadian, Al-Dahle, Letman, Mathur, Schelten, Yang, Fan, Goyal, Hartshorn, Yang, Mitra, Sravankumar, Korenev, Hinsvark, Rao, Zhang, Rodriguez, Gregerson, Spataru, Rozi{\`e}re, Biron, Tang, Chern, Caucheteux, Nayak, Bi, Marra, McConnell, Keller, Touret, Wu, Wong, Ferrer, Nikolaidis, Allonsius, Song, Pintz, Livshits, Esiobu, Choudhary, Mahajan, Garcia-Olano, Perino, Hupkes, Lakomkin, AlBadawy, Lobanova, Dinan, Smith, Radenovic, Zhang, Synnaeve, Lee, Anderson, Nail, Mialon, Pang, Cucurell, Nguyen, Korevaar, Xu, Touvron, Zarov, Ibarra, Kloumann, Misra, Evtimov, Copet, Lee, Geffert, Vranes, Park, Mahadeokar, Shah, van~der Linde, Billock, Hong, Lee, Fu, Chi, Huang, Liu, Wang, Yu, Bitton, Spisak, Park, Rocca, Johnstun, Saxe, Jia, Alwala, Upasani, Plawiak, Li, neth Heafield, Stone, El-Arini, Iyer, Malik, Chiu, Bhalla, Rantala-Yeary, van~der Maaten, Chen, Tan, Jenkins, Martin, Madaan, Malo, Blecher, Landzaat, de~Oliveira, Muzzi, Pasupuleti, Singh, Paluri, Kardas, Oldham,
  Rita, Pavlova, Kambadur, Lewis, Si, Singh, Hassan, Goyal, Torabi, Bashlykov, Bogoychev, Chatterji, Duchenne, cCelebi, Alrassy, Zhang, Li, Vasi{\'c}, Weng, Bhargava, Dubal, Krishnan, Koura, Xu, He, Dong, Srinivasan, Ganapathy, Calderer, Cabral, Stojnic, Raileanu, Girdhar, Patel, Sauvestre, Polidoro, Sumbaly, Taylor, Silva, Hou, Wang, Hosseini, Chennabasappa, Singh, Bell, Kim, Edunov, Nie, Narang, Raparthy, Shen, Wan, Bhosale, Zhang, Vandenhende, Batra, Whitman, Sootla, Collot, Gururangan, Borodinsky, Herman, Fowler, Sheasha, Georgiou, Scialom, Speckbacher, Mihaylov, Xiao, Karn, Goswami, Gupta, Ramanathan, Kerkez, Gonguet, Do, Vogeti, Petrovic, Chu, Xiong, Fu, ney Meers, Martinet, Wang, Tan, Xie, Jia, Wang, Goldschlag, Gaur, Babaei, Wen, Song, Zhang, Li, Mao, Coudert, Yan, Chen, Papakipos, Singh, Grattafiori, Jain, Kelsey, Shajnfeld, Gangidi, Victoria, Goldstand, Menon, Sharma, Boesenberg, Vaughan, Baevski, Feinstein, Kallet, Sangani, Yunus, Lupu, Alvarado, Caples, Gu, Ho, Poulton, Ryan, Ramchandani, Franco,
  Saraf, Chowdhury, Gabriel, Bharambe, Eisenman, Yazdan, James, Maurer, Leonhardi, Huang, Loyd, Paola, Paranjape, Liu, Wu, Ni, Hancock, Wasti, Spence, Stojkovic, Gamido, Montalvo, Parker, Burton, Mejia, Wang, Kim, Zhou, Hu, Chu, Cai, Tindal, Feichtenhofer, Civin, Beaty, Kreymer, Li, Wyatt, Adkins, Xu, Testuggine, David, Parikh, Liskovich, Foss, Wang, Le, Holland, Dowling, Jamil, Montgomery, Presani, Hahn, Wood, Brinkman, Arcaute, Dunbar, Smothers, Sun, Kreuk, Tian, Ozgenel, Caggioni, Guzm'an, Kanayet, Seide, Florez, Schwarz, Badeer, Swee, Halpern, Thattai, Herman, Sizov, Zhang, Lakshminarayanan, Shojanazeri, Zou, Wang, Zha, Habeeb, Rudolph, Suk, Aspegren, Goldman, Molybog, Tufanov, Veliche, Gat, Weissman, Geboski, Kohli, Asher, Gaya, Marcus, Tang, Chan, Zhen, Reizenstein, Teboul, Zhong, Jin, Yang, Cummings, Carvill, Shepard, McPhie, Torres, Ginsburg, Wang, Wu, KamHou, Saxena, Prasad, Khandelwal, Zand, Matosich, Veeraraghavan, Michelena, Li, Huang, Chawla, Lakhotia, Huang, Chen, Garg, Lavender, Silva, Bell,
  Zhang, Guo, Yu, Moshkovich, Wehrstedt, Khabsa, Avalani, Bhatt, Tsimpoukelli, Mankus, Hasson, Lennie, Reso, Groshev, Naumov, Lathi, Keneally, Seltzer, Valko, Restrepo, Patel, Vyatskov, Samvelyan, Clark, Macey, Wang, Hermoso, Metanat, Rastegari, Bansal, Santhanam, Parks, White, Bawa, Singhal, Egebo, Usunier, Laptev, Dong, Zhang, Cheng, Chernoguz, Hart, Salpekar, Kalinli, Kent, Parekh, Saab, Balaji, Rittner, Bontrager, Roux, Doll{\'a}r, Zvyagina, Ratanchandani, Yuvraj, Liang, Alao, Rodriguez, Ayub, Murthy, Nayani, Mitra, Li, Hogan, Battey, Wang, Maheswari, Howes, Rinott, Bondu, Datta, Chugh, Hunt, Dhillon, Sidorov, Pan, Verma, Yamamoto, Ramaswamy, Lindsay, Feng, Lin, Zha, Shankar, Zhang, Wang, Agarwal, Sajuyigbe, Chintala, Max, Chen, Kehoe, Satterfield, Govindaprasad, Gupta, Cho, Virk, Subramanian, Choudhury, Goldman, Remez, Glaser, Best, Kohler, Robinson, Li, Zhang, Matthews, Chou, Shaked, Vontimitta, Ajayi, Montanez, Mohan, Kumar, Mangla, Ionescu, Poenaru, Mihailescu, Ivanov, Li, Wang, Jiang, Bouaziz,
  Constable, Tang, Wang, Wu, Wang, Xia, Wu, Gao, Chen, Hu, Jia, Qi, Li, Zhang, Zhang, Adi, Nam, Wang, Hao, Qian, He, Rait, DeVito, Rosnbrick, Wen, Yang, and Zhao]{llama31}
Abhimanyu Dubey, Abhinav Jauhri, Abhinav Pandey, Abhishek Kadian, Ahmad Al-Dahle, Aiesha Letman, Akhil Mathur, Alan Schelten, Amy Yang, Angela Fan, Anirudh Goyal, Anthony~S. Hartshorn, Aobo Yang, Archi Mitra, Archie Sravankumar, Artem Korenev, Arthur Hinsvark, Arun Rao, Aston Zhang, Aur{\'e}lien Rodriguez, Austen Gregerson, Ava Spataru, Baptiste Rozi{\`e}re, Bethany Biron, Binh Tang, Bobbie Chern, Charlotte Caucheteux, Chaya Nayak, Chloe Bi, Chris Marra, Chris McConnell, Christian Keller, Christophe Touret, Chunyang Wu, Corinne Wong, Cristian~Cant{\'o}n Ferrer, Cyrus Nikolaidis, Damien Allonsius, Daniel Song, Danielle Pintz, Danny Livshits, David Esiobu, Dhruv Choudhary, Dhruv Mahajan, Diego Garcia-Olano, Diego Perino, Dieuwke Hupkes, Egor Lakomkin, Ehab~A. AlBadawy, Elina Lobanova, Emily Dinan, Eric~Michael Smith, Filip Radenovic, Frank Zhang, Gabriele Synnaeve, Gabrielle Lee, Georgia~Lewis Anderson, Graeme Nail, Gr{\'e}goire Mialon, Guanglong Pang, Guillem Cucurell, Hailey Nguyen, Hannah Korevaar, Hu Xu,
  Hugo Touvron, Iliyan Zarov, Imanol~Arrieta Ibarra, Isabel~M. Kloumann, Ishan Misra, Ivan Evtimov, Jade Copet, Jaewon Lee, Jan~Laurens Geffert, Jana Vranes, Jason Park, Jay Mahadeokar, Jeet Shah, Jelmer van~der Linde, Jennifer Billock, Jenny Hong, Jenya Lee, Jeremy Fu, Jianfeng Chi, Jianyu Huang, Jiawen Liu, Jie Wang, Jiecao Yu, Joanna Bitton, Joe Spisak, Jongsoo Park, Joseph Rocca, Joshua Johnstun, Joshua Saxe, Ju-Qing Jia, Kalyan~Vasuden Alwala, K. Upasani, Kate Plawiak, Keqian Li, Ken-591 neth Heafield, Kevin Stone, Khalid El-Arini, Krithika Iyer, Kshitiz Malik, Kuenley Chiu, Kunal Bhalla, Lauren Rantala-Yeary, Laurens van~der Maaten, Lawrence Chen, Liang Tan, Liz Jenkins, Louis Martin, Lovish Madaan, Lubo Malo, Lukas Blecher, Lukas Landzaat, Luke de Oliveira, Madeline Muzzi, Mahesh~Babu Pasupuleti, Mannat Singh, Manohar Paluri, Marcin Kardas, Mathew Oldham, Mathieu Rita, Maya Pavlova, Melissa Hall~Melanie Kambadur, Mike Lewis, Min Si, Mitesh~Kumar Singh, Mona Hassan, Naman Goyal, Narjes Torabi, Nikolay
  Bashlykov, Nikolay Bogoychev, Niladri~S. Chatterji, Olivier Duchenne, Onur cCelebi, Patrick Alrassy, Pengchuan Zhang, Pengwei Li, Petar Vasi{\'c}, Peter Weng, Prajjwal Bhargava, Pratik Dubal, Praveen Krishnan, Punit~Singh Koura, Puxin Xu, Qing He, Qingxiao Dong, Ragavan Srinivasan, Raj Ganapathy, Ramon Calderer, Ricardo~Silveira Cabral, Robert Stojnic, Roberta Raileanu, Rohit Girdhar, Rohit Patel, Romain Sauvestre, Ronnie Polidoro, Roshan Sumbaly, Ross Taylor, Ruan Silva, Rui Hou, Rui Wang, Saghar Hosseini, Sahana Chennabasappa, Sanjay Singh, Sean Bell, Seohyun~Sonia Kim, Sergey Edunov, Shaoliang Nie, Sharan Narang, Sharath~Chandra Raparthy, Sheng Shen, Shengye Wan, Shruti Bhosale, Shun Zhang, Simon Vandenhende, Soumya Batra, Spencer Whitman, Sten Sootla, Stephane Collot, Suchin Gururangan, Sydney Borodinsky, Tamar Herman, Tara Fowler, Tarek Sheasha, Thomas Georgiou, Thomas Scialom, Tobias Speckbacher, Todor Mihaylov, Tong Xiao, Ujjwal Karn, Vedanuj Goswami, Vibhor Gupta, Vignesh Ramanathan, Viktor Kerkez,
  Vincent Gonguet, Virginie Do, Vish Vogeti, Vladan Petrovic, Weiwei Chu, Wenhan Xiong, Wenyin Fu, Whit ney Meers, Xavier Martinet, Xiaodong Wang, Xiaoqing~Ellen Tan, Xinfeng Xie, Xuchao Jia, Xuewei Wang, Yaelle Goldschlag, Yashesh Gaur, Yasmine Babaei, Yiqian Wen, Yiwen Song, Yuchen Zhang, Yue Li, Yuning Mao, Zacharie~Delpierre Coudert, Zhengxu Yan, Zhengxing Chen, Zoe Papakipos, Aaditya~K. Singh, Aaron Grattafiori, Abha Jain, Adam Kelsey, Adam Shajnfeld, Adi Gangidi, Adolfo Victoria, Ahuva Goldstand, Ajay Menon, Ajay Sharma, Alex Boesenberg, Alex Vaughan, Alexei Baevski, Allie Feinstein, Amanda Kallet, Amit Sangani, Anam Yunus, Andrei Lupu, Andres Alvarado, Andrew Caples, Andrew Gu, Andrew Ho, Andrew Poulton, Andrew Ryan, Ankit Ramchandani, Annie Franco, Aparajita Saraf, Arkabandhu Chowdhury, Ashley Gabriel, Ashwin Bharambe, Assaf Eisenman, Azadeh Yazdan, Beau James, Ben Maurer, Ben Leonhardi, Po-Yao~(Bernie) Huang, Beth Loyd, Beto~De Paola, Bhargavi Paranjape, Bing Liu, Bo Wu, Boyu Ni, Braden Hancock, Bram
  Wasti, Brandon Spence, Brani Stojkovic, Brian Gamido, Britt Montalvo, Carl Parker, Carly Burton, Catalina Mejia, Changhan Wang, Changkyu Kim, Chao Zhou, Chester Hu, Ching-Hsiang Chu, Chris Cai, Chris Tindal, Christoph Feichtenhofer, Damon Civin, Dana Beaty, Daniel Kreymer, Shang-Wen Li, Danny Wyatt, David Adkins, David Xu, Davide Testuggine, Delia David, Devi Parikh, Diana Liskovich, Didem Foss, Dingkang Wang, Duc Le, Dustin Holland, Edward Dowling, Eissa Jamil, Elaine Montgomery, Eleonora Presani, Emily Hahn, Emily Wood, Erik Brinkman, Esteban Arcaute, Evan Dunbar, Evan Smothers, Fei Sun, Felix Kreuk, Feng Tian, Firat Ozgenel, Francesco Caggioni, Francisco Guzm'an, Frank~J. Kanayet, Frank Seide, Gabriela~Medina Florez, Gabriella Schwarz, Gada Badeer, Georgia Swee, Gil Halpern, Govind Thattai, Grant Herman, Grigory~G. Sizov, Guangyi Zhang, Guna Lakshminarayanan, Hamid Shojanazeri, Han Zou, Hannah Wang, Han Zha, Haroun Habeeb, Harrison Rudolph, Helen Suk, Henry Aspegren, Hunter Goldman, Igor Molybog, Igor
  Tufanov, Irina-Elena Veliche, Itai Gat, Jake Weissman, James Geboski, James Kohli, Japhet Asher, Jean-Baptiste Gaya, Jeff Marcus, Jeff Tang, Jennifer Chan, Jenny Zhen, Jeremy Reizenstein, Jeremy Teboul, Jessica Zhong, Jian Jin, Jingyi Yang, Joe Cummings, Jon Carvill, Jon Shepard, Jonathan McPhie, Jonathan Torres, Josh Ginsburg, Junjie Wang, Kaixing(Kai) Wu, U KamHou, Karan Saxena, Karthik Prasad, Kartikay Khandelwal, Katayoun Zand, Kathy Matosich, Kaushik Veeraraghavan, Kelly Michelena, Keqian Li, Kun Huang, Kunal Chawla, Kushal Lakhotia, Kyle Huang, Lailin Chen, Lakshya Garg, A Lavender, Leandro Silva, Lee Bell, Lei Zhang, Liangpeng Guo, Licheng Yu, Liron Moshkovich, Luca Wehrstedt, Madian Khabsa, Manav Avalani, Manish Bhatt, Maria Tsimpoukelli, Martynas Mankus, Matan Hasson, Matthew Lennie, Matthias Reso, Maxim Groshev, Maxim Naumov, Maya Lathi, Meghan Keneally, Michael~L. Seltzer, Michal Valko, Michelle Restrepo, Mihir Patel, Mik Vyatskov, Mikayel Samvelyan, Mike Clark, Mike Macey, Mike Wang,
  Miquel~Jubert Hermoso, Mo Metanat, Mohammad Rastegari, Munish Bansal, Nandhini Santhanam, Natascha Parks, Natasha White, Navyata Bawa, Nayan Singhal, Nick Egebo, Nicolas Usunier, Nikolay~Pavlovich Laptev, Ning Dong, Ning Zhang, Norman Cheng, Oleg Chernoguz, Olivia Hart, Omkar Salpekar, Ozlem Kalinli, Parkin Kent, Parth Parekh, Paul Saab, Pavan Balaji, Pedro Rittner, Philip Bontrager, Pierre Roux, Piotr Doll{\'a}r, Polina Zvyagina, Prashant Ratanchandani, Pritish Yuvraj, Qian Liang, Rachad Alao, Rachel Rodriguez, Rafi Ayub, Raghotham Murthy, Raghu Nayani, Rahul Mitra, Raymond Li, Rebekkah Hogan, Robin Battey, Rocky Wang, Rohan Maheswari, Russ Howes, Ruty Rinott, Sai~Jayesh Bondu, Samyak Datta, Sara Chugh, Sara Hunt, Sargun Dhillon, Sasha Sidorov, Satadru Pan, Saurabh Verma, Seiji Yamamoto, Sharadh Ramaswamy, Shaun Lindsay, Sheng Feng, Shenghao Lin, Shengxin~Cindy Zha, Shiva Shankar, Shuqiang Zhang, Sinong Wang, Sneha Agarwal, Soji Sajuyigbe, Soumith Chintala, Stephanie Max, Stephen Chen, Steve Kehoe, Steve
  Satterfield, Sudarshan Govindaprasad, Sumit Gupta, Sung-Bae Cho, Sunny Virk, Suraj Subramanian, Sy Choudhury, Sydney Goldman, Tal Remez, Tamar Glaser, Tamara Best, Thilo Kohler, Thomas Robinson, Tianhe Li, Tianjun Zhang, Tim Matthews, Timothy Chou, Tzook Shaked, Varun Vontimitta, Victoria Ajayi, Victoria Montanez, Vijai Mohan, Vinay~Satish Kumar, Vishal Mangla, Vlad Ionescu, Vlad~Andrei Poenaru, Vlad~T. Mihailescu, Vladimir Ivanov, Wei Li, Wenchen Wang, Wenwen Jiang, Wes Bouaziz, Will Constable, Xia Tang, Xiaofang Wang, Xiaojian Wu, Xiaolan Wang, Xide Xia, Xilun Wu, Xinbo Gao, Yanjun Chen, Ye Hu, Ye Jia, Ye Qi, Yenda Li, Yilin Zhang, Ying Zhang, Yossi Adi, Youngjin Nam, Yu Wang, Yuchen Hao, Yundi Qian, Yuzi He, Zach Rait, Zachary DeVito, Zef Rosnbrick, Zhaoduo Wen, Zhenyu Yang, and Zhiwei Zhao.
\newblock The llama 3 herd of models.
\newblock \emph{ArXiv}, abs/2407.21783, 2024.

\bibitem[Fang et~al.(2024)Fang, Mao, Duan, Zhao, Li, Lin, and Chen]{mmbenchvideo}
Xinyu Fang, Kangrui Mao, Haodong Duan, Xiangyu Zhao, Yining Li, Dahua Lin, and Kai Chen.
\newblock Mmbench-video: A long-form multi-shot benchmark for holistic video understanding.
\newblock \emph{arXiv preprint arXiv:2406.14515}, 2024.

\bibitem[Fei et~al.(2024)Fei, Li, Deng, Wang, Liu, and Wang]{fei2024video}
Jiajun Fei, Dian Li, Zhidong Deng, Zekun Wang, Gang Liu, and Hui Wang.
\newblock Video-ccam: Enhancing video-language understanding with causal cross-attention masks for short and long videos.
\newblock \emph{arXiv preprint arXiv:2408.14023}, 2024.

\bibitem[Fu et~al.(2024{\natexlab{a}})Fu, Dai, Luo, Li, Ren, Zhang, Wang, Zhou, Shen, Zhang, et~al.]{videomme}
Chaoyou Fu, Yuhan Dai, Yondong Luo, Lei Li, Shuhuai Ren, Renrui Zhang, Zihan Wang, Chenyu Zhou, Yunhang Shen, Mengdan Zhang, et~al.
\newblock Video-mme: The first-ever comprehensive evaluation benchmark of multi-modal llms in video analysis.
\newblock \emph{arXiv preprint arXiv:2405.21075}, 2024{\natexlab{a}}.

\bibitem[Fu et~al.(2024{\natexlab{b}})Fu, Lin, Long, Shen, Zhao, Zhang, Dong, Wang, Yin, Ma, et~al.]{fu2024vita}
Chaoyou Fu, Haojia Lin, Zuwei Long, Yunhang Shen, Meng Zhao, Yifan Zhang, Shaoqi Dong, Xiong Wang, Di Yin, Long Ma, et~al.
\newblock Vita: Towards open-source interactive omni multimodal llm.
\newblock \emph{arXiv preprint arXiv:2408.05211}, 2024{\natexlab{b}}.

\bibitem[Han et~al.(2023{\natexlab{a}})Han, Zhang, Shao, Gao, Xu, Xiao, Zhang, Liu, Wen, Guo, et~al.]{imagebind-llm}
Jiaming Han, Renrui Zhang, Wenqi Shao, Peng Gao, Peng Xu, Han Xiao, Kaipeng Zhang, Chris Liu, Song Wen, Ziyu Guo, et~al.
\newblock Imagebind-llm: Multi-modality instruction tuning.
\newblock \emph{arXiv preprint arXiv:2309.03905}, 2023{\natexlab{a}}.

\bibitem[Han et~al.(2023{\natexlab{b}})Han, Yang, Chang, and Wang]{han2023shot2story20k}
Mingfei Han, Linjie Yang, Xiaojun Chang, and Heng Wang.
\newblock Shot2story20k: A new benchmark for comprehensive understanding of multi-shot videos.
\newblock \emph{arXiv preprint arXiv:2312.10300}, 2023{\natexlab{b}}.

\bibitem[He et~al.(2023)He, Bai, Lin, Zhao, Hu, Sheng, Yi, Li, and Liu]{T3bench}
Yuze He, Yushi Bai, Matthieu Lin, Wang Zhao, Yubin Hu, Jenny Sheng, Ran Yi, Juanzi Li, and Yong-Jin Liu.
\newblock T $^3$ bench: Benchmarking current progress in text-to-3d generation.
\newblock \emph{arXiv preprint arXiv:2310.02977}, 2023.

\bibitem[Hessel et~al.(2021)Hessel, Holtzman, Forbes, Bras, and Choi]{hessel2021clipscore}
Jack Hessel, Ari Holtzman, Maxwell Forbes, Ronan~Le Bras, and Yejin Choi.
\newblock Clipscore: A reference-free evaluation metric for image captioning.
\newblock \emph{arXiv preprint arXiv:2104.08718}, 2021.

\bibitem[Hong et~al.(2023)Hong, Zhen, Chen, Zheng, Du, Chen, and Gan]{3dllm}
Yining Hong, Haoyu Zhen, Peihao Chen, Shuhong Zheng, Yilun Du, Zhenfang Chen, and Chuang Gan.
\newblock 3d-llm: Injecting the 3d world into large language models.
\newblock \emph{arXiv}, 2023.

\bibitem[Jia et~al.(2024)Jia, Chen, Yu, Wang, Niu, Liu, Li, and Huang]{sceneverse}
Baoxiong Jia, Yixin Chen, Huangyue Yu, Yan Wang, Xuesong Niu, Tengyu Liu, Qing Li, and Siyuan Huang.
\newblock Sceneverse: Scaling 3d vision-language learning for grounded scene understanding.
\newblock In \emph{European Conference on Computer Vision (ECCV)}, 2024.

\bibitem[Kang et~al.(2023)Kang, Mun, Lee, and Roh]{kang2023noise}
Wooyoung Kang, Jonghwan Mun, Sungjun Lee, and Byungseok Roh.
\newblock Noise-aware learning from web-crawled image-text data for image captioning.
\newblock In \emph{Proceedings of the IEEE/CVF International Conference on Computer Vision}, pages 2942--2952, 2023.

\bibitem[Kesen et~al.(2023)Kesen, Pedrotti, Dogan, Cafagna, Acikgoz, Parcalabescu, Calixto, Frank, Gatt, Erdem, et~al.]{VilMA}
Ilker Kesen, Andrea Pedrotti, Mustafa Dogan, Michele Cafagna, Emre~Can Acikgoz, Letitia Parcalabescu, Iacer Calixto, Anette Frank, Albert Gatt, Aykut Erdem, et~al.
\newblock Vilma: A zero-shot benchmark for linguistic and temporal grounding in video-language models.
\newblock \emph{arXiv preprint arXiv:2311.07022}, 2023.

\bibitem[Kil et~al.(2024)Kil, Mai, Lee, Wang, Cheng, Wang, Liu, Chowdhury, and Chao]{compbench}
Jihyung Kil, Zheda Mai, Justin Lee, Zihe Wang, Kerrie Cheng, Lemeng Wang, Ye Liu, Arpita Chowdhury, and Wei-Lun Chao.
\newblock Compbench: A comparative reasoning benchmark for multimodal llms.
\newblock \emph{arXiv preprint arXiv:2407.16837}, 2024.

\bibitem[Kim et~al.(2022)Kim, Kim, Lee, Yoo, and Lee]{kim2022mutual}
Jin-Hwa Kim, Yunji Kim, Jiyoung Lee, Kang~Min Yoo, and Sang-Woo Lee.
\newblock Mutual information divergence: A unified metric for multimodal generative models.
\newblock \emph{Advances in Neural Information Processing Systems}, 35:\penalty0 35072--35086, 2022.

\bibitem[Krishna et~al.(2017)Krishna, Hata, Ren, Fei-Fei, and Carlos~Niebles]{krishna2017dense}
Ranjay Krishna, Kenji Hata, Frederic Ren, Li Fei-Fei, and Juan Carlos~Niebles.
\newblock Dense-captioning events in videos.
\newblock In \emph{Proceedings of the IEEE international conference on computer vision}, pages 706--715, 2017.

\bibitem[Lee et~al.(2024)Lee, Park, and Kang]{lee2024fleur}
Yebin Lee, Imseong Park, and Myungjoo Kang.
\newblock Fleur: An explainable reference-free evaluation metric for image captioning using a large multimodal model.
\newblock \emph{arXiv preprint arXiv:2406.06004}, 2024.

\bibitem[Lei et~al.(2020)Lei, Yu, Berg, and Bansal]{lei2020tvr}
Jie Lei, Licheng Yu, Tamara~L Berg, and Mohit Bansal.
\newblock Tvr: A large-scale dataset for video-subtitle moment retrieval.
\newblock In \emph{Computer Vision--ECCV 2020: 16th European Conference, Glasgow, UK, August 23--28, 2020, Proceedings, Part XXI 16}, pages 447--463. Springer, 2020.

\bibitem[Li et~al.(2023{\natexlab{a}})Li, Zhang, Yang, Zhang, Pu, and Liu]{MagnifierBench}
Bo Li, Peiyuan Zhang, Jingkang Yang, Yuanhan Zhang, Fanyi Pu, and Ziwei Liu.
\newblock Otterhd: A high-resolution multi-modality model.
\newblock \emph{arXiv preprint arXiv:2311.04219}, 2023{\natexlab{a}}.

\bibitem[Li et~al.(2024{\natexlab{a}})Li, Ge, Ge, Wang, Wang, Zhang, and Shan]{li2024seed}
Bohao Li, Yuying Ge, Yixiao Ge, Guangzhi Wang, Rui Wang, Ruimao Zhang, and Ying Shan.
\newblock Seed-bench: Benchmarking multimodal large language models.
\newblock In \emph{Proceedings of the IEEE/CVF Conference on Computer Vision and Pattern Recognition}, pages 13299--13308, 2024{\natexlab{a}}.

\bibitem[Li et~al.(2024{\natexlab{b}})Li, Ge, Ge, Wang, Wang, Zhang, and Shan]{seedbench}
Bohao Li, Yuying Ge, Yixiao Ge, Guangzhi Wang, Rui Wang, Ruimao Zhang, and Ying Shan.
\newblock Seed-bench: Benchmarking multimodal large language models.
\newblock In \emph{Proceedings of the IEEE/CVF Conference on Computer Vision and Pattern Recognition}, pages 13299--13308, 2024{\natexlab{b}}.

\bibitem[Li et~al.(2024{\natexlab{c}})Li, Zhang, Guo, Zhang, Li, Zhang, Zhang, Li, Liu, and Li]{llavaonevision}
Bo Li, Yuanhan Zhang, Dong Guo, Renrui Zhang, Feng Li, Hao Zhang, Kaichen Zhang, Yanwei Li, Ziwei Liu, and Chunyuan Li.
\newblock Llava-onevision: Easy visual task transfer.
\newblock \emph{arXiv preprint arXiv:2408.03326}, 2024{\natexlab{c}}.

\bibitem[Li et~al.(2024{\natexlab{d}})Li, Zheng, Zhu, Mai, Zhang, Wonka, and Ghanem]{vividzoo}
Bing Li, Cheng Zheng, Wenxuan Zhu, Jinjie Mai, Biao Zhang, Peter Wonka, and Bernard Ghanem.
\newblock Vivid-zoo: Multi-view video generation with diffusion model, 2024{\natexlab{d}}.

\bibitem[Li et~al.(2024{\natexlab{e}})Li, Liu, Wu, Wang, Shen, Qu, Niu, Wang, Chen, and Li]{li2024aria}
Dongxu Li, Yudong Liu, Haoning Wu, Yue Wang, Zhiqi Shen, Bowen Qu, Xinyao Niu, Guoyin Wang, Bei Chen, and Junnan Li.
\newblock Aria: An open multimodal native mixture-of-experts model.
\newblock \emph{arXiv preprint arXiv:2410.05993}, 2024{\natexlab{e}}.

\bibitem[Li and Lu(2024)]{li2024survey}
Jian Li and Weiheng Lu.
\newblock A survey on benchmarks of multimodal large language models.
\newblock \emph{arXiv preprint arXiv:2408.08632}, 2024.

\bibitem[Li et~al.(2023{\natexlab{b}})Li, Li, Savarese, and Hoi]{blip2}
Junnan Li, Dongxu Li, Silvio Savarese, and Steven Hoi.
\newblock Blip-2: Bootstrapping language-image pre-training with frozen image encoders and large language models.
\newblock In \emph{International conference on machine learning}, pages 19730--19742. PMLR, 2023{\natexlab{b}}.

\bibitem[Li et~al.(2023{\natexlab{c}})Li, He, Wang, Li, Wang, Luo, Wang, Wang, and Qiao]{videochat}
Kunchang Li, Yinan He, Yi Wang, Yizhuo Li, Wenhai Wang, Ping Luo, Yali Wang, Limin Wang, and Yu Qiao.
\newblock Videochat: Chat-centric video understanding.
\newblock \emph{arXiv preprint arXiv:2305.06355}, 2023{\natexlab{c}}.

\bibitem[Li et~al.(2024{\natexlab{f}})Li, Wang, He, Li, Wang, Liu, Wang, Xu, Chen, Luo, et~al.]{mvbench}
Kunchang Li, Yali Wang, Yinan He, Yizhuo Li, Yi Wang, Yi Liu, Zun Wang, Jilan Xu, Guo Chen, Ping Luo, et~al.
\newblock Mvbench: A comprehensive multi-modal video understanding benchmark.
\newblock In \emph{Proceedings of the IEEE/CVF Conference on Computer Vision and Pattern Recognition}, pages 22195--22206, 2024{\natexlab{f}}.

\bibitem[Li et~al.(2021)Li, Takehara, Taketomi, Zheng, and Nie{\ss}ner]{deform4dthing}
Yang Li, Hikari Takehara, Takafumi Taketomi, Bo Zheng, and Matthias Nie{\ss}ner.
\newblock 4dcomplete: Non-rigid motion estimation beyond the observable surface.
\newblock \emph{2021 IEEE/CVF International Conference on Computer Vision (ICCV)}, pages 12686--12696, 2021.

\bibitem[Li et~al.(2023{\natexlab{d}})Li, Du, Zhou, Wang, Zhao, and Wen]{POPE2023}
Yifan Li, Yifan Du, Kun Zhou, Jinpeng Wang, Wayne~Xin Zhao, and Ji-Rong Wen.
\newblock Evaluating object hallucination in large vision-language models.
\newblock \emph{arXiv preprint arXiv:2305.10355}, 2023{\natexlab{d}}.

\bibitem[Liang et~al.(2024{\natexlab{a}})Liang, Chen, and Zhang]{liang2024evqascore}
Hao Liang, Zirong Chen, and Wentao Zhang.
\newblock Evqascore: Efficient video question answering data evaluation.
\newblock \emph{arXiv preprint arXiv:2411.06908}, 2024{\natexlab{a}}.

\bibitem[Liang et~al.(2024{\natexlab{b}})Liang, Yin, Xu, Liang, Wang, Plataniotis, Zhao, and Wei]{diffusion4d}
Hanwen Liang, Yuyang Yin, Dejia Xu, Hanxue Liang, Zhangyang Wang, Konstantinos~N Plataniotis, Yao Zhao, and Yunchao Wei.
\newblock Diffusion4d: Fast spatial-temporal consistent 4d generation via video diffusion models.
\newblock \emph{arXiv preprint arXiv:2405.16645}, 2024{\natexlab{b}}.

\bibitem[Lin(2004)]{rouge}
Chin-Yew Lin.
\newblock Rouge: A package for automatic evaluation of summaries.
\newblock In \emph{Text summarization branches out}, pages 74--81, 2004.

\bibitem[Lin et~al.(2024)Lin, Yin, Ping, Molchanov, Shoeybi, and Han]{lin2024vila}
Ji Lin, Hongxu Yin, Wei Ping, Pavlo Molchanov, Mohammad Shoeybi, and Song Han.
\newblock Vila: On pre-training for visual language models.
\newblock In \emph{Proceedings of the IEEE/CVF Conference on Computer Vision and Pattern Recognition}, pages 26689--26699, 2024.

\bibitem[Lin et~al.(2023)Lin, Liu, Zhang, Gao, Qiu, Xiao, Qiu, Lin, Shao, Chen, et~al.]{lin2023sphinx}
Ziyi Lin, Chris Liu, Renrui Zhang, Peng Gao, Longtian Qiu, Han Xiao, Han Qiu, Chen Lin, Wenqi Shao, Keqin Chen, et~al.
\newblock Sphinx: The joint mixing of weights, tasks, and visual embeddings for multi-modal large language models.
\newblock \emph{arXiv preprint arXiv:2311.07575}, 2023.

\bibitem[Liu et~al.(2024{\natexlab{a}})Liu, Li, Li, and Lee]{LLaVA-15}
Haotian Liu, Chunyuan Li, Yuheng Li, and Yong~Jae Lee.
\newblock Improved baselines with visual instruction tuning.
\newblock In \emph{Proceedings of the IEEE/CVF Conference on Computer Vision and Pattern Recognition}, pages 26296--26306, 2024{\natexlab{a}}.

\bibitem[Liu et~al.(2024{\natexlab{b}})Liu, Li, Wu, and Lee]{LLaVA}
Haotian Liu, Chunyuan Li, Qingyang Wu, and Yong~Jae Lee.
\newblock Visual instruction tuning.
\newblock \emph{Advances in neural information processing systems}, 36, 2024{\natexlab{b}}.

\bibitem[Liu et~al.(2024{\natexlab{c}})Liu, Li, Wu, and Lee]{LLaVA-Bench}
Haotian Liu, Chunyuan Li, Qingyang Wu, and Yong~Jae Lee.
\newblock Visual instruction tuning.
\newblock \emph{Advances in neural information processing systems}, 36, 2024{\natexlab{c}}.

\bibitem[Liu et~al.(2024{\natexlab{d}})Liu, Wang, Ma, Wu, Ma, Wei, Jiao, Wu, and Hu]{liu2024kangaroo}
Jiajun Liu, Yibing Wang, Hanghang Ma, Xiaoping Wu, Xiaoqi Ma, Xiaoming Wei, Jianbin Jiao, Enhua Wu, and Jie Hu.
\newblock Kangaroo: A powerful video-language model supporting long-context video input.
\newblock \emph{arXiv preprint arXiv:2408.15542}, 2024{\natexlab{d}}.

\bibitem[Liu et~al.(2025)Liu, Li, Tang, Ge, Shan, and Li]{liu2025st}
Ruyang Liu, Chen Li, Haoran Tang, Yixiao Ge, Ying Shan, and Ge Li.
\newblock St-llm: Large language models are effective temporal learners.
\newblock In \emph{European Conference on Computer Vision}, pages 1--18. Springer, 2025.

\bibitem[Liu et~al.(2023{\natexlab{a}})Liu, Duan, Zhang, Li, Zhang, Zhao, Yuan, Wang, He, Liu, et~al.]{mmbench}
Yuan Liu, Haodong Duan, Yuanhan Zhang, Bo Li, Songyang Zhang, Wangbo Zhao, Yike Yuan, Jiaqi Wang, Conghui He, Ziwei Liu, et~al.
\newblock Mmbench: Is your multi-modal model an all-around player?
\newblock \emph{arXiv preprint arXiv:2307.06281}, 2023{\natexlab{a}}.

\bibitem[Liu et~al.(2023{\natexlab{b}})Liu, Li, Yang, Li, Yin, Liu, Jin, and Bai]{liu2023hidden}
Yuliang Liu, Zhang Li, Biao Yang, Chunyuan Li, Xucheng Yin, Cheng-lin Liu, Lianwen Jin, and Xiang Bai.
\newblock On the hidden mystery of ocr in large multimodal models.
\newblock \emph{arXiv preprint arXiv:2305.07895}, 2023{\natexlab{b}}.

\bibitem[Liu et~al.(2024{\natexlab{e}})Liu, Li, Liu, Wang, Ren, Li, Chen, Sun, and Hou]{tempcompass}
Yuanxin Liu, Shicheng Li, Yi Liu, Yuxiang Wang, Shuhuai Ren, Lei Li, Sishuo Chen, Xu Sun, and Lu Hou.
\newblock Tempcompass: Do video llms really understand videos?
\newblock \emph{arXiv preprint arXiv:2403.00476}, 2024{\natexlab{e}}.

\bibitem[Liu et~al.(2024{\natexlab{f}})Liu, Dong, Liu, Hu, Lu, and Rao]{liu2024oryx}
Zuyan Liu, Yuhao Dong, Ziwei Liu, Winston Hu, Jiwen Lu, and Yongming Rao.
\newblock Oryx mllm: On-demand spatial-temporal understanding at arbitrary resolution.
\newblock \emph{arXiv preprint arXiv:2409.12961}, 2024{\natexlab{f}}.

\bibitem[Maaz et~al.(2024)Maaz, Rasheed, Khan, and Khan]{VideoChatGPT}
Muhammad Maaz, Hanoona Rasheed, Salman Khan, and Fahad~Shahbaz Khan.
\newblock Video-chatgpt: Towards detailed video understanding via large vision and language models.
\newblock In \emph{Proceedings of the 62nd Annual Meeting of the Association for Computational Linguistics (ACL 2024)}, 2024.

\bibitem[Mangalam et~al.(2024)Mangalam, Akshulakov, and Malik]{egoschema}
Karttikeya Mangalam, Raiymbek Akshulakov, and Jitendra Malik.
\newblock Egoschema: A diagnostic benchmark for very long-form video language understanding.
\newblock \emph{Advances in Neural Information Processing Systems}, 36, 2024.

\bibitem[Nagrani et~al.(2024)Nagrani, Zhang, Mehran, Nitesh, Gundavarapu, Jha, Myers, Zhou, Gong, Schmid, Sirotenko, Zhu, Weyand, and GoogleResearch]{Neptune}
Arsha Nagrani, Mingda Zhang, Ramin Mehran, Rachel~Hornung Nitesh, Bharadwaj Gundavarapu, Nilpa Jha, Austin Myers, Xingyi Zhou, Boqing Gong, Cordelia Schmid, Mikhail Sirotenko, Yukun Zhu, Tobias Weyand, and † GoogleResearch.
\newblock Neptune: The long orbit to benchmarking long video understanding.
\newblock \emph{ArXiv}, abs/2412.09582, 2024.

\bibitem[Nguyen et~al.(2024)Nguyen, Bi, Vosoughi, Tian, Fazli, and Xu]{OsCaR}
Nguyen Nguyen, Jing Bi, Ali Vosoughi, Yapeng Tian, Pooyan Fazli, and Chenliang Xu.
\newblock Oscar: Object state captioning and state change representation.
\newblock \emph{arXiv preprint arXiv:2402.17128}, 2024.

\bibitem[Ning et~al.(2023)Ning, Zhu, Xie, Lin, Cui, Yuan, Chen, and Yuan]{Video-Bench}
Munan Ning, Bin Zhu, Yujia Xie, Bin Lin, Jiaxi Cui, Lu Yuan, Dongdong Chen, and Li Yuan.
\newblock Video-bench: A comprehensive benchmark and toolkit for evaluating video-based large language models.
\newblock \emph{arXiv preprint arXiv:2311.16103}, 2023.

\bibitem[OpenAI et~al.(2024)OpenAI, Achiam, Adler, Agarwal, Ahmad, Akkaya, Aleman, Almeida, Altenschmidt, Altman, Anadkat, Avila, Babuschkin, Balaji, Balcom, Baltescu, Bao, Bavarian, Belgum, Bello, Berdine, Bernadett-Shapiro, Berner, Bogdonoff, Boiko, Boyd, Brakman, Brockman, Brooks, Brundage, Button, Cai, Campbell, Cann, Carey, Carlson, Carmichael, Chan, Chang, Chantzis, Chen, Chen, Chen, Chen, Chen, Chess, Cho, Chu, Chung, Cummings, Currier, Dai, Decareaux, Degry, Deutsch, Deville, Dhar, Dohan, Dowling, Dunning, Ecoffet, Eleti, Eloundou, Farhi, Fedus, Felix, Fishman, Forte, Fulford, Gao, Georges, Gibson, Goel, Gogineni, Goh, Gontijo-Lopes, Gordon, Grafstein, Gray, Greene, Gross, Gu, Guo, Hallacy, Han, Harris, He, Heaton, Heidecke, Hesse, Hickey, Hickey, Hoeschele, Houghton, Hsu, Hu, Hu, Huizinga, Jain, Jain, Jang, Jiang, Jiang, Jin, Jin, Jomoto, Jonn, Jun, Kaftan, Łukasz Kaiser, Kamali, Kanitscheider, Keskar, Khan, Kilpatrick, Kim, Kim, Kim, Kirchner, Kiros, Knight, Kokotajlo, Łukasz Kondraciuk, Kondrich,
  Konstantinidis, Kosic, Krueger, Kuo, Lampe, Lan, Lee, Leike, Leung, Levy, Li, Lim, Lin, Lin, Litwin, Lopez, Lowe, Lue, Makanju, Malfacini, Manning, Markov, Markovski, Martin, Mayer, Mayne, McGrew, McKinney, McLeavey, McMillan, McNeil, Medina, Mehta, Menick, Metz, Mishchenko, Mishkin, Monaco, Morikawa, Mossing, Mu, Murati, Murk, Mély, Nair, Nakano, Nayak, Neelakantan, Ngo, Noh, Ouyang, O'Keefe, Pachocki, Paino, Palermo, Pantuliano, Parascandolo, Parish, Parparita, Passos, Pavlov, Peng, Perelman, de~Avila Belbute~Peres, Petrov, de~Oliveira~Pinto, Michael, Pokorny, Pokrass, Pong, Powell, Power, Power, Proehl, Puri, Radford, Rae, Ramesh, Raymond, Real, Rimbach, Ross, Rotsted, Roussez, Ryder, Saltarelli, Sanders, Santurkar, Sastry, Schmidt, Schnurr, Schulman, Selsam, Sheppard, Sherbakov, Shieh, Shoker, Shyam, Sidor, Sigler, Simens, Sitkin, Slama, Sohl, Sokolowsky, Song, Staudacher, Such, Summers, Sutskever, Tang, Tezak, Thompson, Tillet, Tootoonchian, Tseng, Tuggle, Turley, Tworek, Uribe, Vallone, Vijayvergiya,
  Voss, Wainwright, Wang, Wang, Wang, Ward, Wei, Weinmann, Welihinda, Welinder, Weng, Weng, Wiethoff, Willner, Winter, Wolrich, Wong, Workman, Wu, Wu, Wu, Xiao, Xu, Yoo, Yu, Yuan, Zaremba, Zellers, Zhang, Zhang, Zhao, Zheng, Zhuang, Zhuk, and Zoph]{openai2024gpt4}
OpenAI, Josh Achiam, Steven Adler, Sandhini Agarwal, Lama Ahmad, Ilge Akkaya, Florencia~Leoni Aleman, Diogo Almeida, Janko Altenschmidt, Sam Altman, Shyamal Anadkat, Red Avila, Igor Babuschkin, Suchir Balaji, Valerie Balcom, Paul Baltescu, Haiming Bao, Mohammad Bavarian, Jeff Belgum, Irwan Bello, Jake Berdine, Gabriel Bernadett-Shapiro, Christopher Berner, Lenny Bogdonoff, Oleg Boiko, Madelaine Boyd, Anna-Luisa Brakman, Greg Brockman, Tim Brooks, Miles Brundage, Kevin Button, Trevor Cai, Rosie Campbell, Andrew Cann, Brittany Carey, Chelsea Carlson, Rory Carmichael, Brooke Chan, Che Chang, Fotis Chantzis, Derek Chen, Sully Chen, Ruby Chen, Jason Chen, Mark Chen, Ben Chess, Chester Cho, Casey Chu, Hyung~Won Chung, Dave Cummings, Jeremiah Currier, Yunxing Dai, Cory Decareaux, Thomas Degry, Noah Deutsch, Damien Deville, Arka Dhar, David Dohan, Steve Dowling, Sheila Dunning, Adrien Ecoffet, Atty Eleti, Tyna Eloundou, David Farhi, Liam Fedus, Niko Felix, Simón~Posada Fishman, Juston Forte, Isabella Fulford, Leo
  Gao, Elie Georges, Christian Gibson, Vik Goel, Tarun Gogineni, Gabriel Goh, Rapha Gontijo-Lopes, Jonathan Gordon, Morgan Grafstein, Scott Gray, Ryan Greene, Joshua Gross, Shixiang~Shane Gu, Yufei Guo, Chris Hallacy, Jesse Han, Jeff Harris, Yuchen He, Mike Heaton, Johannes Heidecke, Chris Hesse, Alan Hickey, Wade Hickey, Peter Hoeschele, Brandon Houghton, Kenny Hsu, Shengli Hu, Xin Hu, Joost Huizinga, Shantanu Jain, Shawn Jain, Joanne Jang, Angela Jiang, Roger Jiang, Haozhun Jin, Denny Jin, Shino Jomoto, Billie Jonn, Heewoo Jun, Tomer Kaftan, Łukasz Kaiser, Ali Kamali, Ingmar Kanitscheider, Nitish~Shirish Keskar, Tabarak Khan, Logan Kilpatrick, Jong~Wook Kim, Christina Kim, Yongjik Kim, Jan~Hendrik Kirchner, Jamie Kiros, Matt Knight, Daniel Kokotajlo, Łukasz Kondraciuk, Andrew Kondrich, Aris Konstantinidis, Kyle Kosic, Gretchen Krueger, Vishal Kuo, Michael Lampe, Ikai Lan, Teddy Lee, Jan Leike, Jade Leung, Daniel Levy, Chak~Ming Li, Rachel Lim, Molly Lin, Stephanie Lin, Mateusz Litwin, Theresa Lopez, Ryan
  Lowe, Patricia Lue, Anna Makanju, Kim Malfacini, Sam Manning, Todor Markov, Yaniv Markovski, Bianca Martin, Katie Mayer, Andrew Mayne, Bob McGrew, Scott~Mayer McKinney, Christine McLeavey, Paul McMillan, Jake McNeil, David Medina, Aalok Mehta, Jacob Menick, Luke Metz, Andrey Mishchenko, Pamela Mishkin, Vinnie Monaco, Evan Morikawa, Daniel Mossing, Tong Mu, Mira Murati, Oleg Murk, David Mély, Ashvin Nair, Reiichiro Nakano, Rajeev Nayak, Arvind Neelakantan, Richard Ngo, Hyeonwoo Noh, Long Ouyang, Cullen O'Keefe, Jakub Pachocki, Alex Paino, Joe Palermo, Ashley Pantuliano, Giambattista Parascandolo, Joel Parish, Emy Parparita, Alex Passos, Mikhail Pavlov, Andrew Peng, Adam Perelman, Filipe de Avila Belbute~Peres, Michael Petrov, Henrique~Ponde de Oliveira~Pinto, Michael, Pokorny, Michelle Pokrass, Vitchyr~H. Pong, Tolly Powell, Alethea Power, Boris Power, Elizabeth Proehl, Raul Puri, Alec Radford, Jack Rae, Aditya Ramesh, Cameron Raymond, Francis Real, Kendra Rimbach, Carl Ross, Bob Rotsted, Henri Roussez,
  Nick Ryder, Mario Saltarelli, Ted Sanders, Shibani Santurkar, Girish Sastry, Heather Schmidt, David Schnurr, John Schulman, Daniel Selsam, Kyla Sheppard, Toki Sherbakov, Jessica Shieh, Sarah Shoker, Pranav Shyam, Szymon Sidor, Eric Sigler, Maddie Simens, Jordan Sitkin, Katarina Slama, Ian Sohl, Benjamin Sokolowsky, Yang Song, Natalie Staudacher, Felipe~Petroski Such, Natalie Summers, Ilya Sutskever, Jie Tang, Nikolas Tezak, Madeleine~B. Thompson, Phil Tillet, Amin Tootoonchian, Elizabeth Tseng, Preston Tuggle, Nick Turley, Jerry Tworek, Juan Felipe~Cerón Uribe, Andrea Vallone, Arun Vijayvergiya, Chelsea Voss, Carroll Wainwright, Justin~Jay Wang, Alvin Wang, Ben Wang, Jonathan Ward, Jason Wei, CJ Weinmann, Akila Welihinda, Peter Welinder, Jiayi Weng, Lilian Weng, Matt Wiethoff, Dave Willner, Clemens Winter, Samuel Wolrich, Hannah Wong, Lauren Workman, Sherwin Wu, Jeff Wu, Michael Wu, Kai Xiao, Tao Xu, Sarah Yoo, Kevin Yu, Qiming Yuan, Wojciech Zaremba, Rowan Zellers, Chong Zhang, Marvin Zhang, Shengjia
  Zhao, Tianhao Zheng, Juntang Zhuang, William Zhuk, and Barret Zoph.
\newblock Gpt-4 technical report, 2024.

\bibitem[Panickssery et~al.(2024)Panickssery, Bowman, and Feng]{panickssery2404llm}
Arjun Panickssery, Samuel~R. Bowman, and Shi Feng.
\newblock Llm evaluators recognize and favor their own generations.
\newblock \emph{ArXiv}, abs/2404.13076, 2024.

\bibitem[Papineni et~al.(2002)Papineni, Roukos, Ward, and Zhu]{bleu}
Kishore Papineni, Salim Roukos, Todd Ward, and Wei-Jing Zhu.
\newblock Bleu: a method for automatic evaluation of machine translation.
\newblock In \emph{Proceedings of the 40th annual meeting of the Association for Computational Linguistics}, pages 311--318, 2002.

\bibitem[Parcalabescu et~al.(2021)Parcalabescu, Cafagna, Muradjan, Frank, Calixto, and Gatt]{VALSE}
Letitia Parcalabescu, Michele Cafagna, Lilitta Muradjan, Anette Frank, Iacer Calixto, and Albert Gatt.
\newblock Valse: A task-independent benchmark for vision and language models centered on linguistic phenomena.
\newblock \emph{arXiv preprint arXiv:2112.07566}, 2021.

\bibitem[Peng et~al.(2023)Peng, Wang, Dong, Hao, Huang, Ma, and Wei]{peng2023kosmos}
Zhiliang Peng, Wenhui Wang, Li Dong, Yaru Hao, Shaohan Huang, Shuming Ma, and Furu Wei.
\newblock Kosmos-2: Grounding multimodal large language models to the world.
\newblock \emph{arXiv preprint arXiv:2306.14824}, 2023.

\bibitem[Qi et~al.(2024)Qi, Fang, Sun, Wu, Wu, Wang, Lin, and Zhao]{GPT4Point}
Zhangyang Qi, Ye Fang, Zeyi Sun, Xiaoyang Wu, Tong Wu, Jiaqi Wang, Dahua Lin, and Hengshuang Zhao.
\newblock Gpt4point: A unified framework for point-language understanding and generation.
\newblock In \emph{CVPR}, 2024.

\bibitem[Radford et~al.(2021)Radford, Kim, Hallacy, Ramesh, Goh, Agarwal, Sastry, Askell, Mishkin, Clark, et~al.]{radford2021learning}
Alec Radford, Jong~Wook Kim, Chris Hallacy, Aditya Ramesh, Gabriel Goh, Sandhini Agarwal, Girish Sastry, Amanda Askell, Pamela Mishkin, Jack Clark, et~al.
\newblock Learning transferable visual models from natural language supervision.
\newblock In \emph{International conference on machine learning}, pages 8748--8763. PMLR, 2021.

\bibitem[Reid et~al.(2024)Reid, Savinov, Teplyashin, Lepikhin, Lillicrap, Alayrac, Soricut, Lazaridou, Firat, Schrittwieser, et~al.]{gemini2024}
Machel Reid, Nikolay Savinov, Denis Teplyashin, Dmitry Lepikhin, Timothy Lillicrap, Jean-baptiste Alayrac, Radu Soricut, Angeliki Lazaridou, Orhan Firat, Julian Schrittwieser, et~al.
\newblock Gemini 1.5: Unlocking multimodal understanding across millions of tokens of context.
\newblock \emph{arXiv preprint arXiv:2403.05530}, 2024.

\bibitem[Reimers and Gurevych(2019)]{sentencebert}
Nils Reimers and Iryna Gurevych.
\newblock Sentence-bert: Sentence embeddings using siamese bert-networks.
\newblock \emph{arXiv preprint arXiv:1908.10084}, 2019.

\bibitem[Ren et~al.(2023)Ren, Pan, Tang, Zhang, Cao, Zeng, and Liu]{DreamGaussian4D}
Jiawei Ren, Liang Pan, Jiaxiang Tang, Chi Zhang, Ang Cao, Gang Zeng, and Ziwei Liu.
\newblock Dreamgaussian4d: Generative 4d gaussian splatting.
\newblock \emph{arXiv preprint arXiv:2312.17142}, 2023.

\bibitem[Ren et~al.(2024)Ren, Yao, Li, Sun, and Hou]{TimeIT}
Shuhuai Ren, Linli Yao, Shicheng Li, Xu Sun, and Lu Hou.
\newblock Timechat: A time-sensitive multimodal large language model for long video understanding.
\newblock In \emph{Proceedings of the IEEE/CVF Conference on Computer Vision and Pattern Recognition}, pages 14313--14323, 2024.

\bibitem[Santos et~al.(2021)Santos, Colombini, and Avila]{santos2021cider}
Gabriel Oliveira~dos Santos, Esther~Luna Colombini, and Sandra Avila.
\newblock Cider-r: Robust consensus-based image description evaluation.
\newblock \emph{arXiv preprint arXiv:2109.13701}, 2021.

\bibitem[Sarto et~al.(2023)Sarto, Barraco, Cornia, Baraldi, and Cucchiara]{sarto2023positive}
Sara Sarto, Manuele Barraco, Marcella Cornia, Lorenzo Baraldi, and Rita Cucchiara.
\newblock Positive-augmented contrastive learning for image and video captioning evaluation.
\newblock In \emph{Proceedings of the IEEE/CVF conference on computer vision and pattern recognition}, pages 6914--6924, 2023.

\bibitem[Shi et~al.(2022)Shi, Yang, Xu, Yuan, Li, Hu, and Zha]{shi2022emscore}
Yaya Shi, Xu Yang, Haiyang Xu, Chunfeng Yuan, Bing Li, Weiming Hu, and Zheng-Jun Zha.
\newblock Emscore: Evaluating video captioning via coarse-grained and fine-grained embedding matching.
\newblock In \emph{Proceedings of the IEEE/CVF conference on computer vision and pattern recognition}, pages 17929--17938, 2022.

\bibitem[Shu et~al.(2024)Shu, Zhang, Liu, Qin, Zhou, Huang, and Zhao]{shu2024video}
Yan Shu, Peitian Zhang, Zheng Liu, Minghao Qin, Junjie Zhou, Tiejun Huang, and Bo Zhao.
\newblock Video-xl: Extra-long vision language model for hour-scale video understanding.
\newblock \emph{arXiv preprint arXiv:2409.14485}, 2024.

\bibitem[Snover et~al.(2006)Snover, Dorr, Schwartz, Micciulla, and Makhoul]{snover2006study}
Matthew Snover, Bonnie Dorr, Richard Schwartz, Linnea Micciulla, and John Makhoul.
\newblock A study of translation edit rate with targeted human annotation.
\newblock In \emph{Proceedings of the 7th Conference of the Association for Machine Translation in the Americas: Technical Papers}, pages 223--231, 2006.

\bibitem[Song et~al.(2023)Song, Chai, Wang, Zhang, Zhou, Wu, Guo, Ye, Lu, Hwang, et~al.]{moviechat}
Enxin Song, Wenhao Chai, Guanhong Wang, Yucheng Zhang, Haoyang Zhou, Feiyang Wu, Xun Guo, Tian Ye, Yan Lu, Jenq-Neng Hwang, et~al.
\newblock Moviechat: From dense token to sparse memory for long video understanding.
\newblock \emph{arXiv preprint arXiv:2307.16449}, 2023.

\bibitem[Team et~al.(2023)Team, Anil, Borgeaud, Wu, Alayrac, Yu, Soricut, Schalkwyk, Dai, Hauth, et~al.]{gemini2023}
Gemini Team, Rohan Anil, Sebastian Borgeaud, Yonghui Wu, Jean-Baptiste Alayrac, Jiahui Yu, Radu Soricut, Johan Schalkwyk, Andrew~M Dai, Anja Hauth, et~al.
\newblock Gemini: a family of highly capable multimodal models.
\newblock \emph{arXiv preprint arXiv:2312.11805}, 2023.

\bibitem[Team et~al.(2024)Team, Ormazabal, Zheng, d'Autume, Yogatama, Fu, Ong, Chen, Lamprecht, Pham, et~al.]{team2024reka}
Reka Team, Aitor Ormazabal, Che Zheng, Cyprien de~Masson d'Autume, Dani Yogatama, Deyu Fu, Donovan Ong, Eric Chen, Eugenie Lamprecht, Hai Pham, et~al.
\newblock Reka core, flash, and edge: A series of powerful multimodal language models.
\newblock \emph{arXiv preprint arXiv:2404.12387}, 2024.

\bibitem[Thrush et~al.(2022)Thrush, Jiang, Bartolo, Singh, Williams, Kiela, and Ross]{Winoground}
Tristan Thrush, Ryan Jiang, Max Bartolo, Amanpreet Singh, Adina Williams, Douwe Kiela, and Candace Ross.
\newblock Winoground: Probing vision and language models for visio-linguistic compositionality.
\newblock In \emph{Proceedings of the IEEE/CVF Conference on Computer Vision and Pattern Recognition}, pages 5238--5248, 2022.

\bibitem[Tong et~al.(2024{\natexlab{a}})Tong, Brown, Wu, Woo, Middepogu, Akula, Yang, Yang, Iyer, Pan, et~al.]{cambrian1}
Shengbang Tong, Ellis Brown, Penghao Wu, Sanghyun Woo, Manoj Middepogu, Sai~Charitha Akula, Jihan Yang, Shusheng Yang, Adithya Iyer, Xichen Pan, et~al.
\newblock Cambrian-1: A fully open, vision-centric exploration of multimodal llms.
\newblock \emph{arXiv preprint arXiv:2406.16860}, 2024{\natexlab{a}}.

\bibitem[Tong et~al.(2024{\natexlab{b}})Tong, Liu, Zhai, Ma, LeCun, and Xie]{MMVP}
Shengbang Tong, Zhuang Liu, Yuexiang Zhai, Yi Ma, Yann LeCun, and Saining Xie.
\newblock Eyes wide shut? exploring the visual shortcomings of multimodal llms.
\newblock In \emph{Proceedings of the IEEE/CVF Conference on Computer Vision and Pattern Recognition}, pages 9568--9578, 2024{\natexlab{b}}.

\bibitem[Touvron et~al.(2023{\natexlab{a}})Touvron, Lavril, Izacard, Martinet, Lachaux, Lacroix, Rozi{\`e}re, Goyal, Hambro, Azhar, et~al.]{llama}
Hugo Touvron, Thibaut Lavril, Gautier Izacard, Xavier Martinet, Marie-Anne Lachaux, Timoth{\'e}e Lacroix, Baptiste Rozi{\`e}re, Naman Goyal, Eric Hambro, Faisal Azhar, et~al.
\newblock Llama: open and efficient foundation language models. arxiv.
\newblock \emph{arXiv preprint arXiv:2302.13971}, 2023{\natexlab{a}}.

\bibitem[Touvron et~al.(2023{\natexlab{b}})Touvron, Martin, Stone, Albert, Almahairi, Babaei, Bashlykov, Batra, Bhargava, Bhosale, et~al.]{llama2}
Hugo Touvron, Louis Martin, Kevin Stone, Peter Albert, Amjad Almahairi, Yasmine Babaei, Nikolay Bashlykov, Soumya Batra, Prajjwal Bhargava, Shruti Bhosale, et~al.
\newblock Llama 2: Open foundation and fine-tuned chat models.
\newblock \emph{arXiv preprint arXiv:2307.09288}, 2023{\natexlab{b}}.

\bibitem[Vedantam et~al.(2015)Vedantam, Lawrence~Zitnick, and Parikh]{cider}
Ramakrishna Vedantam, C Lawrence~Zitnick, and Devi Parikh.
\newblock Cider: Consensus-based image description evaluation.
\newblock In \emph{Proceedings of the IEEE conference on computer vision and pattern recognition}, pages 4566--4575, 2015.

\bibitem[Wang et~al.(2023{\natexlab{a}})Wang, Ge, Ding, Kankanhalli, and Shan]{GVT-bench}
Guangzhi Wang, Yixiao Ge, Xiaohan Ding, Mohan Kankanhalli, and Ying Shan.
\newblock What makes for good visual tokenizers for large language models?
\newblock \emph{arXiv preprint arXiv:2305.12223}, 2023{\natexlab{a}}.

\bibitem[Wang et~al.(2024{\natexlab{a}})Wang, Bai, Tan, Wang, Fan, Bai, Chen, Liu, Wang, Ge, et~al.]{qwen2_vl}
Peng Wang, Shuai Bai, Sinan Tan, Shijie Wang, Zhihao Fan, Jinze Bai, Keqin Chen, Xuejing Liu, Jialin Wang, Wenbin Ge, et~al.
\newblock Qwen2-vl: Enhancing vision-language model's perception of the world at any resolution.
\newblock \emph{arXiv preprint arXiv:2409.12191}, 2024{\natexlab{a}}.

\bibitem[Wang et~al.(2023{\natexlab{b}})Wang, Lv, Yu, Hong, Qi, Wang, Ji, Yang, Zhao, Song, et~al.]{wang2023cogvlm}
Weihan Wang, Qingsong Lv, Wenmeng Yu, Wenyi Hong, Ji Qi, Yan Wang, Junhui Ji, Zhuoyi Yang, Lei Zhao, Xixuan Song, et~al.
\newblock Cogvlm: Visual expert for pretrained language models.
\newblock \emph{arXiv preprint arXiv:2311.03079}, 2023{\natexlab{b}}.

\bibitem[Wang et~al.(2024{\natexlab{b}})Wang, Ding, Zeng, Zhou, Shen, Luo, and Tao]{HR-Bench}
Wenbin Wang, Liang Ding, Minyan Zeng, Xiabin Zhou, Li Shen, Yong Luo, and Dacheng Tao.
\newblock Divide, conquer and combine: A training-free framework for high-resolution image perception in multimodal large language models.
\newblock \emph{arXiv preprint arXiv:2408.15556}, 2024{\natexlab{b}}.

\bibitem[Wang et~al.(2019)Wang, Wu, Chen, Li, Wang, and Wang]{wang2019vatex}
Xin Wang, Jiawei Wu, Junkun Chen, Lei Li, Yuan-Fang Wang, and William~Yang Wang.
\newblock Vatex: A large-scale, high-quality multilingual dataset for video-and-language research.
\newblock In \emph{Proceedings of the IEEE/CVF international conference on computer vision}, pages 4581--4591, 2019.

\bibitem[Wang et~al.(2024{\natexlab{c}})Wang, Song, Chen, Zhang, and Wang]{wang2024longllava}
Xidong Wang, Dingjie Song, Shunian Chen, Chen Zhang, and Benyou Wang.
\newblock Longllava: Scaling multi-modal llms to 1000 images efficiently via a hybrid architecture.
\newblock \emph{arXiv preprint arXiv:2409.02889}, 2024{\natexlab{c}}.

\bibitem[Wu et~al.(2024)Wu, Yi, Fang, Xie, Zhang, Wei, Liu, Tian, and Wang]{4DGS}
Guanjun Wu, Taoran Yi, Jiemin Fang, Lingxi Xie, Xiaopeng Zhang, Wei Wei, Wenyu Liu, Qi Tian, and Xinggang Wang.
\newblock 4d gaussian splatting for real-time dynamic scene rendering.
\newblock In \emph{Proceedings of the IEEE/CVF Conference on Computer Vision and Pattern Recognition (CVPR)}, pages 20310--20320, 2024.

\bibitem[Xu et~al.(2016)Xu, Mei, Yao, and Rui]{xu2016msr}
Jun Xu, Tao Mei, Ting Yao, and Yong Rui.
\newblock Msr-vtt: A large video description dataset for bridging video and language.
\newblock In \emph{Proceedings of the IEEE conference on computer vision and pattern recognition}, pages 5288--5296, 2016.

\bibitem[Xu et~al.(2023)Xu, Shao, Zhang, Gao, Liu, Lei, Meng, Huang, Qiao, and Luo]{MME}
Peng Xu, Wenqi Shao, Kaipeng Zhang, Peng Gao, Shuo Liu, Meng Lei, Fanqing Meng, Siyuan Huang, Yu Qiao, and Ping Luo.
\newblock Lvlm-ehub: A comprehensive evaluation benchmark for large vision-language models.
\newblock \emph{arXiv preprint arXiv:2306.09265}, 2023.

\bibitem[Xue et~al.(2024)Xue, Chen, Li, Hu, Zhu, Li, Fang, Tang, Yang, Liu, et~al.]{xue2024longvila}
Fuzhao Xue, Yukang Chen, Dacheng Li, Qinghao Hu, Ligeng Zhu, Xiuyu Li, Yunhao Fang, Haotian Tang, Shang Yang, Zhijian Liu, et~al.
\newblock Longvila: Scaling long-context visual language models for long videos.
\newblock \emph{arXiv preprint arXiv:2408.10188}, 2024.

\bibitem[Yang et~al.(2024)Yang, Yang, Zhang, Hui, Zheng, Yu, Li, Liu, Huang, Dong, Wei, Lin, Yang, Tu, Zhang, Yang, Yang, Zhou, Lin, Dang, Lu, Bao, Yang, Yu, Li, Xue, Zhang, Zhu, Men, Lin, Li, Xia, Ren, Ren, Fan, Su, Zhang, Wan, Liu, Cui, Zhang, Qiu, Quan, and Wang]{qwen25}
Qwen~An Yang, Baosong Yang, Beichen Zhang, Binyuan Hui, Bo Zheng, Bowen Yu, Chengyuan Li, Dayiheng Liu, Fei Huang, Guanting Dong, Haoran Wei, Huan Lin, Jian Yang, Jianhong Tu, Jianwei Zhang, Jianxin Yang, Jiaxin Yang, Jingren Zhou, Junyang Lin, Kai Dang, Keming Lu, Keqin Bao, Kexin Yang, Le Yu, Mei Li, Mingfeng Xue, Pei Zhang, Qin Zhu, Rui Men, Runji Lin, Tianhao Li, Tingyu Xia, Xingzhang Ren, Xuancheng Ren, Yang Fan, Yang Su, Yi-Chao Zhang, Yunyang Wan, Yuqi Liu, Zeyu Cui, Zhenru Zhang, Zihan Qiu, Shanghaoran Quan, and Zekun Wang.
\newblock Qwen2.5 technical report.
\newblock \emph{ArXiv}, abs/2412.15115, 2024.

\bibitem[Yao et~al.(2024)Yao, Yu, Zhang, Wang, Cui, Zhu, Cai, Li, Zhao, He, et~al.]{yao2024minicpm}
Yuan Yao, Tianyu Yu, Ao Zhang, Chongyi Wang, Junbo Cui, Hongji Zhu, Tianchi Cai, Haoyu Li, Weilin Zhao, Zhihui He, et~al.
\newblock Minicpm-v: A gpt-4v level mllm on your phone.
\newblock \emph{arXiv preprint arXiv:2408.01800}, 2024.

\bibitem[Ye et~al.(2024)Ye, Xu, Liu, Hu, Yan, Qian, Zhang, Huang, and Zhou]{ye2024mplug}
Jiabo Ye, Haiyang Xu, Haowei Liu, Anwen Hu, Ming Yan, Qi Qian, Ji Zhang, Fei Huang, and Jingren Zhou.
\newblock mplug-owl3: Towards long image-sequence understanding in multi-modal large language models.
\newblock \emph{arXiv preprint arXiv:2408.04840}, 2024.

\bibitem[Ye et~al.(2023)Ye, Xu, Xu, Ye, Yan, Zhou, Wang, Hu, Shi, Shi, et~al.]{OwlEval}
Qinghao Ye, Haiyang Xu, Guohai Xu, Jiabo Ye, Ming Yan, Yiyang Zhou, Junyang Wang, Anwen Hu, Pengcheng Shi, Yaya Shi, et~al.
\newblock mplug-owl: Modularization empowers large language models with multimodality.
\newblock \emph{arXiv preprint arXiv:2304.14178}, 2023.

\bibitem[Yu et~al.(2024)Yu, Yang, Li, Wang, Lin, Liu, Wang, and Wang]{Mm-vet}
Weihao Yu, Zhengyuan Yang, Linjie Li, Jianfeng Wang, Kevin Lin, Zicheng Liu, Xinchao Wang, and Lijuan Wang.
\newblock Mm-vet: Evaluating large multimodal models for integrated capabilities.
\newblock In \emph{International conference on machine learning}. PMLR, 2024.

\bibitem[Yue et~al.(2024)Yue, Ni, Zhang, Zheng, Liu, Zhang, Stevens, Jiang, Ren, Sun, et~al.]{mmmu}
Xiang Yue, Yuansheng Ni, Kai Zhang, Tianyu Zheng, Ruoqi Liu, Ge Zhang, Samuel Stevens, Dongfu Jiang, Weiming Ren, Yuxuan Sun, et~al.
\newblock Mmmu: A massive multi-discipline multimodal understanding and reasoning benchmark for expert agi.
\newblock In \emph{Proceedings of the IEEE/CVF Conference on Computer Vision and Pattern Recognition}, pages 9556--9567, 2024.

\bibitem[Zhang et~al.(2023)Zhang, Li, and Bing]{videollama}
Hang Zhang, Xin Li, and Lidong Bing.
\newblock Video-llama: An instruction-tuned audio-visual language model for video understanding.
\newblock \emph{arXiv preprint arXiv:2306.02858}, 2023.

\bibitem[Zhang et~al.(2024{\natexlab{a}})Zhang, Chen, Wang, Liu, Wang, and Qiao]{4diffusion}
Haiyu Zhang, Xinyuan Chen, Yaohui Wang, Xihui Liu, Yunhong Wang, and Yu Qiao.
\newblock 4diffusion: Multi-view video diffusion model for 4d generation.
\newblock \emph{arXiv preprint arXiv:2405.20674}, 2024{\natexlab{a}}.

\bibitem[Zhang et~al.(2024{\natexlab{b}})Zhang, Dong, Zang, Cao, Qian, Chen, Guo, Duan, Wang, Ouyang, et~al.]{zhang2024internlm}
Pan Zhang, Xiaoyi Dong, Yuhang Zang, Yuhang Cao, Rui Qian, Lin Chen, Qipeng Guo, Haodong Duan, Bin Wang, Linke Ouyang, et~al.
\newblock Internlm-xcomposer-2.5: A versatile large vision language model supporting long-contextual input and output.
\newblock \emph{arXiv preprint arXiv:2407.03320}, 2024{\natexlab{b}}.

\bibitem[Zhang et~al.(2024{\natexlab{c}})Zhang, Zhang, Li, Zeng, Yang, Zhang, Wang, Tan, Li, and Liu]{zhang2024long}
Peiyuan Zhang, Kaichen Zhang, Bo Li, Guangtao Zeng, Jingkang Yang, Yuanhan Zhang, Ziyue Wang, Haoran Tan, Chunyuan Li, and Ziwei Liu.
\newblock Long context transfer from language to vision.
\newblock \emph{arXiv preprint arXiv:2406.16852}, 2024{\natexlab{c}}.

\bibitem[Zhang et~al.(2019)Zhang, Kishore, Wu, Weinberger, and Artzi]{bertscore}
Tianyi Zhang, Varsha Kishore, Felix Wu, Kilian~Q Weinberger, and Yoav Artzi.
\newblock Bertscore: Evaluating text generation with bert.
\newblock \emph{arXiv preprint arXiv:1904.09675}, 2019.

\bibitem[Zhang et~al.(2024{\natexlab{d}})Zhang, Wu, Li, Li, Ma, Liu, and Li]{llava_video}
Yuanhan Zhang, Jinming Wu, Wei Li, Bo Li, Zejun Ma, Ziwei Liu, and Chunyuan Li.
\newblock Video instruction tuning with synthetic data.
\newblock \emph{arXiv preprint arXiv:2410.02713}, 2024{\natexlab{d}}.

\bibitem[Zhang et~al.(2024{\natexlab{e}})Zhang, Wen, Fu, Wang, Zhang, Wang, and Jin]{zhang2024beyond}
Yi-Fan Zhang, Qingsong Wen, Chaoyou Fu, Xue Wang, Zhang Zhang, Liang Wang, and Rong Jin.
\newblock Beyond llava-hd: Diving into high-resolution large multimodal models.
\newblock \emph{arXiv preprint arXiv:2406.08487}, 2024{\natexlab{e}}.

\bibitem[Zhou et~al.(2024)Zhou, Shu, Zhao, Wu, Xiao, Yang, Xiong, Zhang, Huang, and Liu]{MLVU}
Junjie Zhou, Yan Shu, Bo Zhao, Boya Wu, Shitao Xiao, Xi Yang, Yongping Xiong, Bo Zhang, Tiejun Huang, and Zheng Liu.
\newblock Mlvu: A comprehensive benchmark for multi-task long video understanding.
\newblock \emph{arXiv preprint arXiv:2406.04264}, 2024.

\bibitem[Zhou et~al.(2018)Zhou, Xu, and Corso]{zhou2018towards}
Luowei Zhou, Chenliang Xu, and Jason Corso.
\newblock Towards automatic learning of procedures from web instructional videos.
\newblock In \emph{Proceedings of the AAAI Conference on Artificial Intelligence}, 2018.

\bibitem[Zhu et~al.(2023)Zhu, Chen, Shen, Li, and Elhoseiny]{minigpt}
Deyao Zhu, Jun Chen, Xiaoqian Shen, Xiang Li, and Mohamed Elhoseiny.
\newblock Minigpt-4: Enhancing vision-language understanding with advanced large language models.
\newblock \emph{arXiv preprint arXiv:2304.10592}, 2023.

\bibitem[Ziyu et~al.(2023)Ziyu, Xiaojian, Yixin, Zhidong, Siyuan, and Qing]{3dvista}
Zhu Ziyu, Ma Xiaojian, Chen Yixin, Deng Zhidong, Huang Siyuan, and Li Qing.
\newblock 3d-vista: Pre-trained transformer for 3d vision and text alignment.
\newblock In \emph{ICCV}, 2023.

\end{thebibliography}
}
\clearpage
\twocolumn[{%
\renewcommand\twocolumn[1][]{#1}%
\maketitlesupplementary
\begin{center}
    \centering
    \vspace{-20pt}
    \normalsize
    \label{fig:teaser}
\end{center}%
}]

\renewcommand{\thesection}{\Alph{section}}
\renewcommand{\thetable}{\Roman{table}}
\renewcommand{\thefigure}{\Roman{figure}}
\setcounter{section}{0}
\setcounter{table}{0}
\setcounter{figure}{0}
\startcontents
\printcontents{}{1}{}  % 只打印这部分的目录

\section{More related work}
\label{sec:more_retated_work}
\textbf{Benchmark datasets for image and video captioning.}
The development of image captioning has been driven by several influential datasets\cite{chen2015microsoft, agrawal2019nocaps, dong2024benchmarking}. COCO \cite{chen2015microsoft} stands as one of the most widely used benchmarks and covers diverse daily scenes and objects, making it a robust benchmark for evaluating captioning models.
The ground-truth captions provided by early benchmark datasets typically contain limited information. Yet, recent  MLLMs are capable of generating more detailed captions, making these datasets not challenging enough for evaluating modern models' capabilities of producing rich, nuanced descriptions that capture fine-grained visual details and complex relationships between objects.
To fill this gap, Dong et al.\cite{dong2024benchmarking} propose DetailCaps, a new benchmark featuring longer and more detailed captions annotated by human experts and powerful MLLMs like GPT-4V.
On the other hand, several datasets\cite{xu2016msr, chen2011collecting,krishna2017dense, zhou2018towards, wang2019vatex, lei2020tvr, han2023shot2story20k} have been established for 2D video captioning.
MSR-VTT\cite{xu2016msr} provides 20 descriptions per video clip for the open domain 2D video captioning.
% The MSVD dataset\cite{chen2011collecting} contains 18000 short videos with multilingual annotation.
ActivityNet Captions\cite{krishna2017dense} provide temporally localized multiple-sentence descriptions for video captioning.
For domain-specific applications, YouCook2\cite{zhou2018towards} presents task-oriented instructional cooking videos. 

\textbf{Reference-free captioning metrics.} 
We use reference-based metrics \cite{cider,bleu,meteor,rouge, kang2023noise, doddington2002automatic, santos2021cider, snover2006study} in the main paper.
Recently, reference-free caption metrics\cite{hessel2021clipscore,lee2024fleur,sarto2023positive,shi2022emscore,liang2024evqascore, kim2022mutual} has emerged in the image and video captioning metrics field.
Reference-free metrics eliminate the need for human-annotated references, reducing evaluation costs and effort.
They are also ideal for open-ended scenarios, accommodating multiple valid image descriptions and overcoming the limitations of reference-based methods that rely on potentially incomplete captions.
For example, CLIPScore\cite{hessel2021clipscore} uses CLIP embeddings to compute the similarity between generated captions and their associated visual content, offering a flexible way to assess captions in open-ended settings.

\section{More details about 4D-Bench}
\label{sec:4DBench}
\subsection{More details about 4D object representation}

We chose multi-view videos as the representation for 4D objects, as we found recent advanced MLLMs \cite{liu2024oryx,li2024aria,lin2024vila,ye2024mplug, cheng2024videollama, yao2024minicpm, shu2024video,fu2024vita,liu2024kangaroo, fei2024video, wang2024longllava, zhang2024long, xue2024longvila, zhang2024beyond, liu2025st, cambrian1, zhang2024internlm,dai2024nvlm, lin2024vila, team2024reka, peng2023kosmos, awadalla2023openflamingo, wang2023cogvlm, lin2023sphinx} are primarily designed to take texts and 2D images/videos as inputs.

We render the multi-view videos for 4D objects collected from Objaverse-XL\cite{objaversexl}.
For each 4D object, we render a 2D video from a single view up to 125 frames and utilize pixel change detection to identify motion within the 2D video, determining the frame indices for the start and end of the motion. Based on these indices, we render videos from 23 additional views, ensuring that all 24-view videos cover the identified motion frames.
The camera positions are evenly distributed around the normalized 4D object with slight jitters, the camera positions are chosen with a radius from 2.2m to 2.6m and a height from 0.8m to 1.2m.

\subsection{More details about CLIP-based data curation}

We propose a CLIP-based classifier to automatically select high-quality  4D objects, such that low-quality ones, such as oversimplified geometry, lack of texture, and poor aesthetic quality, are removed.

To build the training dataset, we manually annotate thousands of 4D objects into three categories: high quality, textureless, and low overall quality.
The "low overall quality" category typically refers to objects with significant deformation or portions that are largely outside the camera view.
After that, for each object, we choose the first frame of the video from the first view and its corresponding label to build the training dataset.
%We use this dataset to fine-tune the CLIP visual encoder by adding a linear layer as the classification head. 
We build the CLIP-based classifier by adding a linear layer as the classification head to fine-tune the CLIP visual encoder, and then use this dataset to fine-tune the classifier.

%During inference, we input eight images taken from the first frame of videos captured from different views of a single object.
During inference, we feed the first frame from 8 views of the 4D object into the CLIP-based classifier.
The final label of the object is determined through majority voting across the predictions made for these eight images.
Objects classified as high quality are retained, ensuring the dataset is highly usable.

\subsection{Additional statistical data analysis}
\begin{figure}[t]
    \centering
    \includegraphics[width=1.0\linewidth]{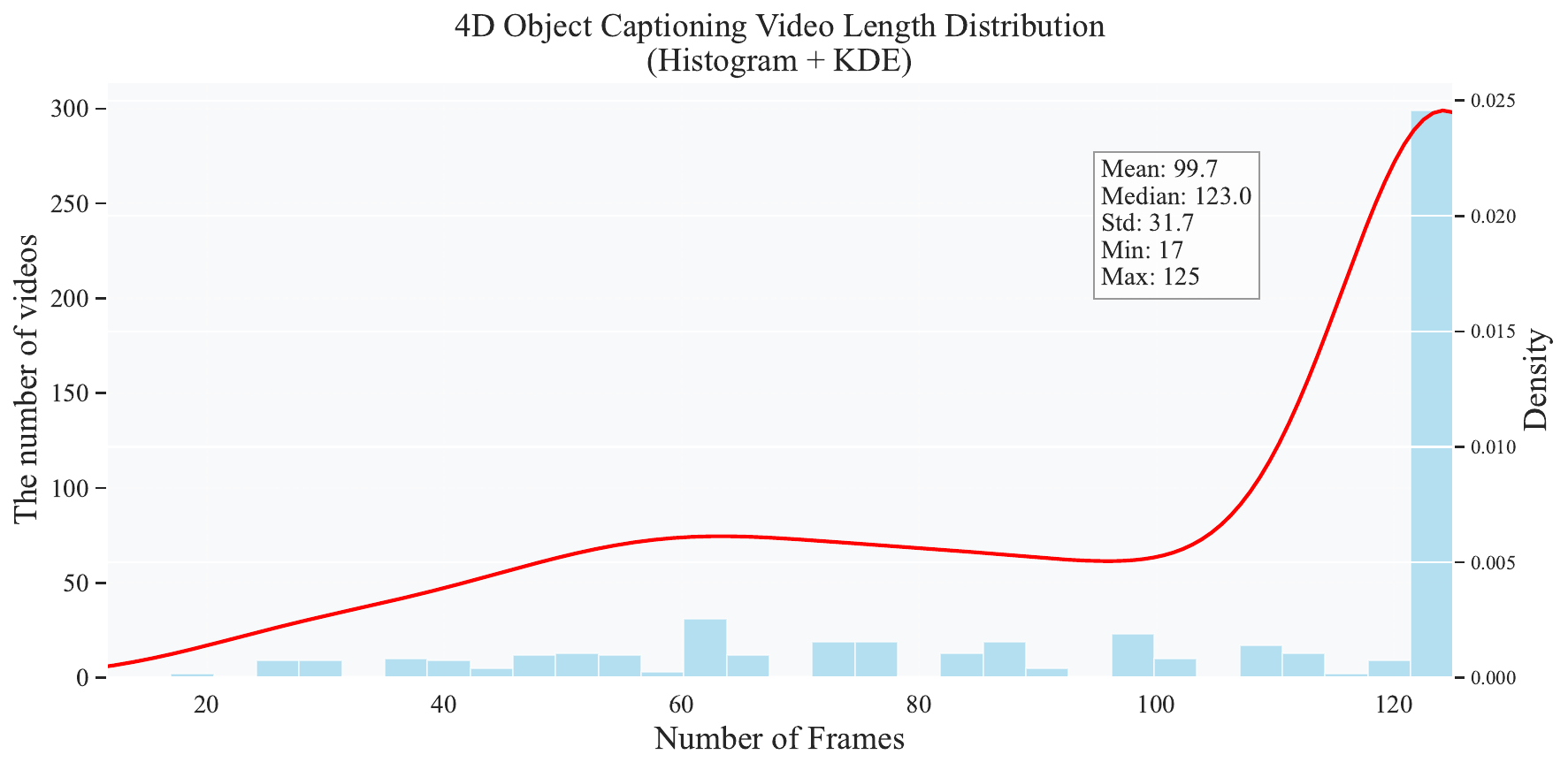}
    \caption{The frame-length distribution of multi-view videos used in the 4D object captioning task}
    \label{fig:caption_frame_distribution}
\end{figure}

\begin{figure}[hbt!]
    \centering
    \includegraphics[width=0.9\linewidth]{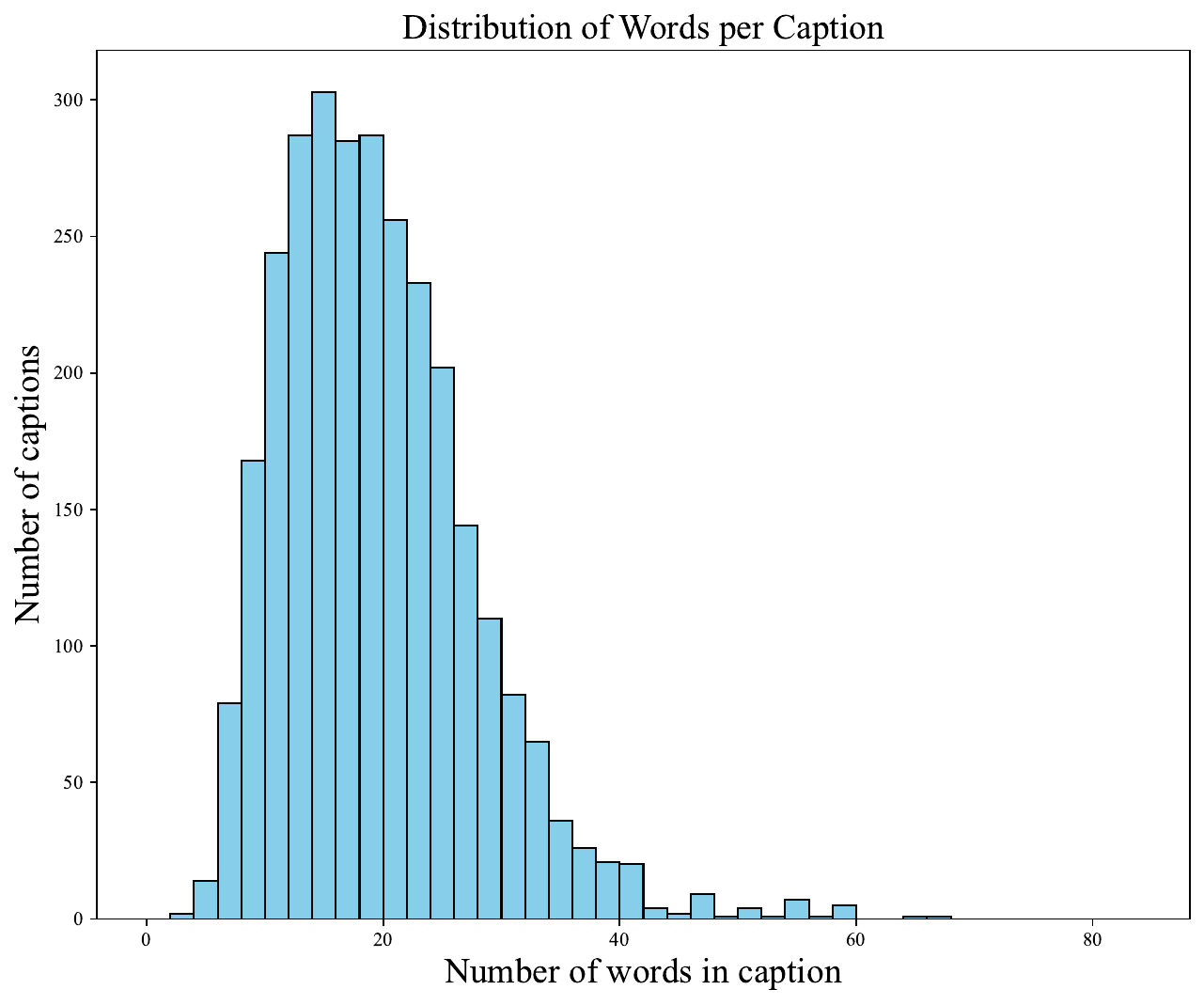}
    \caption{The length distribution of ground-truth captions used in the 4D object captioning task}
    \label{fig:caption_length}
\end{figure}

\textbf{4D object captioning statistics.} 
For the 4D object captioning task, we collected 580 4D objects, where each object is rendered into 24-view videos and has 5 human-annotated captions.
 \cref{fig:caption_frame_distribution} shows the frame-length distribution of multi-view videos, where the videos contain 99.73 frames per 4D object on average. 
The human-annotated captions have an average length of 19.05 words, and their length distribution is illustrated in \cref{fig:caption_length}.

\begin{figure}
    \centering
    \includegraphics[width=1.0\linewidth]{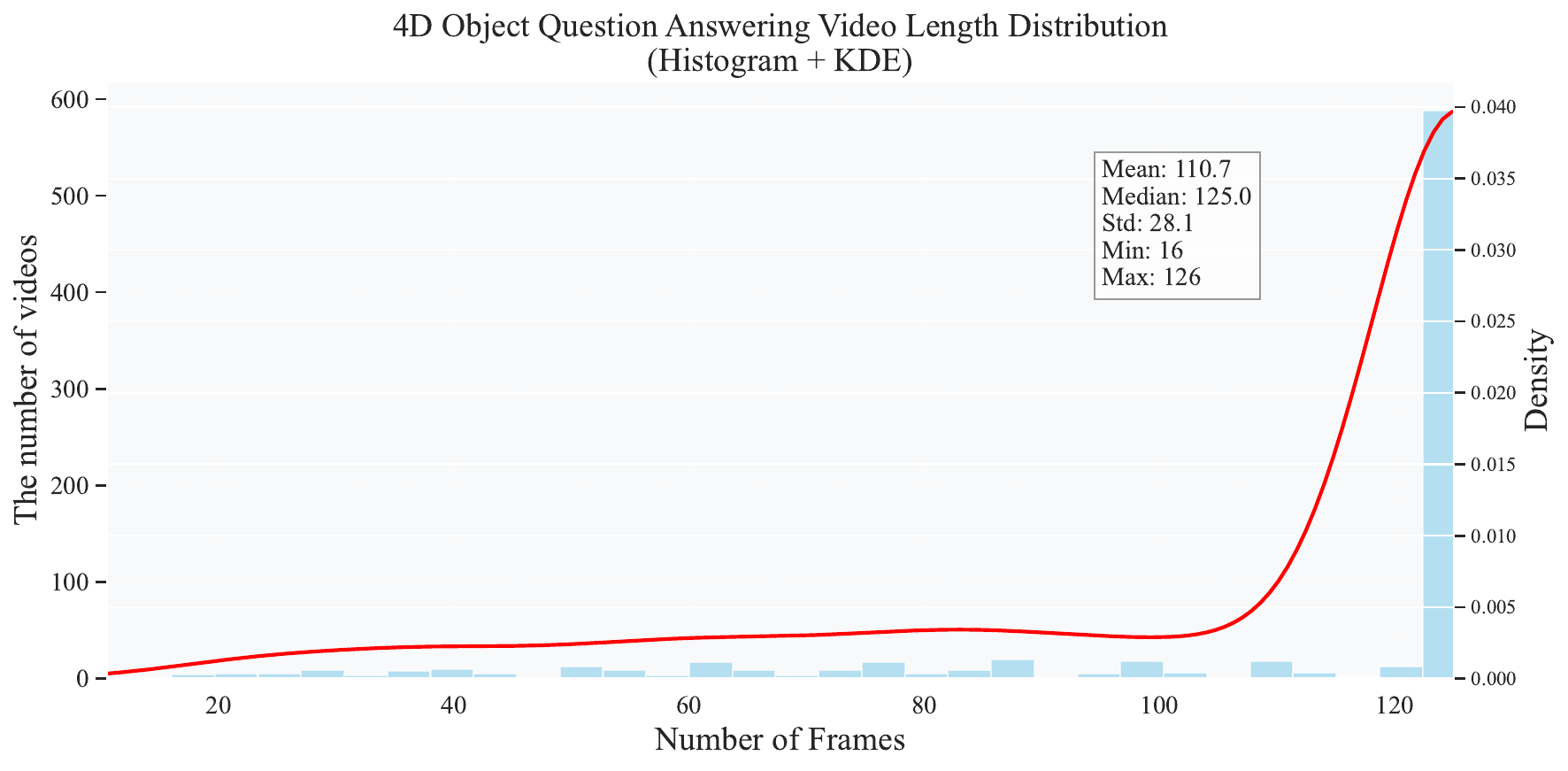}
    \caption{The frame-length distribution of multi-view videos used in the 4D object question answering task}
    \label{fig:vqa_frame_distribution}
\end{figure}

\begin{figure}[hbt!]
    \centering
    \includegraphics[width=1.1 \linewidth]{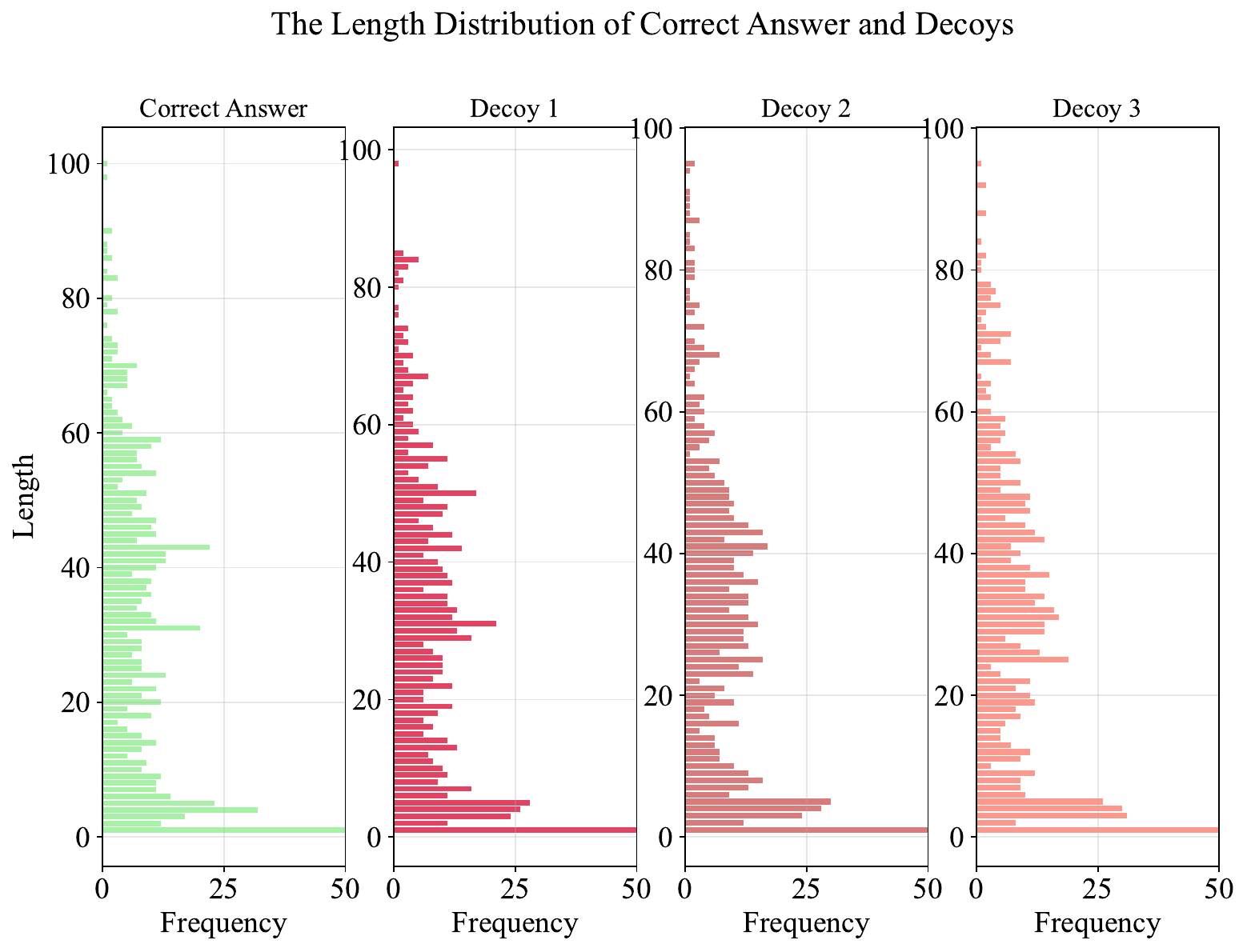}
    \caption{The truncated length distribution of correct answers and decoys used in 4D object question answering dataset}
    \label{fig:choice_length_distribution}
\end{figure}

\textbf{4D object question answering statistics.}
In the 4D object QA dataset, the multi-view videos contain an average of 101 frames per object, with the frame length distribution shown in \cref{fig:vqa_frame_distribution}.
\cref{fig:choice_length_distribution} illustrates that the length distributions of the answer options are roughly similar, avoiding bias caused by answer length. 
%preventing answer leakage due to length discrepancies.

\subsection{More details about evaluation metrics}
\begin{figure*}[hbt]
    \centering
    \includegraphics[width=0.9\textwidth]{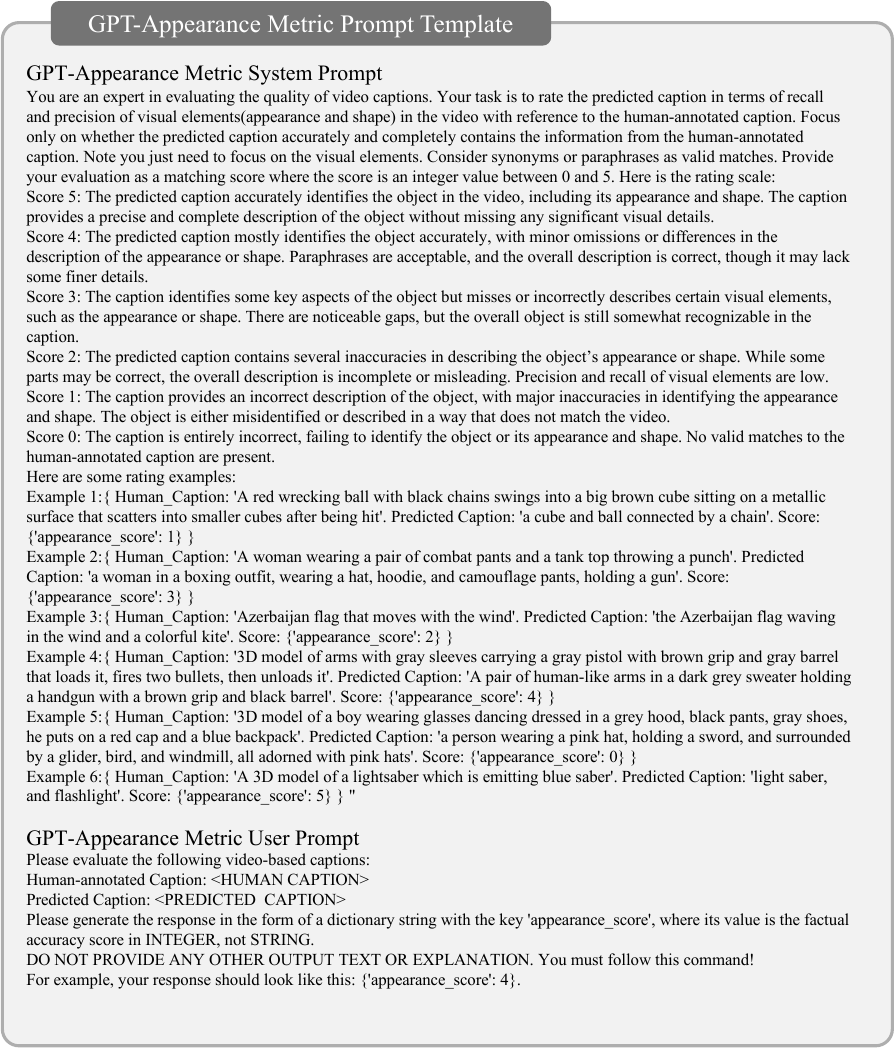}
    \caption{Prompt used in GPT-Appearance metric}
    \label{fig:gpt_4o_appearance_prompt}
\end{figure*}

\begin{figure*}[hbt]
    \centering
    \includegraphics[width=0.9\linewidth]{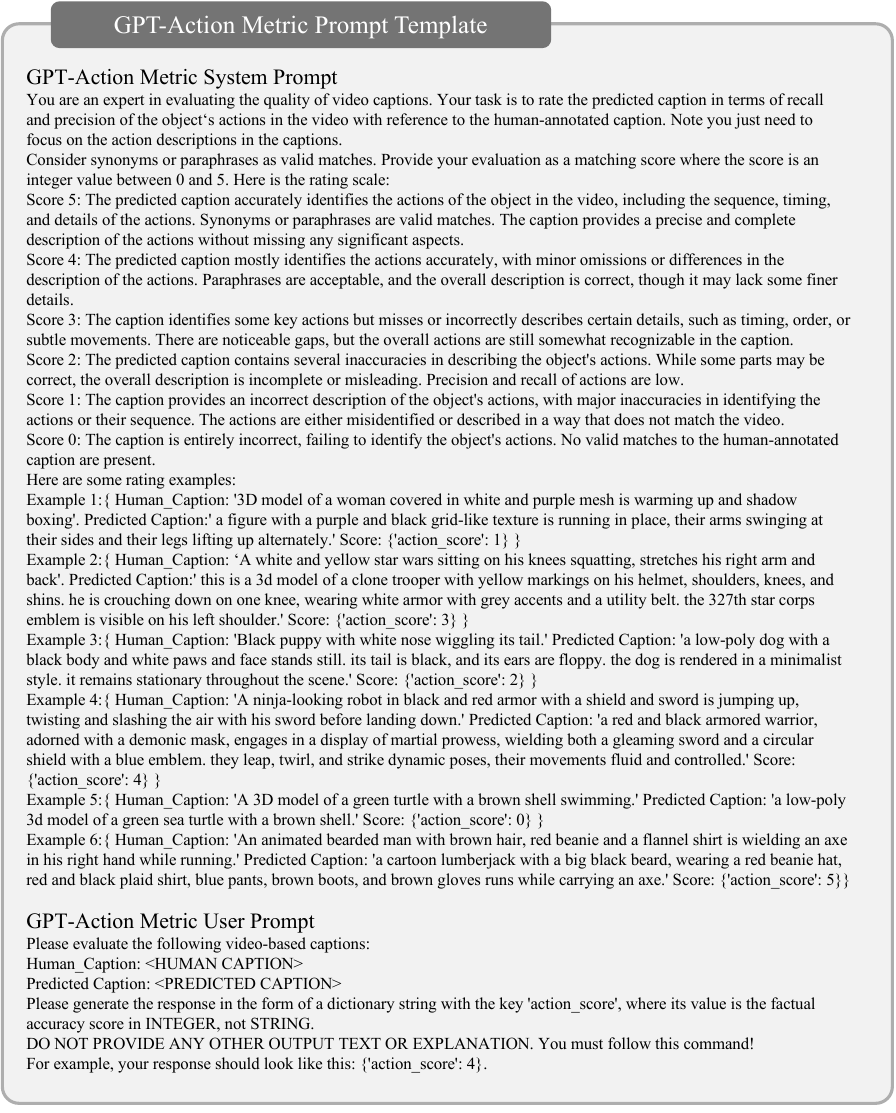}
    \caption{Prompt used in GPT-Action metric}
    \label{fig:gpt_4o_action_prompt}
\end{figure*}

\cref{fig:gpt_4o_appearance_prompt} and \cref{fig:gpt_4o_action_prompt} present the prompt template designed to guide GPT-4o in assessing the correspondence between generated and human-annotated captions in terms of appearance and action descriptions.
The prompt templates incorporate a comprehensive scoring rubric ranging from 0 to 5, where each score level is defined based on the accuracy and completeness of visual appearance/action descriptions.
% The prompt templates are also accompanied by representative examples demonstrating various scoring scenarios, facilitating consistent and systematic assessment of caption quality.
To ensure consistent evaluation, the template also provides carefully selected example pairs of human and machine-generated captions, along with their corresponding scores.

\section{More experimental details on 4D-Bench}
\label{sec:exp_details}
\subsection{More experimental details of 4D object captioning}
In the 4D object captioning experiments, all models adhere to a common function $C = M(V, t)$, where $V$, $t$, $M$ and $C$ denote the multi-view video input, text prompt (instruction), MLLM being tested, and generated caption respectively. The quality of generated captions is evaluated by computing various metric scores through comparison with human-annotated reference captions.

\begin{figure}[t]
    \centering
    \includegraphics[width=1.0\linewidth]{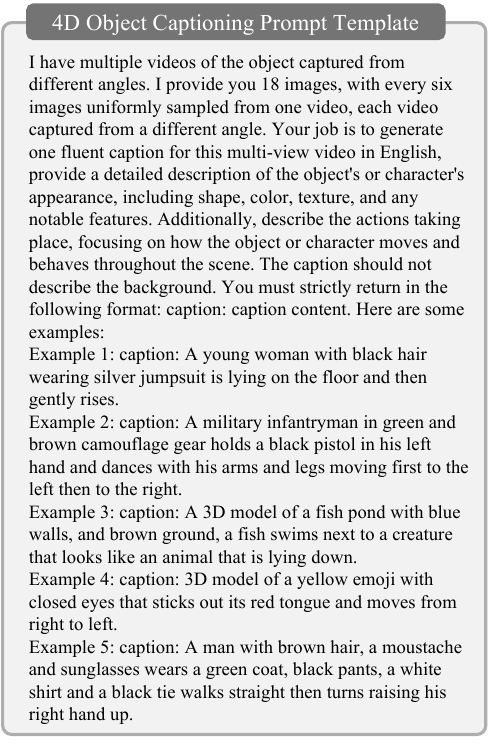}
    \caption{The prompt provided to the evaluated MLLMs in the 4D object captioning task. In this prompt, we describe the video information, caption requirement, and output format. We also provide several caption examples to guide the style of captions generated by MLLMs.}
    \label{fig:caption_prompt}
\end{figure}

\cref{fig:caption_prompt} shows the prompt we use to prompt the MLLMs to generate captions.
It's notable that we give them caption examples because we found that different MLLMs may generate captions in vastly different styles when not provided with examples, which could impact the results due to stylistic variations.
By providing examples, we aim to minimize the influence of different writing styles, allowing us to control experimental variables better and obtain more objective evaluation results that reflect the models' actual understanding capabilities rather than differences in writing style.

\subsection{More experimental details of 4D object question answering}
\begin{figure}[hbt]
    \centering
    \includegraphics[width=1.0\linewidth]{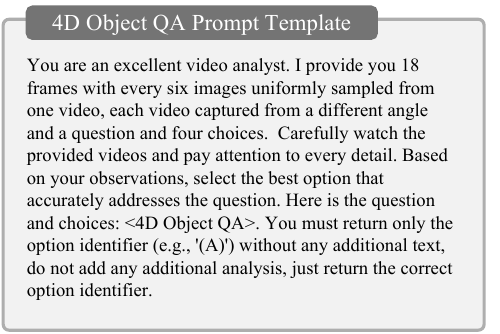}
    \caption{The prompt provided to the evaluated MLLMs in the 4D object QA task. In this prompt, we detailed the video information, questions and options, and the output format.}
    \label{fig:vqa_prompt}
\end{figure}

In the 4D object question answering experiments, all models operate under a shared function $P(A) = M(V, t, QA)$, where $V$, $t$, $QA$, $M$, $A$, and $P$ represent the multi-view video input, text prompt (instruction), question and four answer options, MLLM being tested, model output, and output processor, respectively. We add output processor to extract the selected answer option as we found that some open-source models sometimes struggled to strictly follow the prompt instructions that explicitly defined the required output format. 
\cref{fig:vqa_prompt} shows the prompt we use to prompt the MLLMs to complete the 4D Object QA task.

Since some open-source MLLMs may not always strictly follow the specified output format requirements, we implemented an output processor function to standardize answer extraction using the following code. This function is designed to extract a single letter answer choice (A, B, C, or D) from model responses. It first attempts to find a letter enclosed in parentheses (e.g., "(A)"). If no match is found, it looks for standalone letters that are bordered by spaces or punctuation marks to ensure consistent extraction regardless of the response format.

% The processor finds letter answers by checking for parentheses format "(A)" first, then for standalone letters if needed.
\lstset{
  language=Python,         % 指定语言
  basicstyle=\ttfamily\footnotesize,    % 基本字体
  keywordstyle=\color{blue}, % 关键字颜色
  commentstyle=\color{green}, % 注释颜色
  stringstyle=\color{red},   % 字符串颜色
  frame=single,           % 边框样式
  breaklines=true         % 自动换行
}
\begin{lstlisting}
def extract_answer_option(text):
    paren_pattern = r'\(([A-D])\)'
    matches = re.findall(paren_pattern, text)
    if matches:
        return matches[0]
    isolated_pattern = r'(?:^|[\s\(\.,;:])([A-D])(?:[\s\)\.,;:]|$)'
    matches = re.findall(isolated_pattern, text)
    if matches:
        return matches[0]
    return None
\end{lstlisting}

\section{Additional evaluation results on 4D-Bench}
In this section, we first provide additional analysis for the 4D object captioning in Sec. \ref{sec:analysis}. Then, Sec. \ref{sec:4dbench_examples} and Sec. \ref{sec:4dbench_qa_examples} provide additional evaluation results on the 4D object captioning and 4D object QA tasks of 4D-Bench, respectively.

\subsection{Analysis for 4D object captioning evaluation}
\label{sec:analysis}

\subsection{Additional qualitative results of 4D object captioning }
\label{sec:4dbench_examples}
\begin{figure}[t]
    \centering
    \includegraphics[width=0.85\linewidth]{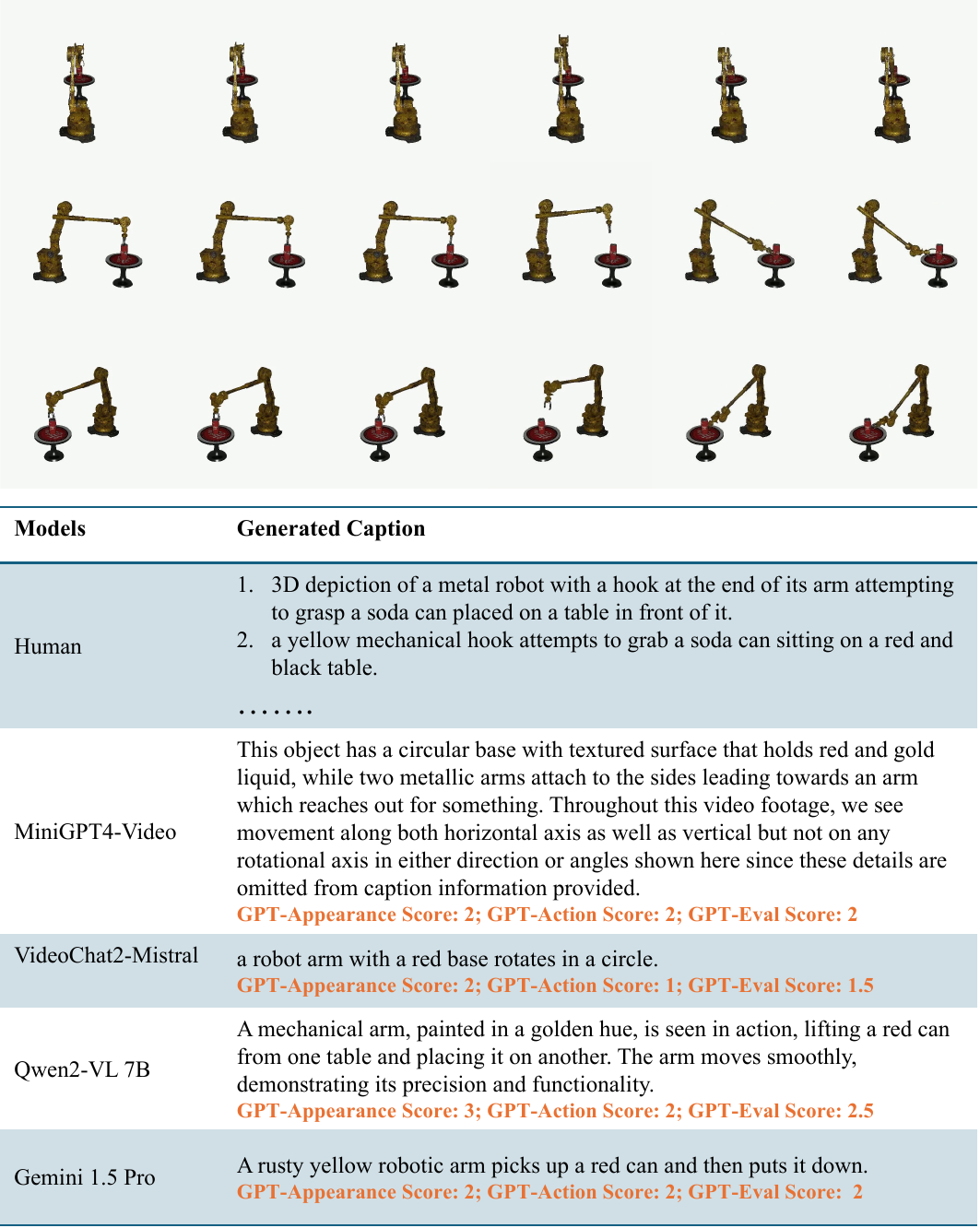}
     \caption{Qualitative results of different MLLMs on the 4D object captioning task of 4D-Bench}
    \label{fig:caption_exp1}
\end{figure}

\begin{figure}[t]
    \centering
    \includegraphics[width=0.85\linewidth]{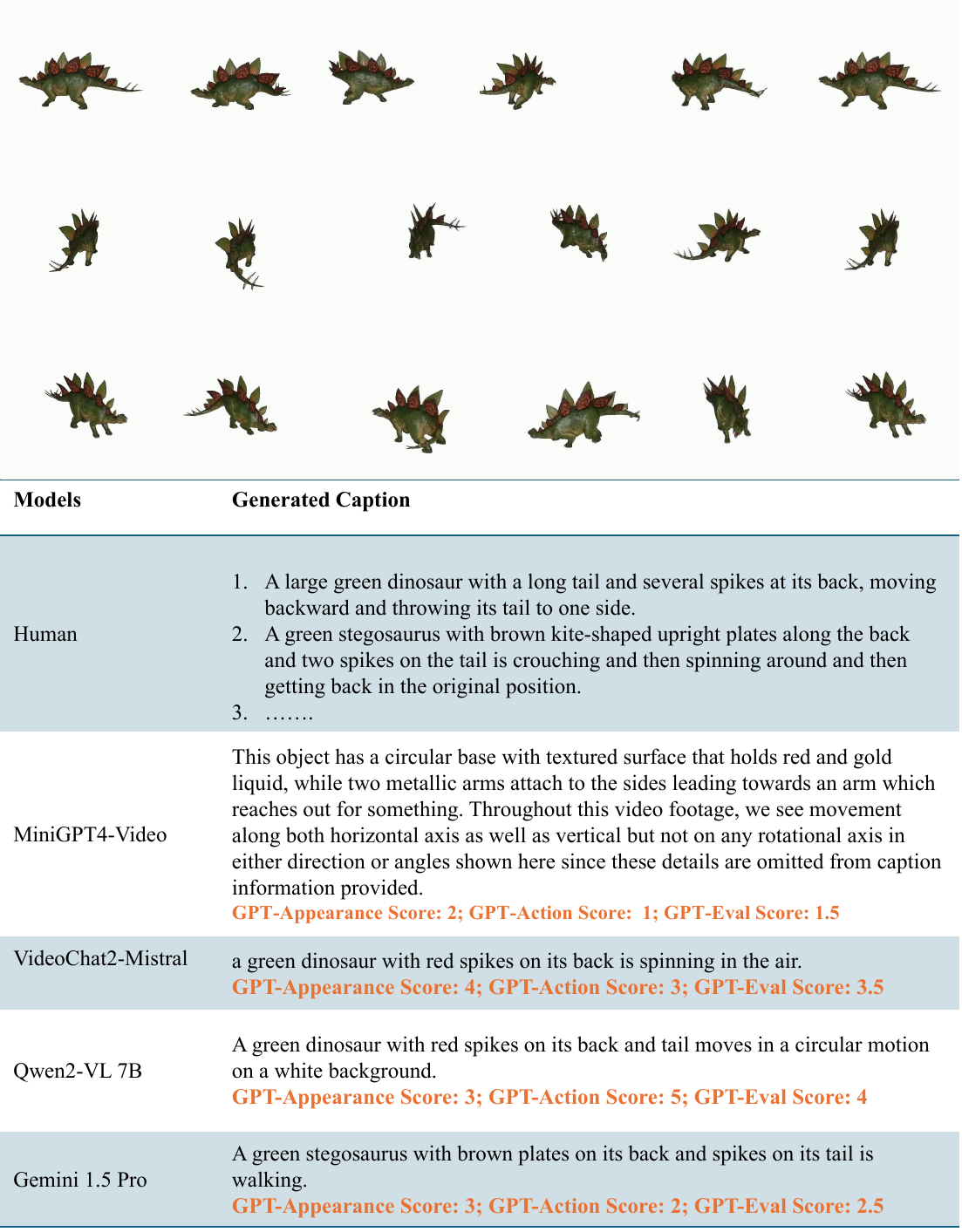}
     \caption{Qualitative results of different MLLMs on the 4D object captioning task of 4D-Bench}
    \label{fig:caption_exp2}
\end{figure}

\begin{figure}[hbt]
    \centering
    \includegraphics[width=0.85\linewidth]{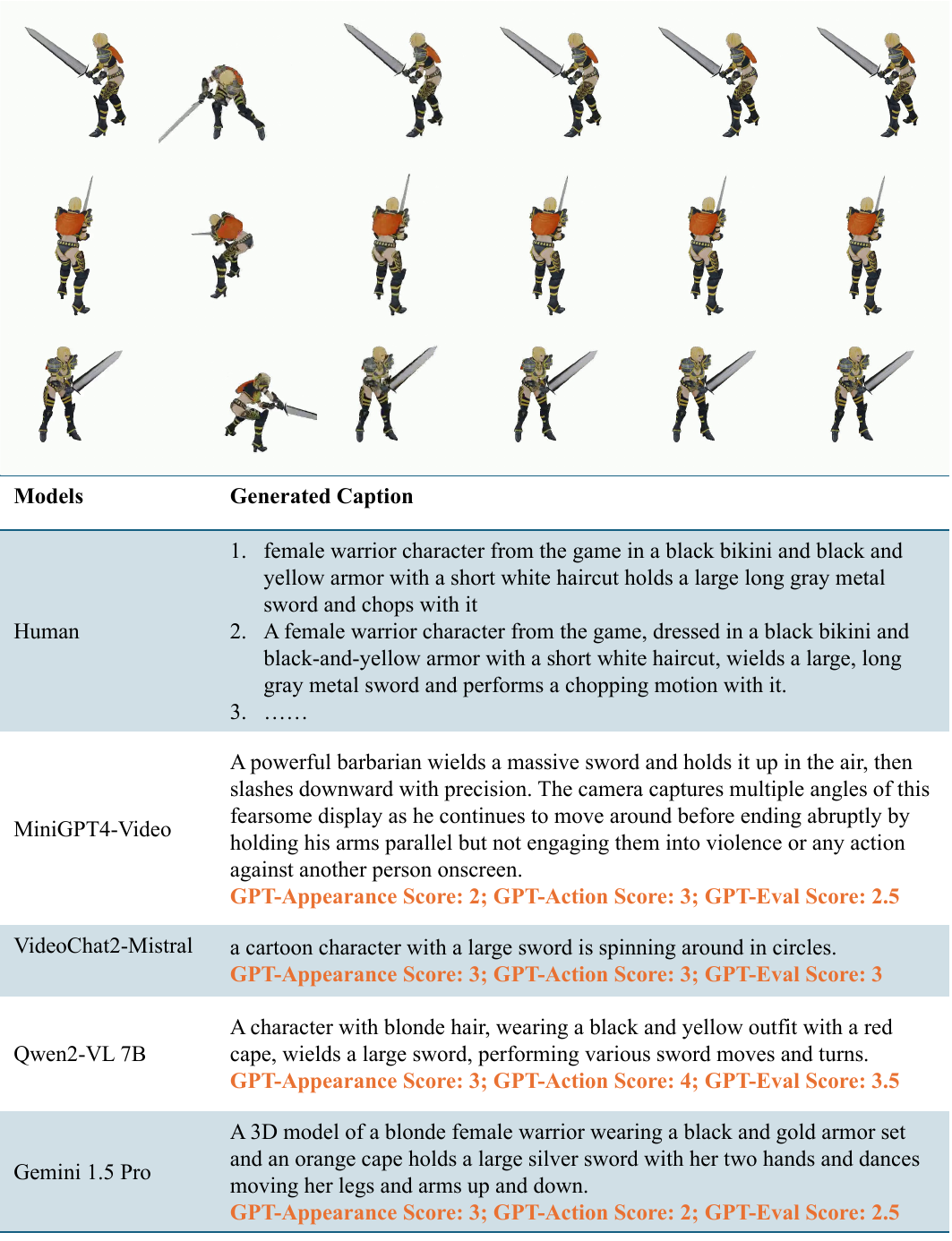}
     \caption{Qualitative results of different MLLMs on the 4D object captioning task of 4D-Bench}
    \label{fig:caption_exp3}
\end{figure}

\begin{figure}[hbt]
    \centering
    \includegraphics[width=0.85\linewidth]{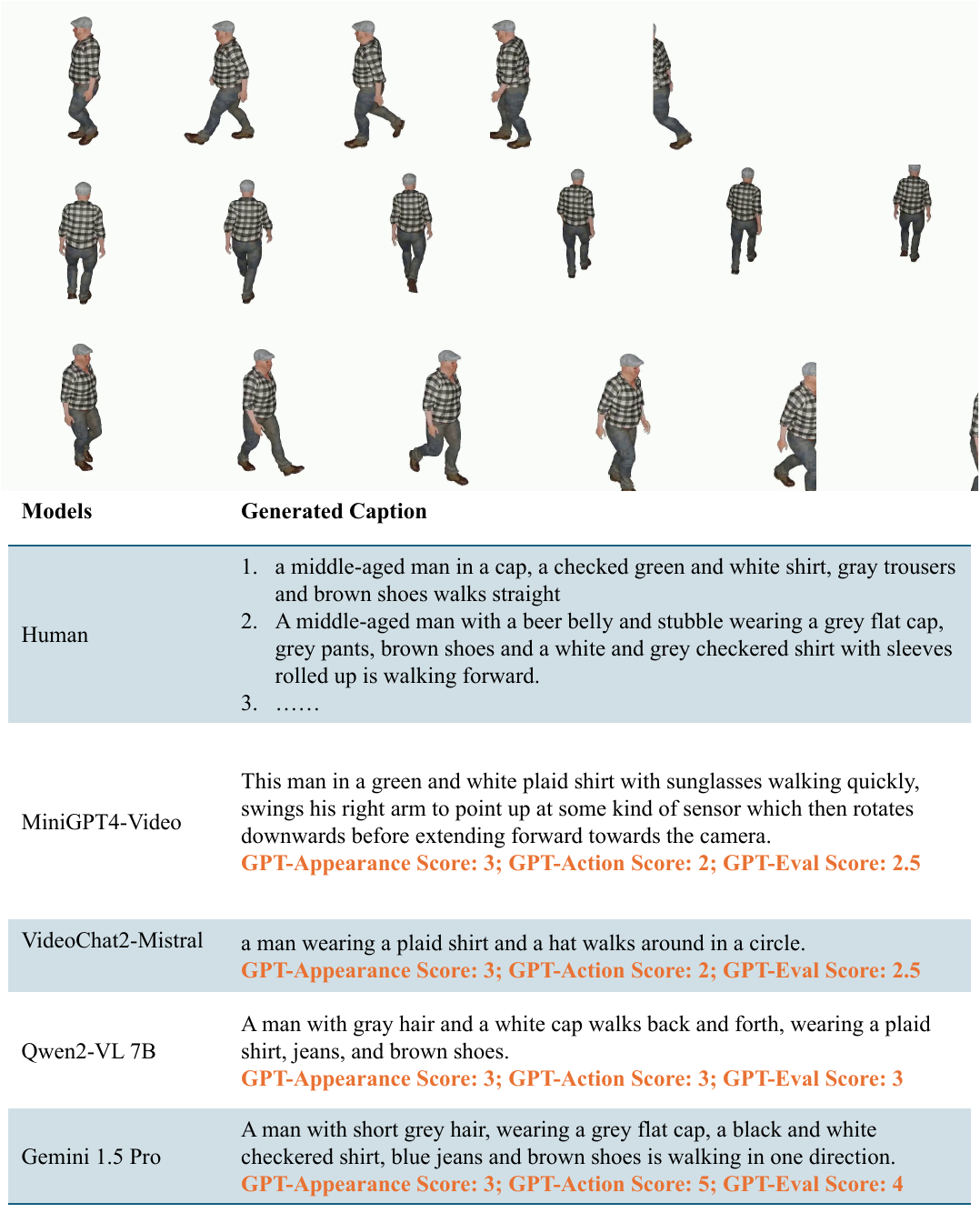}
    \caption{Qualitative results of different MLLMs on the 4D object captioning task of 4D-Bench}
    \label{fig:caption_exp4}
\end{figure}

% In this section, we provide some 4D object captioning examples and .
% You can visualize more results at \href{https://4dbench.github.io/#/caption}{\texttt{\textcolor{green}{https://4dbench.github.io/caption/}}}.
Figs. \ref{fig:caption_exp1}, \ref{fig:caption_exp2} \ref{fig:caption_exp3} and \ref{fig:caption_exp4} show  4D object captioning results of MiniGPT4-Video \cite{minigpt4-video}, VideoChat2-Mistral \cite{mvbench}, Qwen2-VL-7B \cite{qwen2_vl} and Gemini 1.5 Pro \cite{gemini2024},  given various 4D objects in our 4D-Bench. For example,  Fig. \ref{fig:caption_exp1} illustrates  MiniGPT4-Video, VideoChat2-Mistral, Qwen2-VL-7B, and Gemini 1.5 Pro achieve low GPT-Action scores.

\subsection{Additional qualitative results of 4D Object questing answering }
\label{sec:4dbench_qa_examples}
\begin{figure}[hbt]
    \centering
    \includegraphics[width=1\linewidth]{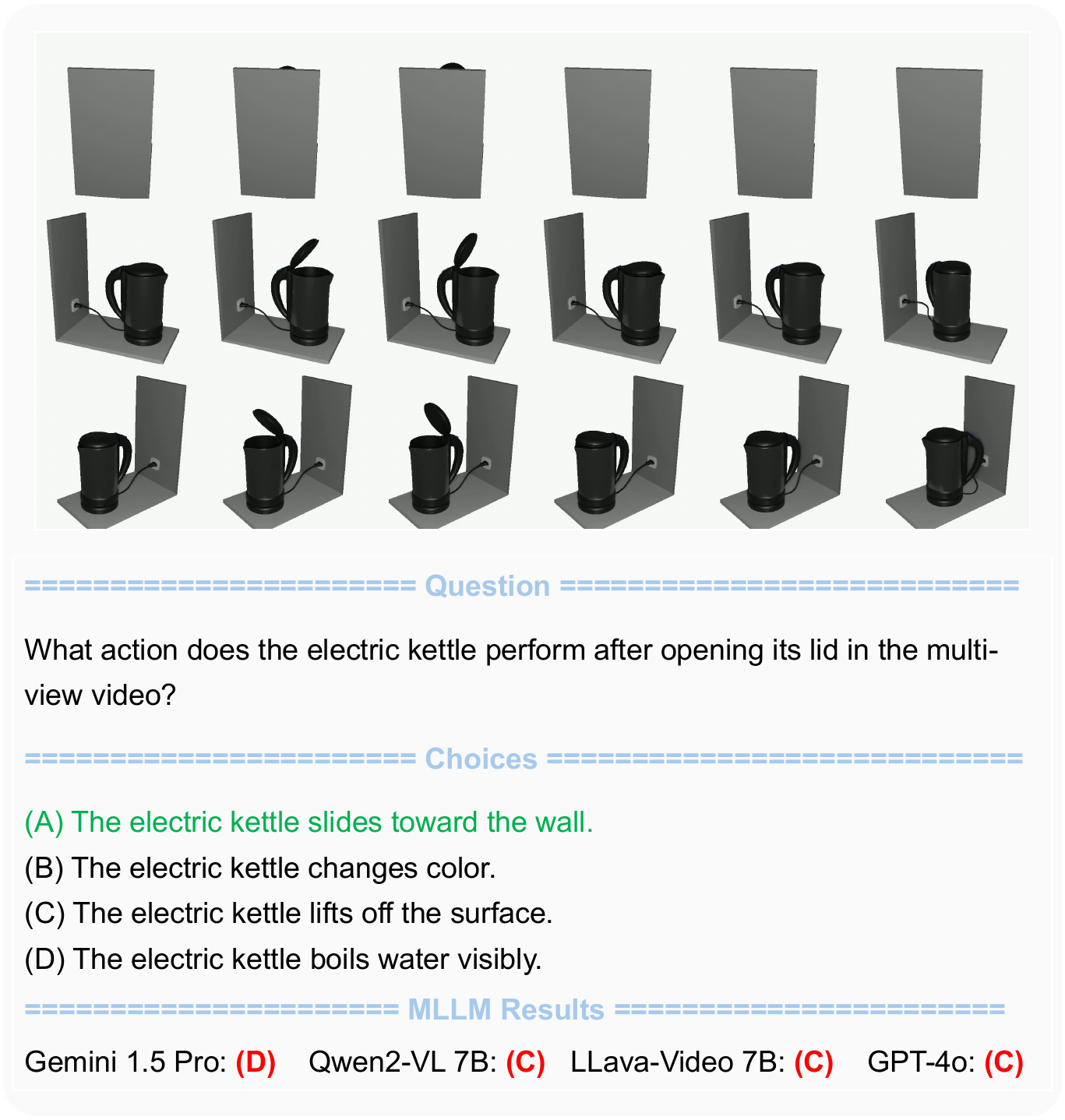}
    \caption{Qualitative results of different MLLMs on the 4D object question answering task of 4D-Bench}
    \label{fig:vqa_exp1}
\end{figure}

\begin{figure}[hbt]
    \centering
    \includegraphics[width=1\linewidth]{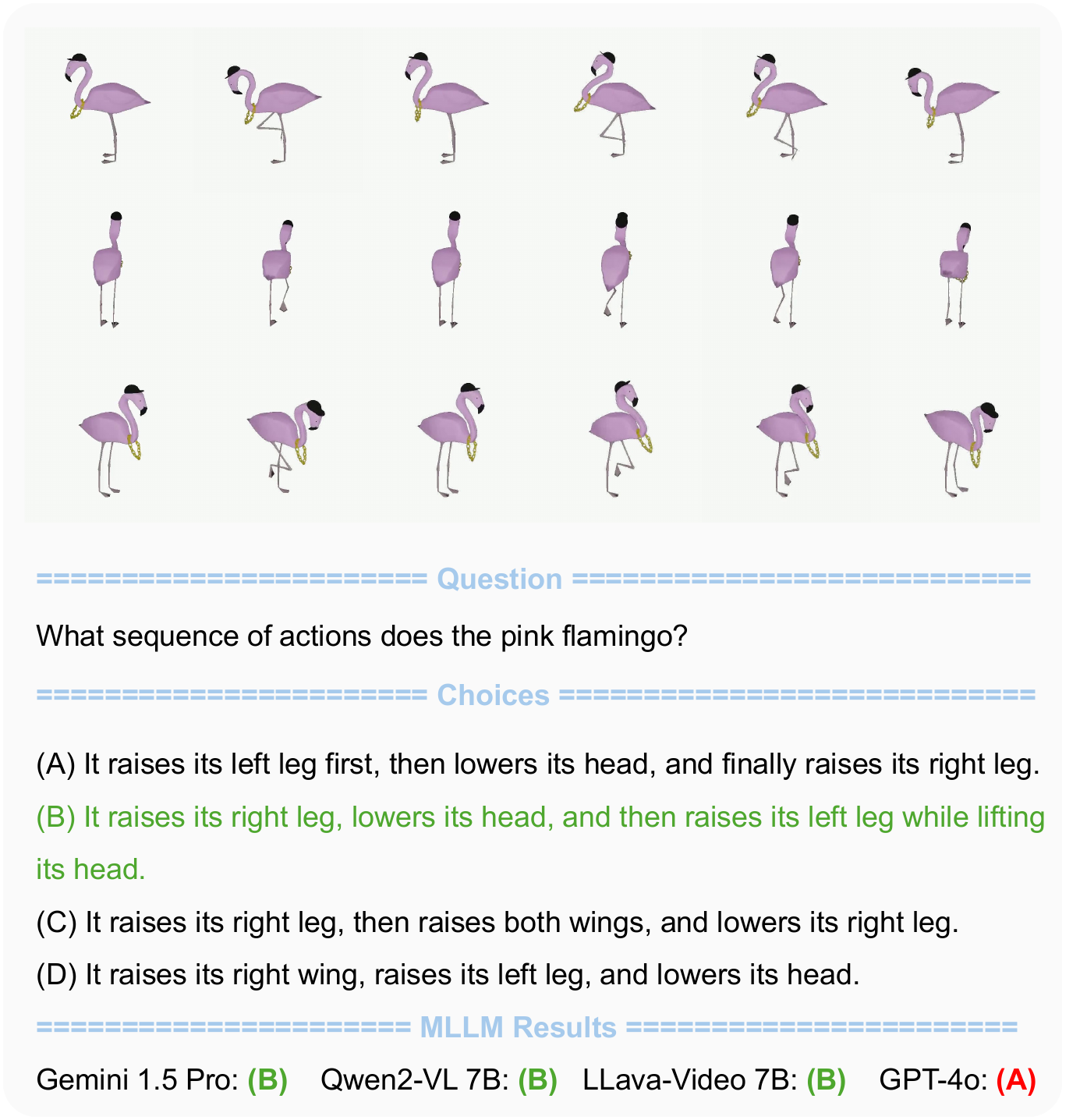}
    \caption{Qualitative results of different MLLMs on the 4D object question answering task of 4D-Bench}
    \label{fig:vqa_exp2}
\end{figure}

\begin{figure}[hbt]
    \centering
    \includegraphics[width=1.0\linewidth]{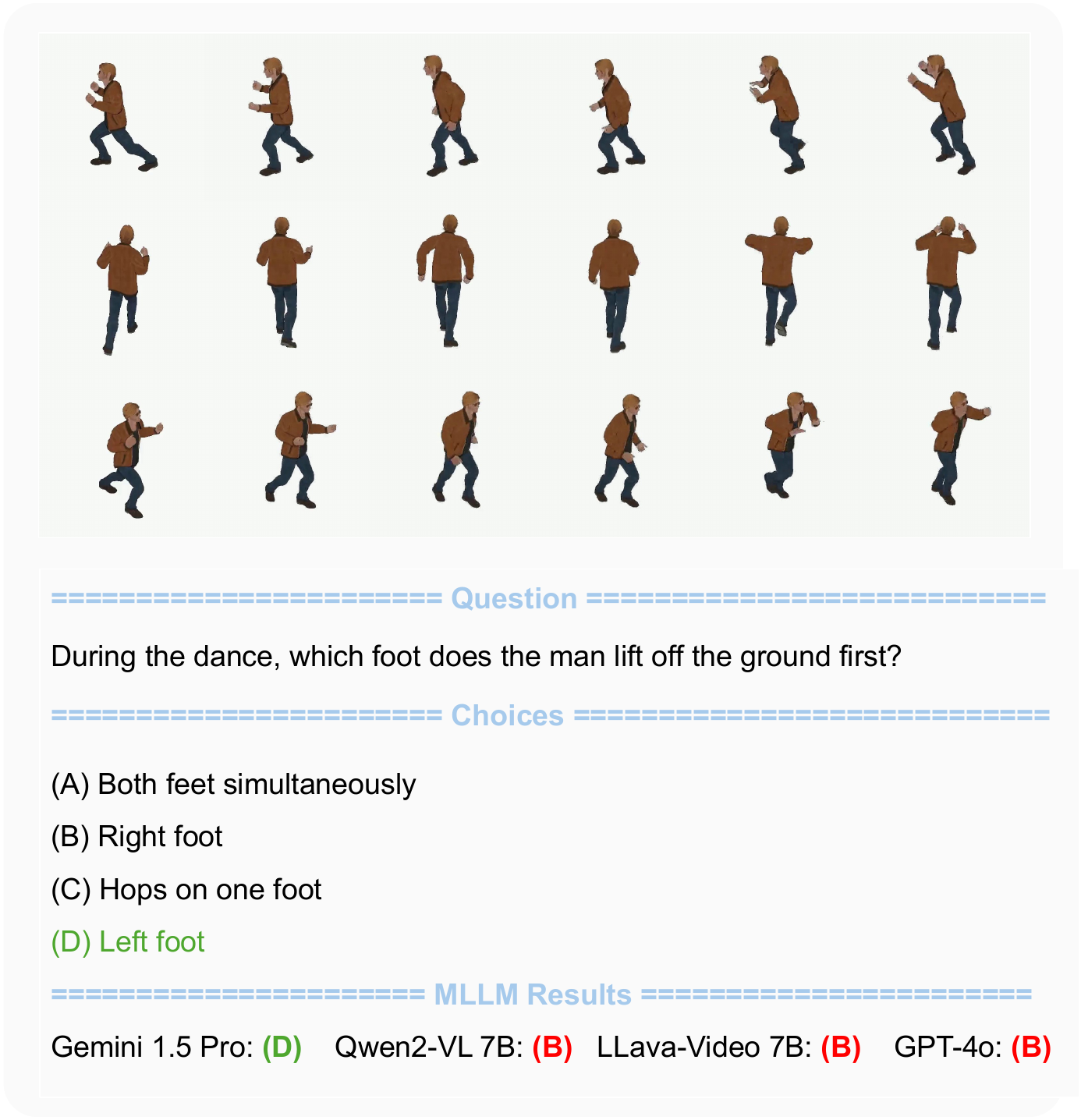}
    \caption{Qualitative results of different MLLMs on the 4D object question answering task of 4D-Bench}
    \label{fig:vqa_exp3}
\end{figure}

\begin{figure}[hbt]
    \centering
    \includegraphics[width=1.0\linewidth]{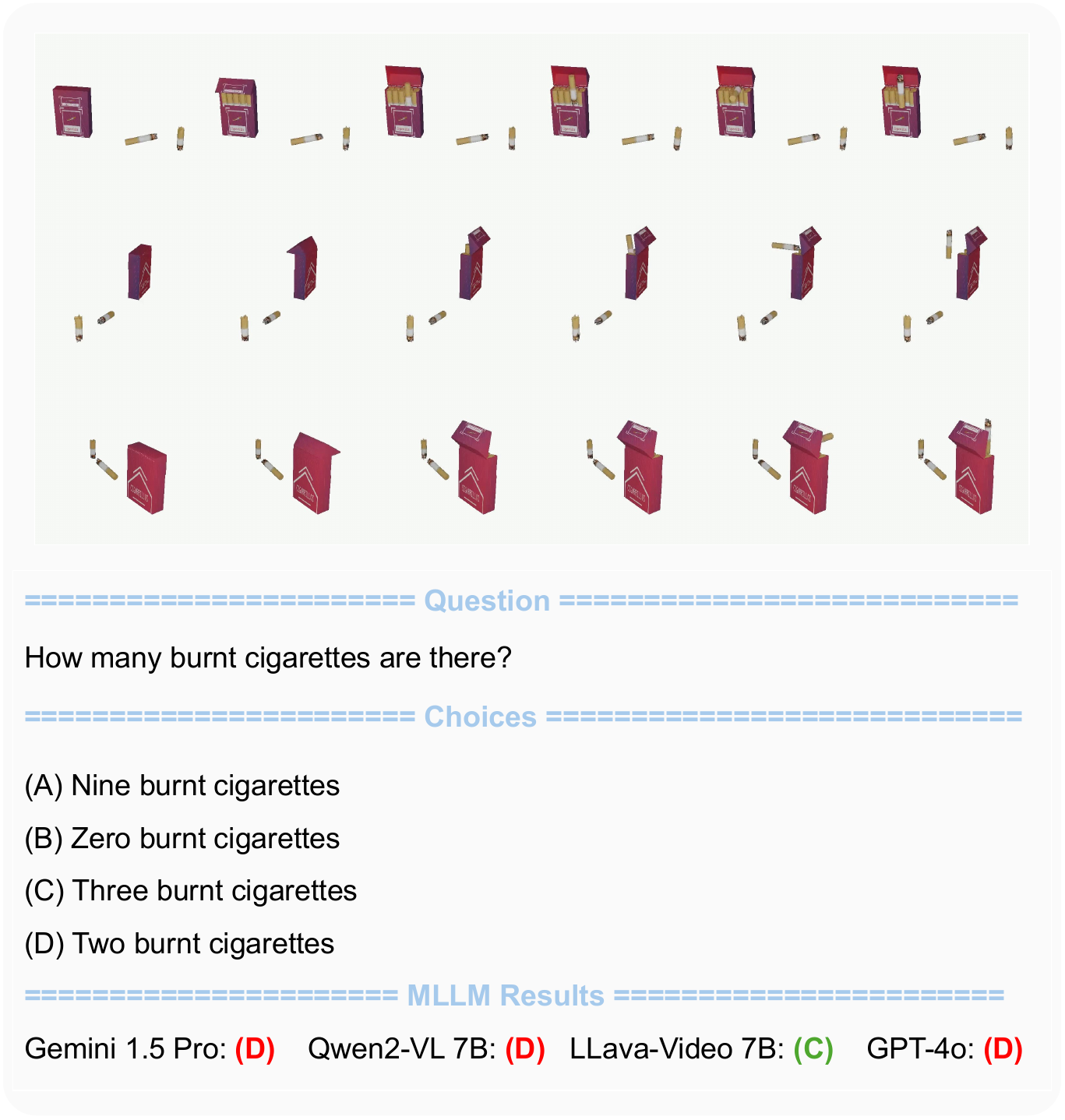}
    \caption{Qualitative results of different MLLMs on the 4D object question answering task of 4D-Bench}
    \label{fig:vqa_exp4}
\end{figure}

Figs. \ref{fig:vqa_exp1},  \ref{fig:vqa_exp2}, \ref{fig:vqa_exp3} and \ref{fig:vqa_exp4} illustrate more 4D object QA results of advanced MLLMs. 
Fig. \ref{fig:vqa_exp2} shows an easy sample on the subtask of \textit{Temporal Relationship}, where all MLLMs choose the correct answer except for GPT-4o.
Fig. \ref{fig:vqa_exp3} shows a more difficult example of \textit{Temporal Relationship}, where Qwen2-VL 7B, GPT-4o and LLava-Video picks the wrong answer.
Fig. \ref{fig:vqa_exp4} shows qualitative results of MLLMs on the \textit{Object Counting} subtask, where only LLava-Video 7B answered the question correctly. 
Fig. \ref{fig:vqa_exp1} illustrates all MLLMs (including GPT-4o and Gemini 1.5 pro)  pick the wrong option on the subtask of \textit{Action}, indicating the limited capabilities of MLLMs in action understanding of 4D objects. 
\clearpage

\clearpage

% \vspace{3cm}  % 根据需要调整具体数值

% \vspace{-1cm}  % 根据需要调整具体数值

\end{document}